\documentclass[sigconf,nonacm]{acmart} 

\usepackage{enumitem}

\usepackage{algorithm}
\usepackage{algpseudocode}
\algtext*{EndProcedure}
\algtext*{EndFunction}
\algtext*{EndWhile}
\algtext*{EndFor}
\algtext*{EndIf}
\algrenewcommand{\algorithmicreturn}{\hspace{\algorithmicindent}\State \textbf{return}}
\algrenewcommand\textproc{}

\usepackage[noabbrev,capitalise]{cleveref}

\usepackage{mathtools}

\usepackage{subcaption}

\usepackage[binary-units,per-mode=symbol]{siunitx}

\usepackage{soul}

\usepackage{comment}

\AtBeginDocument{%
  \providecommand\BibTeX{{%
    \normalfont B\kern-0.5em{\scshape i\kern-0.25em b}\kern-0.8em\TeX}}}

\copyrightyear{2021}
\acmYear{2021} 
\setcopyright{acmcopyright}\acmConference[GECCO '21]{2021 Genetic and Evolutionary Computation Conference}{July 10--14, 2021}{Lille, France}
\acmBooktitle{2021 Genetic and Evolutionary Computation Conference (GECCO '21), July 10--14, 2021, Lille, France}
\acmPrice{15.00}
\acmDOI{10.1145/3449639.3459311}
\acmISBN{978-1-4503-8350-9/21/07}

\begin{document}

\title[Seeking Quality Diversity in Evolutionary Co-design of Soft Tensegrity Modular Robots]{Seeking Quality Diversity in Evolutionary Co-design of Morphology and Control of Soft Tensegrity Modular Robots}{}



\author{Enrico Zardini}
\affiliation{%
  \institution{University of Trento}
  \city{Trento}
  \country{Italy}
}
\orcid{0000-0002-7475-7183}
\email{enrico.zardini@unitn.it}

\author{Davide Zappetti}
\affiliation{%
  \institution{\'{E}cole Polytechnique F\'{e}d\'{e}rale de}
  \city{Lausanne}
  \country{Switzerland}
}
\orcid{}
\email{davide.zappetti@epfl.ch}

\author{Davide Zambrano}
\affiliation{%
  \institution{\'{E}cole Polytechnique F\'{e}d\'{e}rale de}
  \city{Lausanne}
  \country{Switzerland}
}
\orcid{}
\email{davide.zambrano@epfl.ch}

\author{Giovanni Iacca}
\affiliation{%
  \institution{University of Trento}
  \city{Trento}
  \country{Italy}
}
\orcid{0000-0001-9723-1830}
\email{giovanni.iacca@unitn.it}

\author{Dario Floreano}
\affiliation{%
  \institution{\'{E}cole Polytechnique F\'{e}d\'{e}rale de}
  \city{Lausanne}
  \country{Switzerland}
}
\orcid{}
\email{dario.floreano@epfl.ch}

\renewcommand{\shortauthors}{Zardini et al.}



\begin{abstract}
Designing optimal soft modular robots is difficult, due to non-trivial interactions between morphology and controller. Evolutionary algorithms (EAs), combined with physical simulators, represent a valid tool to overcome this issue. In this work, we investigate algorithmic solutions to improve the Quality Diversity of co-evolved designs of Tensegrity Soft Modular Robots (TSMRs) for two robotic tasks, namely goal reaching and squeezing trough a narrow passage. To this aim, we use three different EAs, i.e., MAP-Elites and two custom algorithms: one based on Viability Evolution (ViE) and NEAT (ViE-NEAT), the other named Double Map MAP-Elites (DM-ME) and devised to seek diversity while co-evolving robot morphologies and neural network (NN)-based controllers. In detail, DM-ME extends MAP-Elites in that it uses two distinct feature maps, referring to morphologies and controllers respectively, and integrates a mechanism to automatically define the NN-related feature descriptor. Considering the fitness, in the goal-reaching task ViE-NEAT outperforms MAP-Elites and results equivalent to DM-ME. Instead, when considering diversity in terms of ``illumination'' of the feature space, DM-ME outperforms the other two algorithms on both tasks, providing a richer pool of possible robotic designs, whereas ViE-NEAT shows comparable performance to MAP-Elites on goal reaching, although it does not exploit any map.
\end{abstract}

\ccsdesc[500]{Computing methodologies~Evolutionary robotics}
\ccsdesc[500]{Computing methodologies~Genetic algorithms}

\keywords{Soft Tensegrity Modular Robots, Co-evolution, NEAT, Viability Evolution, MAP-Elites, Quality Diversity}

\maketitle


\section{Introduction}
\label{sec:intro}
Soft robots might be one of the key technologies of the future. Indeed, their robustness and adaptive morphology allow them to overcome situations and perform tasks in which traditional hard robots are of limited applicability \cite{mintchev2016adaptive}. For example, they can perform locomotion on rough terrains with risking damage \cite{mintchev2018soft}, or squeeze through narrow passages \cite{cheney2015evolving}. Hence, they can be employed in the exploration of hard-to-access environments \cite{hallawa2020morphological}, as well as in the medical field \cite{zhang2019worm}. Recently, researchers have applied the soft robotic paradigm to modular robotics \cite{zhang2020modular}, developing novel soft modular robots able to explore a larger morpho-functional space than their rigid counterparts while exhibiting robustness and mechanical adaptability \cite{germann2014soft,mintchev2016adaptive}.

Designing soft modular robots is notably difficult due to the hard-to-model dynamics of soft materials and the non-trivial interactions between morphology and controller \cite{laschi2016soft}. These difficulties, together with the lack of analytical methods, make evolutionary algorithms (EAs), coupled with physics simulators, one of the most promising design tools \cite{howard2019evolving}. Indeed, not only can EAs discover unconventional high-performing solutions, but they can also provide a pool of possible designs by "illuminating" the design space \cite{mouret2015illuminating}. This latter aspect is especially significant in soft modular robots, where the influence of the morphology and controller parameters on the robot performance is hard to predict, thus having a number of eligible candidates for the physical realization is desirable. We should remark however that, although other paradigms to design soft modular robots exist \cite{howison2020reality, lee2017soft, vergara2017soft, lee2016soft, morin2014elastomeric}, only a few involve the use of EAs: notably and extensively the voxel-based one (VSRs) \cite{cheney2013unshackling,shah2020shape} and the tensegrity soft modular robots (TSMRs) \cite{zappetti2017bioinspired}, the latter in a preliminary work with an open-loop controller \cite{Zappetti_2018} and in another work \cite{tsmrbehaviouralrepertoires} where the Multi-dimensional Archive of Phenotypic Elites (MAP-Elites) \cite{mouret2015illuminating} was applied to diversify only behavioral properties rather than morphological ones.

This work also focuses on TSMRs. More precisely, on the joint optimization of TSMR morphologies and NN-based controllers, as the optimization of either morphology only or controller only via EAs does not leverage the potential synergy between them. Specifically, we perform the co-optimization using three different approaches, namely MAP-Elites and two proposed custom EAs: 1) Viability Evolution (ViE) \cite{vie} coupled with Neuro-Evolution of Augmenting Topologies (NEAT) \cite{neat}, denoted as \emph{ViE-NEAT}, and 2) a method based on MAP-Elites, dubbed as \emph{Double Map MAP-Elites} (DM-ME). The latter makes use of two maps, associated to an entity-related and a controller-related feature descriptor (FD), respectively, and includes a dimensionality reduction mechanism \cite{dr_pca_me} for the automatic definition of the controller-related FD.

We experimentally compare the three approaches on two robotic tasks: goal reaching and squeezing through a narrow passage. Our results show that DM-ME outperforms the other two approaches at illuminating the search space on both tasks. In addition, although ViE-NEAT does not keep any map, it achieves very similar performance to MAP-Elites on goal reaching (in terms of illumination).

The rest of this paper is structured as follows: \cref{sec:background} deals with the background and related work; \cref{sec:dm-me} introduces the proposed methods, i.e., ViE-NEAT and DM-ME; \cref{sec:setup} describes the experimental setup; \cref{sec:results} presents the numerical results. Finally, \cref{sec:conclusion} provides the main conclusions of our study.


\section{Background}
\label{sec:background}
This section introduces the TSMRs employed in this work and the EAs related to the algorithms we have considered. It also introduces the mechanisms that have been integrated in DM-ME.

\subsection{Tensegrity soft modular robots}
\label{subsec:tsmr}
The term \emph{tensegrity} has been coined by the architect R. Buckminster Fuller \cite{fuller1961tensegrity} and denotes a structure that maintains its mechanical integrity through the pre-stretching of some elements constantly in tension (known as strings or cables) connected in a network with other elements constantly under compression (called rods or struts). Several kinds of tensegrity structures exist: this work focuses on the icosahedron tensegrity, which consists of 6 rods connected by 24 pre-stretched cables. This particular structure has been chosen due to its properties: the ability to deform in all directions; the symmetry; the small number of rods and cables; the possibility to carry a payload in the inner volume \cite{zappetti2017bioinspired}. The last point is particularly important since it allows to equip the structure with a servomotor, which enables structural contractions, and different kinds of sensors.

The term \emph{modular} denotes robots composed of a variable number of building blocks. In this work, the icosahedron tensegrity has been used as a module. In detail, the nodes of the icosahedron draft eight triangular faces, which are exploited to link the modules together by connecting the triangles' vertices. In the experiments, we take in consideration only linear chains of modules, both for simplicity and feasibility of a future hardware implementation. The appearance of a TSMR in the simulation framework that we have used (see \cref{subsec:framework}) is shown in \cref{fig:tasks}.

The term \emph{soft} refers to the fact that these robots are not rigid. Indeed, they are able to deform their structure by contracting and expanding their modules.

\subsection{Neuro-evolution of augmenting topologies}
\label{subsec:neat}
NEAT \cite{neat} is a well-known neuro-evolutionary algorithm capable of evolving both topology and weights of artificial neural networks (NNs). It exploits an advanced genetic encoding based on the \emph{historical marking} mechanism. In practice, each topological feature is marked with a number that allows to match homologous genes while performing crossover; in this way different genotypes can be aligned and crossed over. Moreover, the initialization of the population is addressed through a \emph{complexification} process: the initial population consists of minimal networks, which are gradually made more complex as the search progresses. In this way the evolution will focus on the most promising parts of a very high-dimensional search space. Diversity and innovations are preserved through speciation: the population is split into species by means of a clustering method that groups individuals by similarity and the competition is limited to individuals within each species. It is also worth mentioning the mechanism known as species-elitism, which preserves a certain number of species (the best ones) from extinction due to stagnation, i.e., no improvement for long time. In this work, we used the NEAT implementation provided by \texttt{NEAT-Python}\footnote{\url{https://github.com/CodeReclaimers/neat-python}}.

\subsection{Diversity-driven evolutionary algorithms}
\label{subsec:diversity-eas}
Among the various EAs proposed in the literature for seeking diversity (either explicitly or implicitly) the following two have been taken into account here: Viability Evolution (``ViE'') \cite{vie, maesani2016memetic}, which is characterized by a good balance between exploitation and exploration, and MAP-Elites \cite{mouret2015illuminating}, which is focused on ``illuminating'' the search space.

ViE is an evolutionary paradigm based on the elimination of \textit{unviable} individuals. In detail, ViE makes use of \textit{viability boundaries} that discriminate between viable individuals, which satisfy specific requirements, and unviable ones, not satisfying them. The boundaries in question are dynamic: at the beginning they are set so as to include all the individuals belonging to the population; then, they are progressively shrunk either towards the optimal values or towards target values that correspond to some constraints. In this work we used ViE to evolve a population of TSMR morphologies; since there are no specific constraints in this case, the viability boundary represents a limit on the worst fitness.

MAP-Elites is a Quality Diversity (QD) algorithm, i.e., an algorithm that aims at generating a large collection (archive) of diverse and high-performing solutions (see \cite{qd_score,multi_maps_me} for more details on QD). In particular, it aims at discovering the best-performing solution of each cell of a user-defined feature space, which is a lower dimensional space with respect to the original search space. In order to be added to the archive, a solution is first projected onto the feature space, producing a feature descriptor; if the corresponding cell is empty or contains an individual with worse performance, the solution is added to the archive (replacing the occupant, if present), otherwise it is discarded. In this work we used MAP-Elites to evolve morphology-controller pairs.

\subsection{MAP-Elites with multiple maps}
\label{subsec:double-map-in-literature}
Multiple feature maps have already been employed in previous research, although for the evolution of \emph{a single entity}. For example, in \cite{multi_maps_me}, two MAP-Elites archives are employed for the evolution of the controller of a wheeled robot in a maze navigation task. The two FDs used there are the endpoint of the simulation and the main orientation of the robot for each fifth of the simulation; thus, they are two behavioral properties of the evolved controller. Multiple archives are used also in \cite{openendedmulticontainers} (although with a different QD algorithm), but even there the FDs represent only behavioral features. Instead, the DM-ME algorithm introduced here is devised for the evolution of \emph{two entities}: a generic entity (in our case, a TSMR morphology) and its related controller (a NN). Hence, the FDs of the two archives refer to distinct objects.

\subsection{Automatic feature descriptor definition}
\label{subsec:automatic-fd-definition}
The selection of the FD is a critical factor for MAP-Elites. A mechanism to define it in an automatic way has been introduced in \cite{dr_pca_me}. In practice, the FD is obtained by applying a Dimensionality Reduction (DR) algorithm, i.e., the Principal Component Analysis (PCA), on the sensory data extracted during the evaluation of the individual; the DR's output vector represents the FD that is used for the insertion in the archive. Nevertheless, the FD in question consists of continuous values, which is acceptable for an unstructured archive, such as the one used in \cite{dr_pca_me}, but not for MAP-Elites. Hence, an additional processing step to discretize the feature values is required. Basically, a first discretization to determine the cells boundaries is performed at the beginning of the evolution using the values retrieved by the PCA for the initial population. Then, the boundaries are recomputed every time the PCA is fitted to data or a value outside the current boundaries is returned. It is worth also mentioning that the PCA is scale-sensitive; hence, the sensory data are not directly provided as input to the PCA, but is standardized first by removing the mean and dividing by the standard deviation. As regards the \textit{standardizer}, it is fitted to data every time the fitting is done for the PCA. In conclusion, the only information that is required is the size of the FD (i.e., the number of cells per dimension).


\section{Proposed Methods}
\label{sec:dm-me}
We present now the algorithmic details of the proposed methods, i.e., ViE-NEAT and Double Map MAP-Elites (DM-ME).

\subsection{ViE-NEAT}
ViE-NEAT consists in the parallel evolution of two populations, a population of entities (morphologies) and one of NN-based controllers. The former is evolved using ViE, the latter using NEAT. The only interaction between the two populations happens at the evaluation time: in detail, at each generation the individuals belonging to the two populations are randomly paired for evaluation. In principle pairing should be one-to-one. However, since in the NEAT implementation used here the population is not fixed in size, it may happen that an individual from one population is paired with two individuals from the other, due to the bigger size of the other population at play. The pseudocode is shown in \cref{alg:vie-neat-coev}.

\subsection{Double Map MAP-Elites (DM-ME)}

DM-ME deals with individuals represented by an <entity-NN> pair: the entity encoding depends on the domain of application, whereas NEAT's genetic encoding is used for NNs.

\begin{algorithm}[!th]
    \caption{ViE-NEAT}\label{alg:vie-neat-coev}
    \begin{algorithmic}[1]
        \Procedure{main}{}():
            \State $P_{ViE}$ $\gets$ initViEPopulation() \Comment{entity pop.}
            \State $P_{NEAT}$ $\gets$ initNEATPopulation() \Comment{NN pop.}
            \For {$t$ = 1 $\rightarrow$ $T$}
                \State evaluate($P_{ViE}$, $P_{NEAT}$) \Comment{random pairing used}
                \State $P_{ViE}$ $\gets$ runOneGeneration($P_{ViE}$)
                \State $P_{NEAT}$ $\gets$ runOneGeneration($P_{NEAT}$)
            \EndFor
        \EndProcedure
    \end{algorithmic}
\end{algorithm}
\begin{algorithm}[!th]
    \caption{Double Map Map-Elites (DM-ME)}\label{alg:dm-me}
    \begin{algorithmic}[1]
        \Procedure{main}{$numInitialSolutions$, $archiveSize$}:
            \State $A_E$ $\gets$ $\emptyset$ \Comment{entity archive}
            \State $A_{NN}$ $\gets$ $\emptyset$ \Comment{NN archive}
            \State $PCA$ $\gets$ initPCA($archiveSize$)
            \State $X$ $\gets$ generateRandomSolutions($numInitialSolutions$)
            \State evaluate($X$) \Comment{get fitness and sensory data}
            \State addAllToMap($A_E$, $X$)
            \State addToNNMap($A_{NN}$, $PCA$, $X$)
            \For {$t$ = 1 $\rightarrow$ $T$}
                \State $X$ $\gets$ randomSelection($A_E$, $A_{NN}$)
                \State $X$ $\gets$ randomVariation($X$)
                \State evaluate($X$)
                \State addAllToMap($A_E$, $X$)
                \State addToNNMap($A_{NN}$, $PCA$, $X$)
            \EndFor
        \EndProcedure
        \hrulefill
        \Procedure{addAllToMap}{$A$, $X$}:
            \For {$x$ $\in$ $X$}
                \State addToMap($A$, $x$)
            \EndFor
        \EndProcedure
        \hrulefill
        \Procedure{addToNNMap}{$A_{NN}$, $PCA$, $X$}:
            \State $SD_X$ $\gets$ getSensoryData($X$)
            \If {$PCA$.notFitted()}
                \State $PCA$.fit($SD_X$)
            \ElsIf {$isFittingTime$}
                \State $SD_A$ $\gets$ getSensoryData($A_{NN}$)
                \State $PCA$.fit($SD_A$)
                \State reInsert($A_{NN}$, $PCA$)
            \EndIf
            \State $PCA$.transform($X$, $SD_X$) \Comment{compute NN FD}
            \If {outOfBounds($X$)}
                \State $PCA$.reComputeBounds($A_{NN}$, $X$)
                \State reInsert($A_{NN}$, $PCA$)
                \State $PCA$.transform($X$, $SD_X$)
            \EndIf
            \State addAllToMap($A_{NN}$, $X$)
        \EndProcedure
    \end{algorithmic}
\end{algorithm}

The functioning of DM-ME is illustrated in \cref{alg:dm-me}. At the beginning, the archives are initialized with a large number of randomly generated <entity-NN> pairs, exploiting the corresponding sensory data to fit the PCA. Then, until the desired number of generations has been reached, a batch of elites is sampled from the two maps, half from each of them. Next, the individuals are mutated as follows: with equal probability, either only the entity is mutated, only the NN is mutated, or both of them are mutated. The mutation of the entity is domain-dependent, whereas the NNs are mutated according to NEAT's mutation operator (crossover is not used). After mutation, the individuals are evaluated and added to the archives based on the respective FDs. In this regard, each of the two archives is an independent MAP-Elites map, and the \texttt{addToMap()} archive insertion procedure is handled accordingly: an individual (an <entity-NN> pair) is added to an archive if the corresponding cell is empty or occupied by a worse individual, hence, it may be added to one map, but not to the other, depending on the fitness of the currently stored individuals. The peculiarity lies in the fact that a FD related to the entity is used for the first map, whereas a FD related to the NN is employed for the second one. As shown in \cite{multi_maps_me}, the use of multiple adequate FDs (in the form of multiple maps for MAP-Elites) leads to better results, especially in difficult domains, since different perspectives are taken into account. As for our DM-ME, the two FDs employed are obtained in different ways. In particular, the entity is projected as usual on a user-defined feature space, in order to obtain the corresponding FD. Instead, for the NN, the mechanism described in \cref{subsec:automatic-fd-definition} is used. More specifically, the PCA is fitted using the sensory data of the individuals contained in the NN archive with a frequency that decreases exponentially, as in \cite{dr_pca_me}; obviously, at every fitting, it is necessary to re-compute the FDs of the individuals that are already in the archive and re-insert them, which is done also when the space is discretized between two PCA updates due to the presence of values outside the current cells boundaries.


\section{Experimental setup}
\label{sec:setup}
This section presents the details of our experimental setup.

\subsection{Implementation details}
\label{subsec:framework}
Apart from NEAT, for which we used the existing \texttt{NEAT-Python} library, the other EAs involved in the experimentation were implemented from scratch in Python, and coupled with a custom TSMR simulation framework developed in C++, named Tensoft. The latter is based on the NASA Tensegrity Robotics Toolkit\footnote{\url{https://github.com/NASA-Tensegrity-Robotics-Toolkit/NTRTsim}}, a collection of tools for modeling and simulating tensegrity robots, built on the top of the Bullet Physics\footnote{\url{https://github.com/bulletphysics/bullet3/archive/2.88.tar.gz}} engine. Our source code is available at: \url{https://github.com/lis-epfl/Tensoft-G21}.

\subsection{Tasks}
\label{subsec:tasks}
TSMRs have only recently become a subject of study, as demonstrated by \cite{3d_printed_tsmr}; hence, apart for locomotion \cite{Zappetti_2018}, their capabilities are still largely unknown. Here, we investigate two tasks that are fairly more complex than plain locomotion: \emph{goal reaching} and \emph{squeezing}.

\begin{figure}[!tbh]
    \centering
    \begin{subfigure}{0.495\linewidth}
      \centering
      \includegraphics[width=\linewidth]{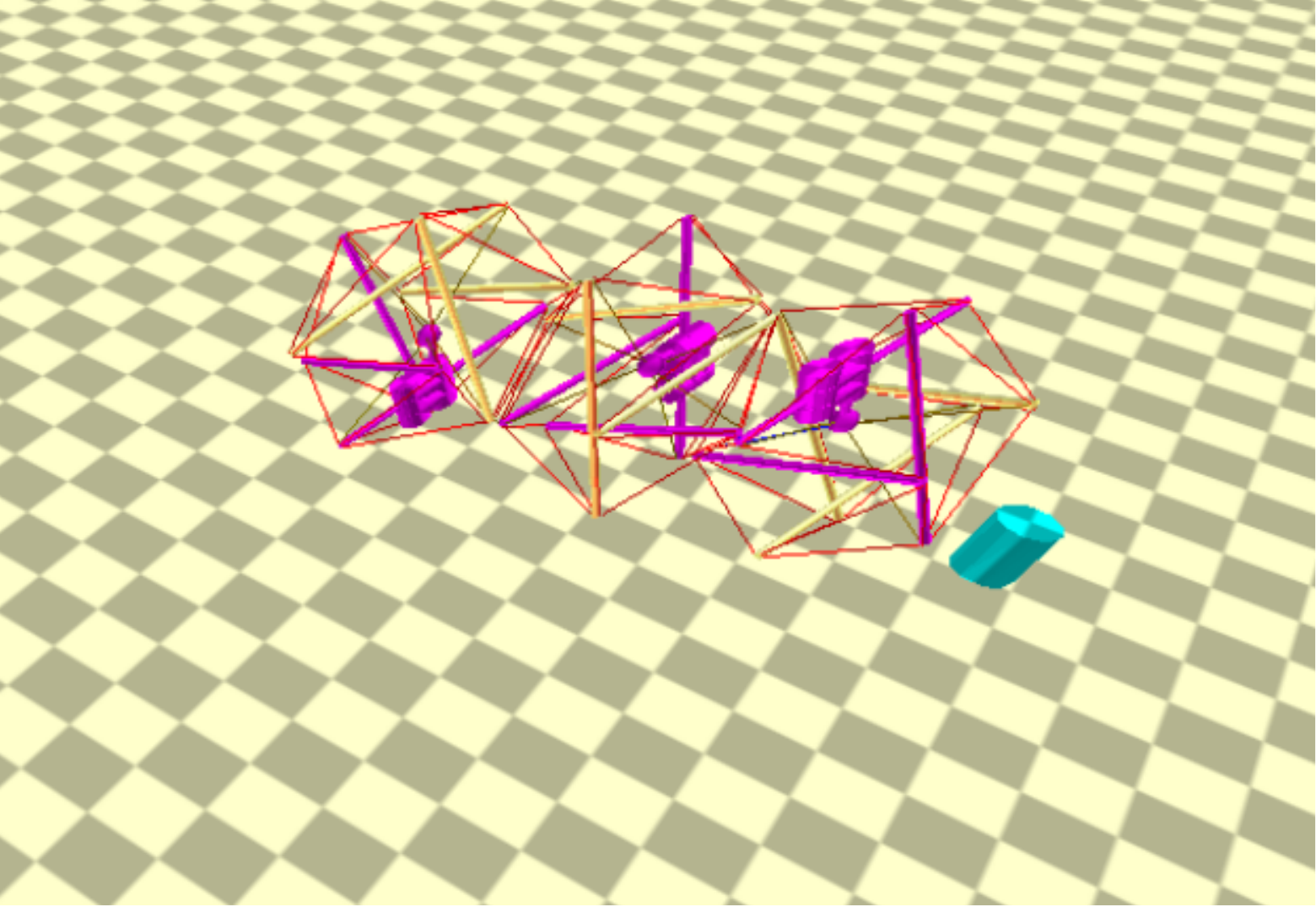}
      \label{fig:task-gr-end-zoom}
    \end{subfigure}
    \begin{subfigure}{0.495\linewidth}
      \centering
      \includegraphics[width=\linewidth]{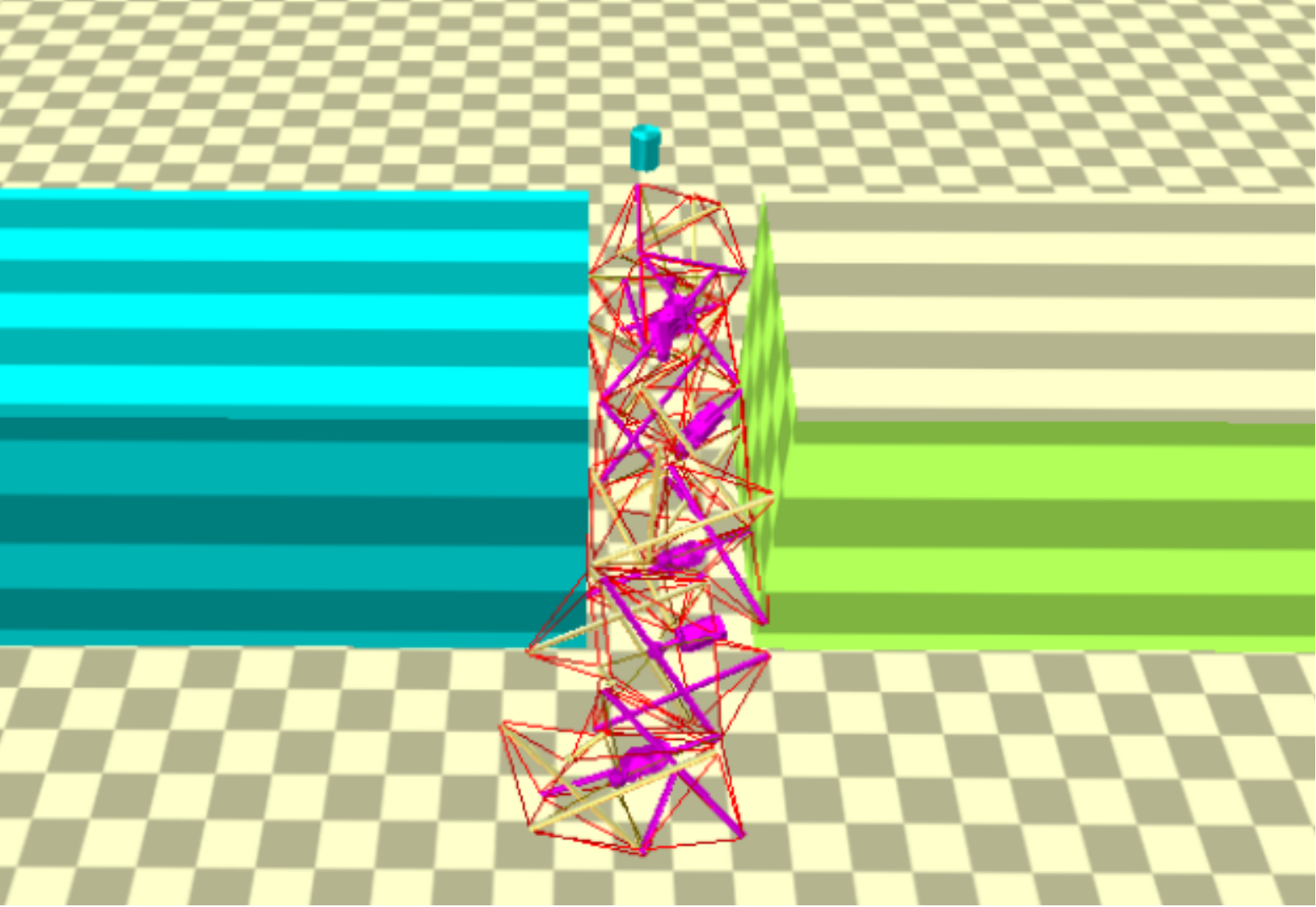}
      \label{fig:task-sgr-end-zoom}
    \end{subfigure}
    \vspace{-.5cm}
    \caption{Example of final situation for goal reaching (left) and squeezing (right). The cyan cylinder depicts the target.}
    \label{fig:tasks}
    \vspace{-.5cm}
\end{figure}

Goal reaching consists in having the robot moving towards a target placed somewhere in the environment. While this is a typical task in robotics research, modular soft structures have been barely taken into account in this regard. In practice, the objective consists in finding a controller able to drive the associated morphology towards different targets, based on the sensory information provided, i.e., distance and bearing to the target. In detail, distance and bearing are computed with respect to the position and the direction of the front face of the head module, respectively. An example of final situation of the task in Tensoft is shown in \cref{fig:tasks} (left).

\begin{table}[!t]
    \centering
    \caption{Main parameters for all the experiments.}
    \vspace{-8pt}
    \begin{tabular}{c|c|c}
                                         & Goal Reaching                 & Squeezing             \\ \hline
        \# Runs                          & 10                            & 10                    \\ \hline
        Run's budget                     & 45000 evaluations             & 45000 evaluations     \\ \hline
        \# Targets                       & 4                             & 2                     \\ \hline
        Distance                         & \SI{45}{cm}                   & \SI{60}{cm}           \\ \hline
        Bearing                          & 90° l, 45° l, 45° r, 90° r    & 5° l, 5° r            \\ \hline
        \# Simulation seeds              & 2                             & 2                     \\ \hline
        Simulation time                  & \SI{40}{s}                    & \SI{40}{s}            \\
    \end{tabular}
    \label{tab:exps-params}
\end{table}

\begin{table}[!t]
    \centering
    \caption{Parameter setting employed for ViE-NEAT.}
    \vspace{-8pt}
    \resizebox{\linewidth}{!}{
        \begin{subtable}[h]{0.52\linewidth}
            \centering
            \begin{tabular}{c|c}
                \multicolumn{2}{c}{ViE}                          \\ \hline
                Population size (init.)      & 48                \\ \hline
                \# Mutants                   & 48                \\ 
                \multicolumn{2}{c}{} \\
                \multicolumn{2}{c}{} \\
            \end{tabular}
        \end{subtable}
        \begin{subtable}[h]{0.52\linewidth}
            \centering
            \begin{tabular}{c|c}
                \multicolumn{2}{c}{NEAT}                \\ \hline
                Population size             & 54        \\ \hline
                Individual elitism          & 3         \\ \hline
                Species elitism             & 2         \\ \hline
                Compatibility threshold     & 2.85      \\ 
            \end{tabular}
        \end{subtable}
    }
    \label{tab:coev-exps}
\end{table}

\begin{table}[!t]
    \centering
    \caption{Parameter setting for MAP-Elites and DM-ME.}
    \vspace{-8pt}
    \begin{tabular}{c|c|c}
                                  & MAP-Elites                    & DM-ME                         \\ \hline
        \# Initial solutions      & 1080                          & 1080                          \\ \hline
        Batch size                & 24                            & 24                            \\ \hline
        Archive(s) size           & [9, 10, 9, 10]                & [9, 10] [9,10]                \\ \hline
        PCA updates               & \multicolumn{2}{c}{[0, 50, 150, 350, 750, 1550]}              \\ \hline
        Trajectory sampling       & \SI{1}{s}                     & \SI{1}{s}                     \\
    \end{tabular}
    \label{tab:map-elites-exps}
\end{table}

Squeezing consists in having the robot shrinking through a restricted aperture that is narrower than the robot size in terms of width and/or height. Due to the high degree of complexity, this task has not been addressed extensively in previous research. In this work, squeezing has been dealt with as an advanced form of goal reaching, by positioning the target beyond an aperture narrower than the maximum robot width. At the beginning of the simulation, the robot is enclosed by four walls, and it has to pass through the aperture to reach the target. In this case the sensory information provided includes distance and bearing to the target, distance and bearing with respect to the center of the entrance of the aperture (computed analogously to the ones referred to the target), and the presence of obstacles in front of the head module within a certain range (set to \SI{10}{cm} in our experiments). An example of final situation of the task in Tensoft is shown in \cref{fig:tasks} (right).

\subsection{Encoding}
\label{subsec:encoding}
The TSMR morphology encoding includes global properties, affecting the entire robot, and local ones, specific to each module. In detail, the global genes are the number of modules and the stiffness, which determines the degree of deformability of the modules. Instead, the local ones include only the number $\rho$, which identifies the triangular face that is connected to the next module in the chain. This influences not only the weight balance, but also how the robot moves since the servomotor acts always on the same pair of faces. The allowed mutations include: the addition of a new module in a random position; the deletion of a random module in the chain; the change of the robot stiffness; the change of the connection face to the next module (local mutation).

As concerns the controllers, the encoding and the allowed mutations are defined by NEAT. In particular, the TSMR controller has been implemented as a feed-forward NN that at each timestep of the simulation takes as input the sensory information provided for the considered task and produces as output the actuation parameters, i.e., the frequency $f$ and phase $\phi$ of a sinusoidal signal that controls contraction/expansion for all modules.

\subsection{Algorithm and task configurations}
\label{subsec:algo}
Three different approaches have been applied to both the goal reaching and squeezing tasks:
\begin{itemize}[leftmargin=*]
    \item co-evolution of morphology, evolved through ViE, and controller, evolved through NEAT (i.e., using ViE-NEAT);
    \item evolution of morphology-controller pairs through MAP-Elites;
    \item evolution of morphology-controller pairs using DM-ME.
\end{itemize}
The main parameters for all the experiments, common to the three approaches, are reported for each task in \cref{tab:exps-params}. For a fair comparison, the budget, limited by the computational resources available, has been defined for all algorithms in terms of total number of solution evaluations.

Each solution is evaluated against different targets placed at different orientations and, since the simulations are noisy, for each target multiple simulations are run, with different seeds for the actuation noise. 
All the simulations contribute to the fitness of the individual, to be minimized, which is computed for both tasks as:
\begin{gather}
    \nonumber \hspace{-15pt} \overline{d_t} = \frac{1}{N_s}\sum_{s=1}^{N_s} d_{ts} \quad t \in \{1,...\ ,N_t\} \\
    \begin{multlined}[][0.9\linewidth] f = \frac{1}{N_t}\sum_{t=1}^{N_t} \overline{d_t} + \frac{1}{2} \times (\max_{t=1 \dots N_t}^{} \overline{d_t} - \frac{1}{N_t}\sum_{t=1}^{N_t} \overline{d_t}) \end{multlined}
    \label{eq:fitness}
\end{gather}
where $N_s$ is the number of simulation seeds, $N_t$ is the number of targets, $d_{ts}$ is the final distance of the individual from the $t$-th target using the $s$-th seed, and $f$ is the resulting fitness.

In practice, first the average final distance across simulation seeds is calculated for each target. Then, the fitness of the individual is computed as the mean value of the average distances plus a penalty equal to half of the difference between the worst (maximum) and the mean average distance across targets. Actually, in the squeezing experiments $d_{ts}$ represents the final distance with any potential bonus already deducted: if the robot succeeds in crossing the entrance of the aperture by at least \SI{4}{cm}, a fixed bonus of \SI{4}{cm} is deducted from the final distance. In this way, the individuals capable of entering the aperture are favored by evolution.

As concerns the squeezing experiments, we set the wall's width to \SI{16.67}{cm}, its height to \SI{12}{cm}, and the aperture width to \SI{8}{cm}. In particular, the wall's width, which corresponds to the length of the aperture, is set to 1.5 times the length of a TSMR module. Instead, the width of the aperture is set to
72\% of a module's width (note that the width of a module is equal to its length, i.e., about \SI{11}{cm}).

A brief description of the approach-specific configurations, which are common to the two tasks, is provided in the following paragraphs. Indeed, only the number of controller input nodes is task-dependent (2 for goal reaching, 5 for squeezing, see \cref{subsec:tasks}).


\paragraph{ViE-NEAT}
The parameter configuration of the ViE-NEAT approach is shown in \cref{tab:coev-exps}. This configuration is characterized by elitism at both individual and species level, a one-to-one (on average) morphology-controller pairing for evaluation (a slightly larger NEAT population is required to compensate for the presence of elite individuals), and a relatively low threshold for speciation.

If an individual from one population is paired with two individuals from the other (due to the bigger size of the other population at play), the fitness of that individual is computed as:
\begin{equation}
    f = \frac{1}{N_p}\sum_{p=1}^{N_p} f_p + \frac{1}{2} \times (\max_{p=1 \dots N_p}^{} f_p - \frac{1}{N_p}\sum_{p=1}^{N_p} f_p)
    \label{eq:cumulative-fitness}
\end{equation}
where $N_p$ is the number of paired individuals belonging to the other population, $f_p$ is the fitness obtained with the $p$-th paired individual, according to Eq. \eqref{eq:fitness}, and $f$ is the resulting fitness. Since the controllers are not associated to a fixed morphology, the number of output nodes, which is morphology-dependent, is set to two times the maximum allowed number of modules, i.e., $20$ $(2 \times 10)$, and only the required number of outputs is considered at the evaluation time (the same holds for the MAP-Elites and DM-ME experiments).

\paragraph{MAP-Elites and DM-ME}
The configuration used for each of the two algorithms is reported in \cref{tab:map-elites-exps}; the only difference lies in the archive(s) size. The mutation of individuals (morphology-controller pairs) in MAP-Elites is performed as in DM-ME (see \cref{sec:dm-me}).

As regards the feature descriptors, the FD used for MAP-Elites corresponds to the concatenation of the FDs employed in the two archives of DM-ME, i.e., the morphology-related and the controller-related FDs (see the archives size in \cref{tab:map-elites-exps}). As a consequence, the archive filled by MAP-Elites has a higher number of cells than the ones explored by DM-ME (8100 cells vs 90 cells $\times$ 2 archives). Nevertheless, a comparison with equal number of cells would require using different FDs for the two algorithms, which would make the comparison unfair. Moreover, DM-ME has been specifically designed to reduce the total number of cells and thus to increase the selective pressure (with equal FDs) w.r.t. the original MAP-Elites.

In detail, the morphology-related FD is represented by the number of modules and the module stiffness; instead, the controller-related FD is provided by the PCA, exploiting the trajectory of the head module as sensory data. Since each morphology-controller pair is simulated with multiple targets, the sensory data vector are defined as the concatenation of the various trajectories: 320 elements in the case of goal reaching (4 targets $\times$ \SI{40}{s} $\times$ 2 coordinates), and 160 elements in the case of squeezing (2 targets instead of 4). In addition, since a solution is simulated multiple times with different seeds, the average trajectory across seeds is taken as representative for each target. Actually, the size of the controller-related FD has been chosen so that the number of cells is the same for the morphology-related and the controller-related feature space.



\section{Results}
\label{sec:results}
In the following, we present the results achieved in the goal reaching and squeezing experiments. We must remark that the fitness trends, the heatmaps and the statistics shown here have been computed after applying the following transformation to the fitness values obtained throughout the evolution:
\begin{equation}
    f_s = d_{init} - f
    \label{eq:res-diff-based-fit}
\end{equation}
where $d_{init}$ is the initial distance from the target (\SI{45}{cm} for goal reaching and \SI{60}{cm} for squeezing, see \cref{tab:exps-params}), $f$ is the fitness of the individual measured according to Eq. \eqref{eq:fitness} or \eqref{eq:cumulative-fitness} during the evolutionary process, and $f_s$ is the fitness used for plots and statistics computation. In this way, the minimization problem is presented as a maximization one, to facilitate the analysis in terms of QD.
\begin{figure}[!ht]
    \centering
    \vspace{-8pt}
    \includegraphics[width=.9\linewidth,trim=0 0 0 1cm,clip]{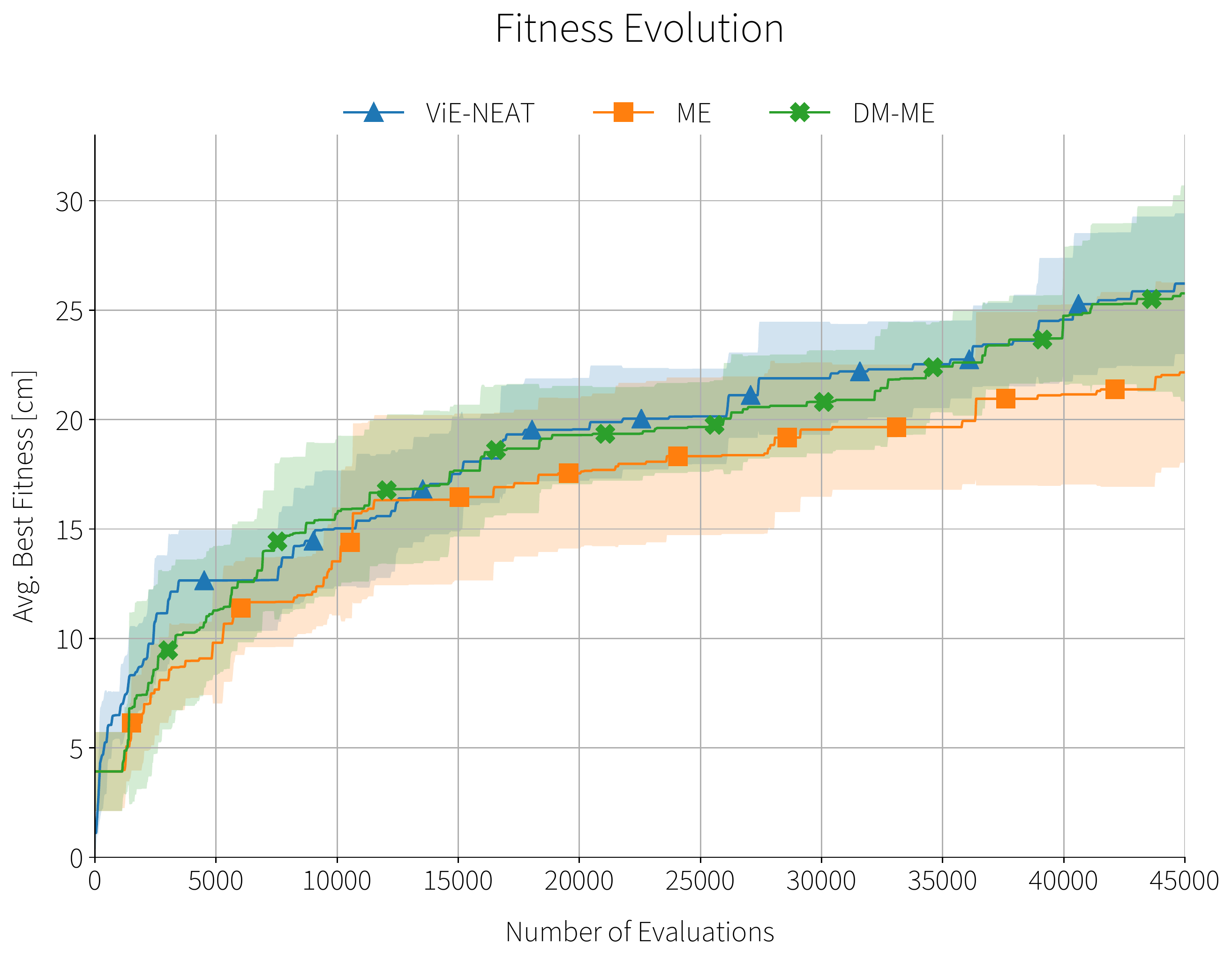}
    \vspace{-0.3cm}
    \caption{Best fitness (mean $\pm$ std. dev. across $10$ runs) over evaluations for the goal reaching task. In particular, the trends shown are related to: the morphology population for ViE-NEAT; the one and only archive for MAP-Elites (ME); the morphology archive for DM-ME. The fitness values (to be maximized) shown here have been obtained through Eq. \eqref{eq:res-diff-based-fit}.}
    \label{fig:fitness-ba-gr}
\end{figure}
\begin{figure}[!ht]
    \centering
    \vspace{-8pt}
    \includegraphics[width=.9\linewidth,trim=0 0 0 1cm,clip]{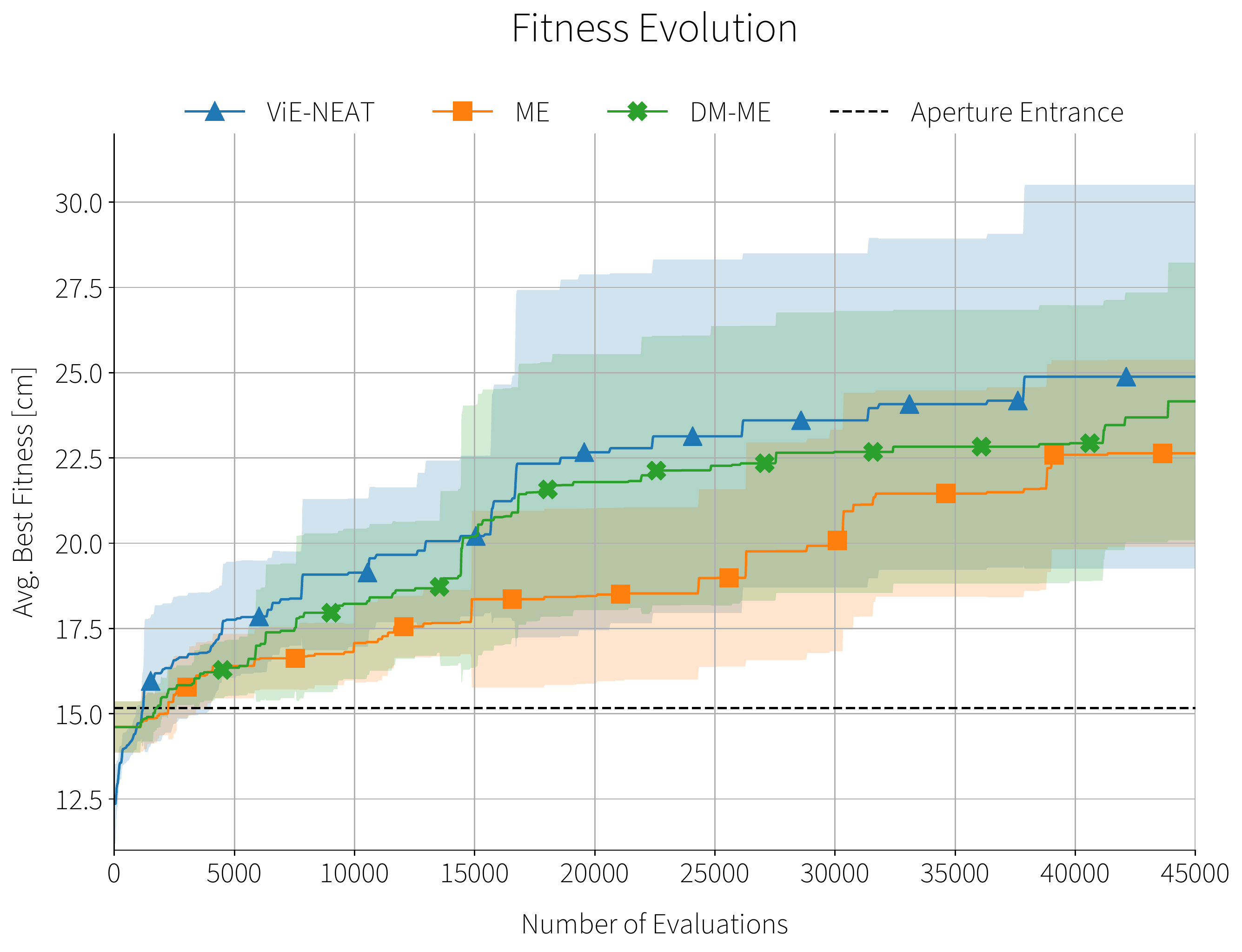}
    \vspace{-0.3cm}
    \caption{Best fitness (mean $\pm$ std. dev. across $10$ runs) over evaluations for the squeezing task. In particular, the trends shown are related to: the morphology population for VIE-NEAT; the one and only archive for MAP-Elites (ME); the morphology archive for DM-ME. The fitness values (to be maximized) shown here have been obtained through Eq. \eqref{eq:res-diff-based-fit}.}
    \label{fig:fitness-ba-sgr}
\end{figure}

All the experiments have been executed using a High Performance Computing (HPC) infrastructure. In detail, for each run, one node with 2 Intel Broadwell processors running at 2.6 GHz, with 14 cores each (28 cores in total), and 16/32 GB of RAM has been used.

\begin{table}[!ht]
    \centering
    \caption{Descriptive statistics of the best fitness across $10$ runs for the goal reaching task (up); Wilcoxon rank-sum test ($\alpha\,{=}\,0.05$) applied to the distributions of best fitness (bottom).}\label{tab:gr}
    \vspace{-8pt}
        \centering
        \begin{tabular}{c|c|c|c|c|c} 
            Algorithm         & Mean     & Std    & Min     & Max     & Median   \\ \hline
            ViE-NEAT          & 26.21    & 3.22   & 21.76   & 30.79   & 26.08    \\ \hline
            MAP-Elites        & 22.16    & 4.13   & 15.40   & 29.14   & 21.50    \\ \hline   
            DM-ME             & 25.76    & 4.93   & 20.66   & 39.36   & 24.22    \\
        \end{tabular}
        \begin{tabular}{c|c|c} 
            \multicolumn{2}{c|}{Algorithms}        & p-value         \\ \hline
            ViE-NEAT          & MAP-Elites         & 0.034293        \\ \hline
            ViE-NEAT          & DM-ME              & 0.545350        \\ \hline   
            MAP-Elites        & DM-ME              & 0.082099        \\
        \end{tabular}
\end{table}
\begin{table}[!ht]
    \centering
    \caption{Descriptive statistics of the best fitness across $10$ runs for the squeezing task (up); Wilcoxon rank-sum test ($\alpha\,{=}\,0.05$) applied to the distributions of best fitness (bottom).}\label{tab:sgr}
    \vspace{-8pt}
        \begin{tabular}{c|c|c|c|c|c} 
            Algorithm         & Mean     & Std    & Min     & Max     & Median   \\ \hline
            ViE-NEAT      & 24.88    & 5.63   & 18.81   & 35.94   & 22.75    \\ \hline
            MAP-Elites                & 22.63    & 2.74   & 18.60   & 26.05   & 22.63    \\ \hline   
            DM-ME             & 24.16    & 4.07   & 18.99   & 31.00   & 23.49    \\
        \end{tabular}
        \begin{tabular}{c|c|c} 
            \multicolumn{2}{c|}{Algorithms}        & p-value         \\ \hline
            ViE-NEAT          & MAP-Elites         & 0.496292        \\ \hline
            ViE-NEAT          & DM-ME              & 0.820596        \\ \hline   
            MAP-Elites        & DM-ME              & 0.496292        \\
        \end{tabular}
    \label{tab:sgr-fit}
\end{table}
\begin{table}[!ht]
    \centering
    \caption{Descriptive statistics of the QD score of the morphology archives generated in the goal reaching (up) and squeezing (bottom) experiments.}\label{tab:qd}
    \vspace{-8pt}
        \begin{tabular}{c|c|c|c|c|c} 
            Algorithm         & Mean      & Std     & Min      & Max      & Median    \\ \hline
            ViE-NEAT          & 535.61    & 59.21   & 461.81   & 641.67   & 524.60    \\ \hline
            MAP-Elites        & 552.27    & 35.41   & 519.24   & 642.73   & 544.26    \\ \hline   
            DM-ME             & 656.08    & 35.05   & 589.36   & 707.55   & 655.60    \\
        \end{tabular}
        \begin{tabular}{c|c|c|c|c|c} 
            Algorithm         & Mean       & Std     & Min       & Max       & Median     \\ \hline
            ViE-NEAT          & 1196.26    & 60.05   & 1099.57   & 1291.28   & 1203.39    \\ \hline
            MAP-Elites        & 1293.25    & 15.30   & 1264.05   & 1314.11   & 1296.67    \\ \hline   
            DM-ME             & 1322.52    & 14.60   & 1296.39   & 1341.12   & 1324.52    \\
        \end{tabular}
    \label{tab:qd-stats}
\end{table}
\begin{table}[!ht]
    \centering
    \caption{Wilcoxon rank-sum test ($\alpha\,{=}\,0.05$) applied to the QD score of the morphology archives generated in the goal reaching (GR) and squeezing (SQ) experiments.}\label{tab:qd-ranksum}
    \vspace{-8pt}
    \begin{tabular}{c|c|c|c} 
        \multicolumn{2}{c|}{Algorithms}        & p-value (GR)       & p-value (SQ)        \\ \hline
        ViE-NEAT          & MAP-Elites         & 0.325751           & 0.000881            \\ \hline
        ViE-NEAT          & DM-ME              & 0.000881           & 0.000157            \\ \hline   
        MAP-Elites        & DM-ME              & 0.000507           & 0.002497            \\
    \end{tabular}
\end{table}

\paragraph{Goal reaching}
\cref{fig:fitness-ba-gr} shows the best fitness trend (mean $\pm$ std. dev. across $10$ runs) for ViE-NEAT, MAP-Elites and DM-ME. In detail, ViE-NEAT and DM-ME exhibit an almost equivalent performance on average, whereas MAP-Elites turns out to be worse. The descriptive statistics of the best fitness across runs are reported in \cref{tab:gr} (up) for each algorithm. It can be seen that both ViE-NEAT and DM-ME obtain a higher mean best fitness value than MAP-Elites. It is also worth highlighting that the best individual across all the goal reaching experiments has been discovered by DM-ME\footnote{The videos of the best individuals are available at \url{https://tinyurl.com/ynaav7ek}.}. Both the equivalence of ViE-NEAT and DM-ME and the superiority of ViE-NEAT with respect to MAP-Elites are statistically significant, see the results of the Wilcoxon rank-sum test in \cref{tab:gr} (bottom). The table also shows that in this case DM-ME does not statistically outperform MAP-Elites. 

\begin{figure*}[!th]
    \centering
    \begin{subfigure}{0.495\textwidth}
      \centering
      \includegraphics[width=\textwidth]{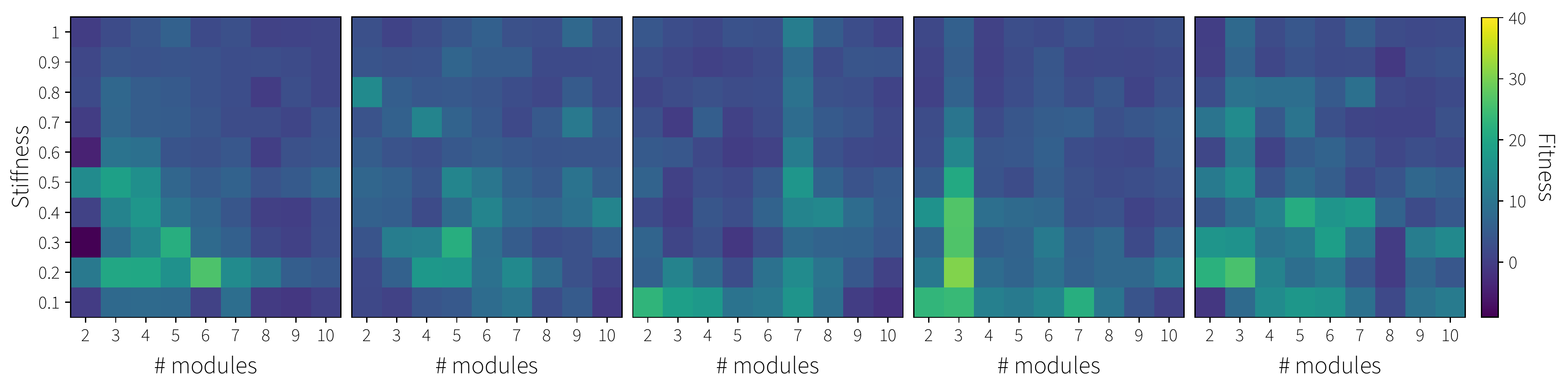}
      \caption{ViE-NEAT, morphology archive (projection), goal reaching}
      \label{fig:heatmaps-co-evo-gr}
    \end{subfigure}
    \begin{subfigure}{0.495\textwidth}
      \centering
      \includegraphics[width=\textwidth]{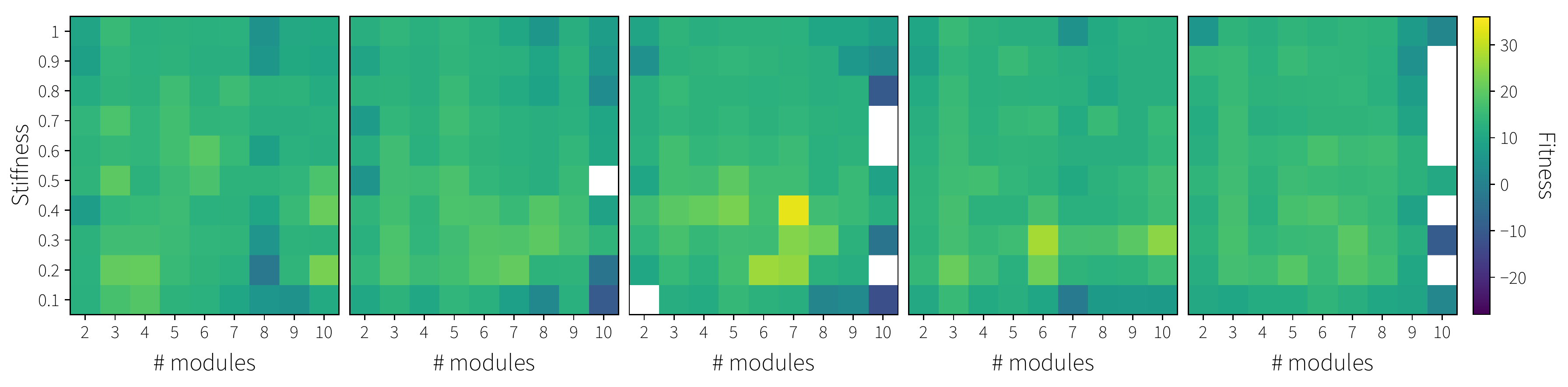}
      \caption{ViE-NEAT, morphology archive (projection), squeezing}
      \label{fig:heatmaps-co-evo-sgr}
    \end{subfigure} \\ \vspace{5pt}
    \begin{subfigure}{0.495\textwidth}
      \centering
      \includegraphics[width=\textwidth]{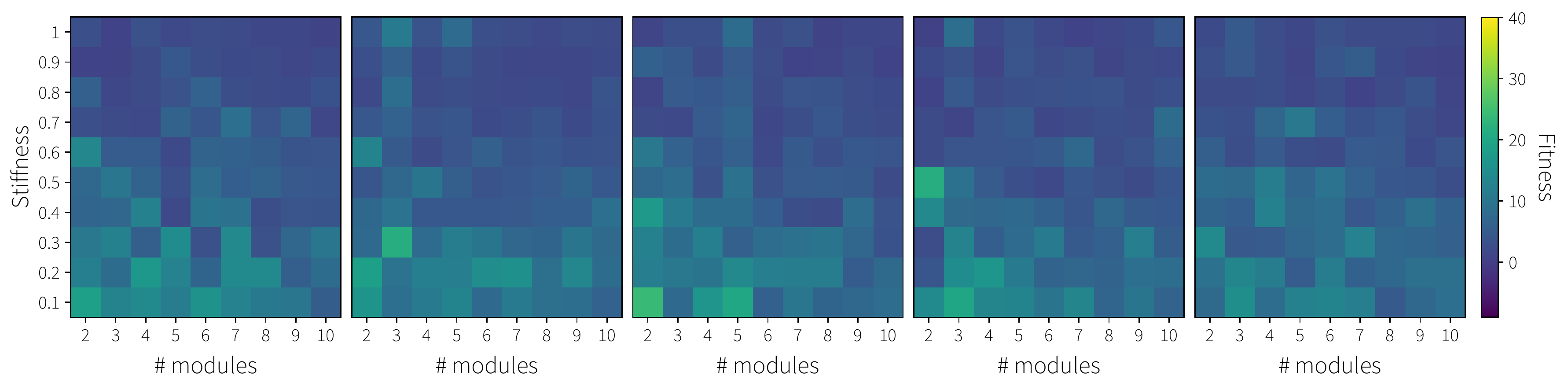}
      \caption{MAP-Elites, morphology archive (projection), goal reaching}
      \label{fig:heatmaps-me-morph-gr}
    \end{subfigure}
    \begin{subfigure}{0.495\textwidth}
      \centering
      \includegraphics[width=\textwidth]{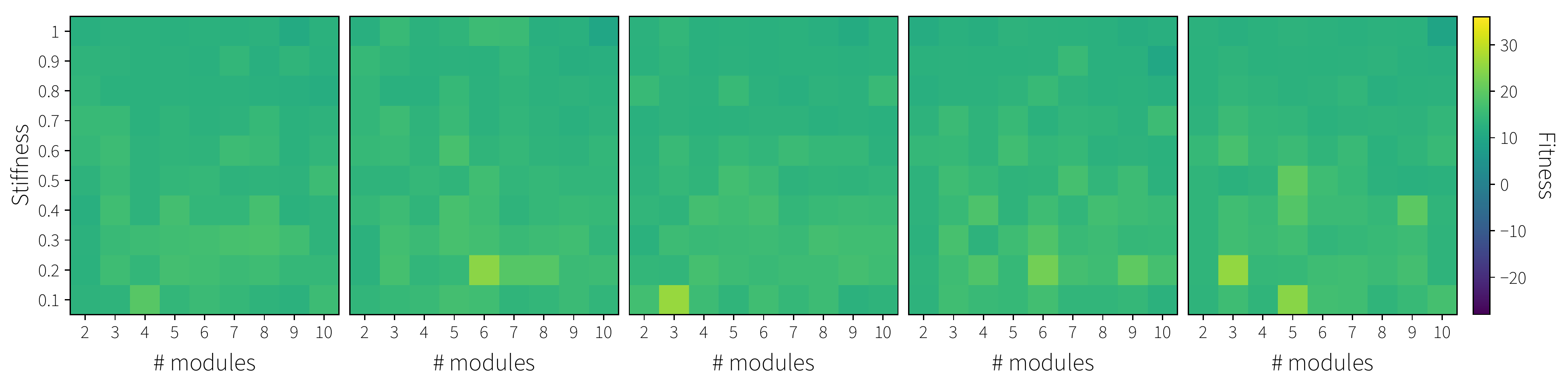}
      \caption{MAP-Elites, morphology archive (projection), squeezing}
      \label{fig:heatmaps-me-morph-sgr}
    \end{subfigure} \\ \vspace{5pt}
    \begin{subfigure}{0.495\textwidth}
      \centering
      \includegraphics[width=\textwidth]{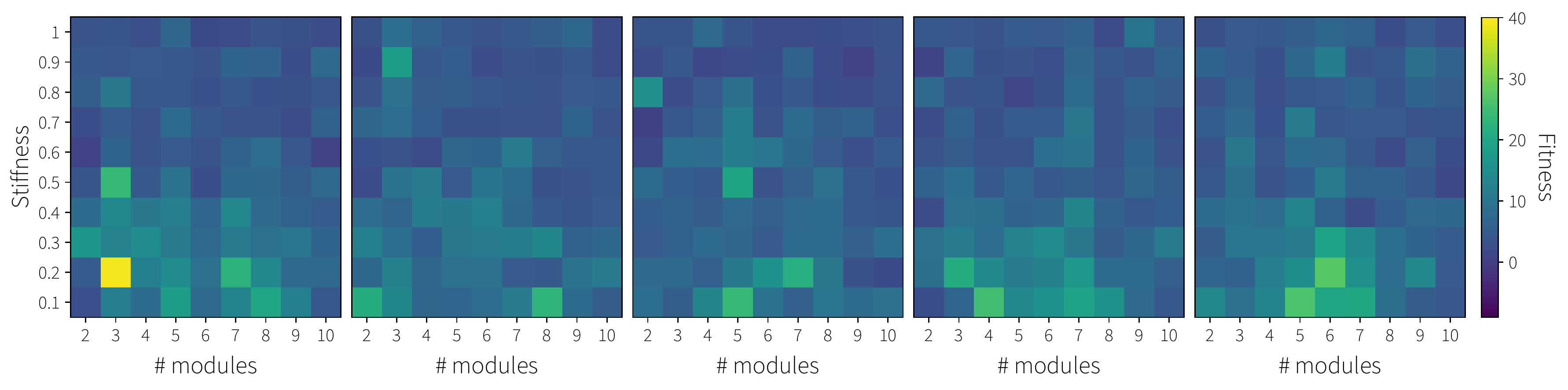}
      \caption{DM-ME, morphology archive, goal reaching}
      \label{fig:heatmaps-dm-me-morph-gr}
    \end{subfigure}
    \begin{subfigure}{0.495\textwidth}
      \centering
      \includegraphics[width=\textwidth]{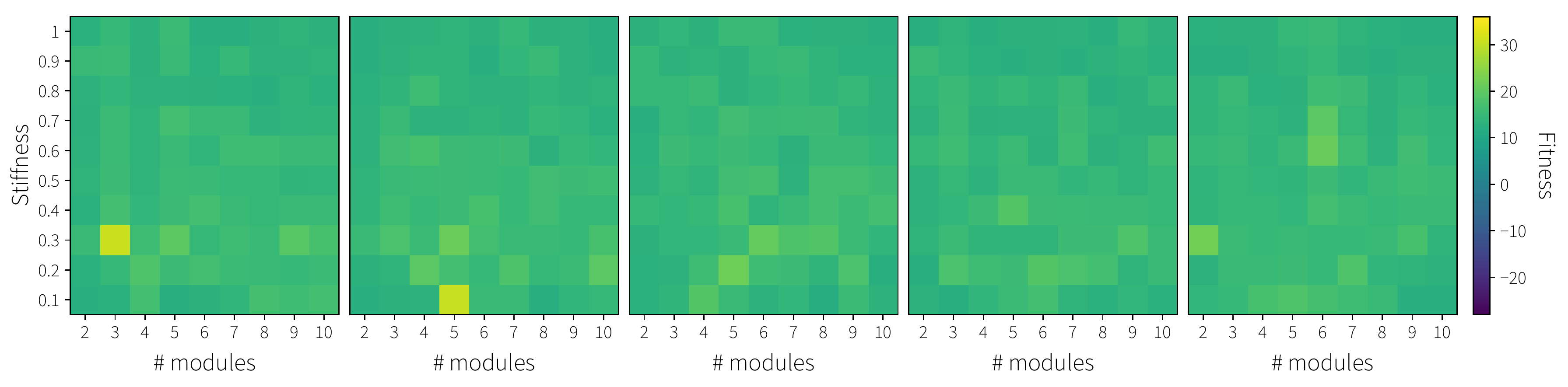}
      \caption{DM-ME, morphology archive, squeezing}
      \label{fig:heatmaps-dm-me-morph-sgr}
    \end{subfigure} \\ \vspace{5pt}
    \begin{subfigure}{0.495\textwidth}
      \centering
      \includegraphics[width=\textwidth]{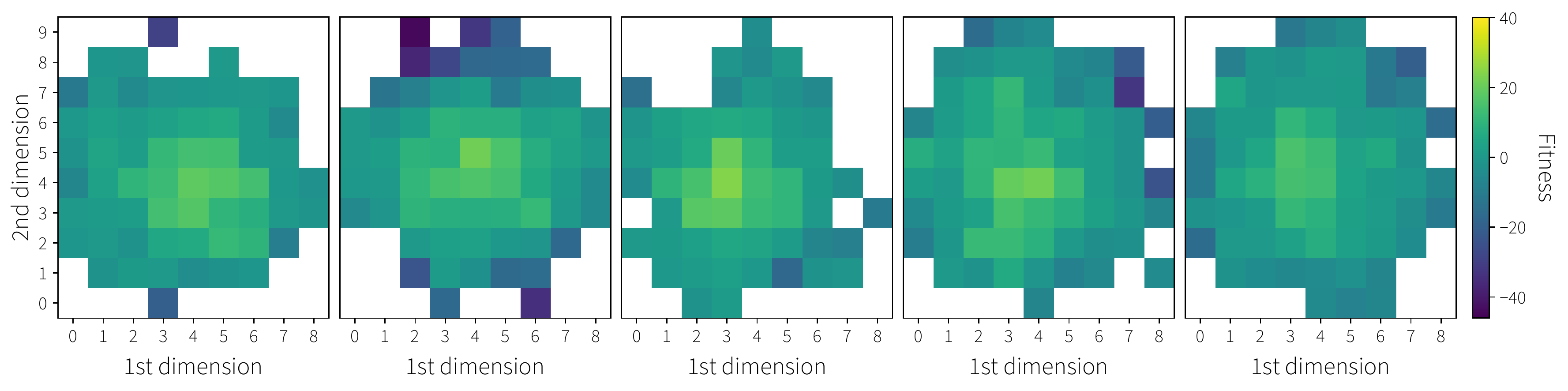}
      \caption{MAP-Elites, controller archive (projection), goal reaching}
      \label{fig:heatmaps-me-contr-gr}
    \end{subfigure}
    \begin{subfigure}{0.495\textwidth}
      \centering
      \includegraphics[width=\textwidth]{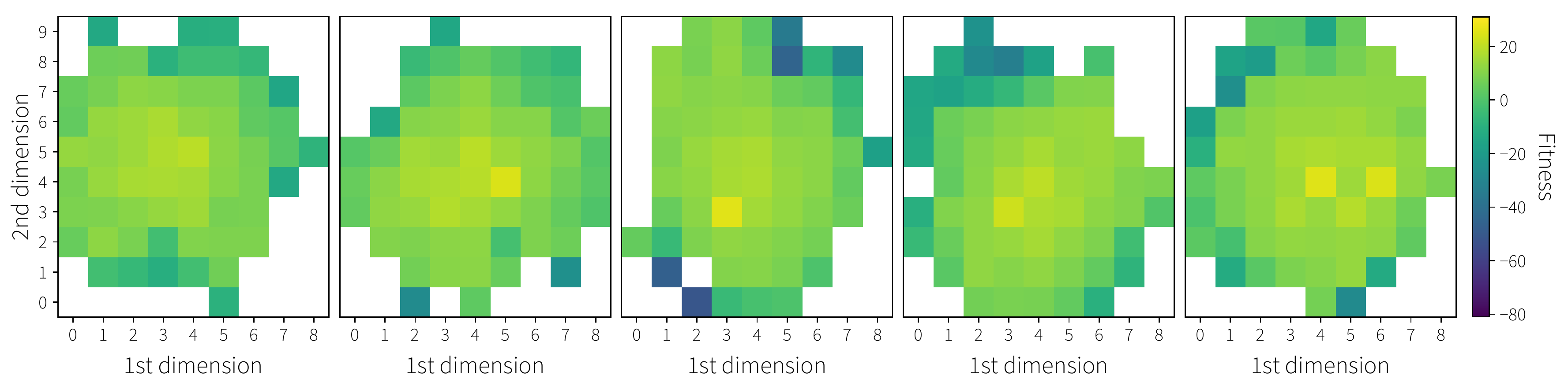}
      \caption{MAP-Elites, controller archive (projection), squeezing}
      \label{fig:heatmaps-me-contr-sgr}
    \end{subfigure} \\ \vspace{5pt}
    \begin{subfigure}{0.495\textwidth}
      \centering
      \includegraphics[width=\textwidth]{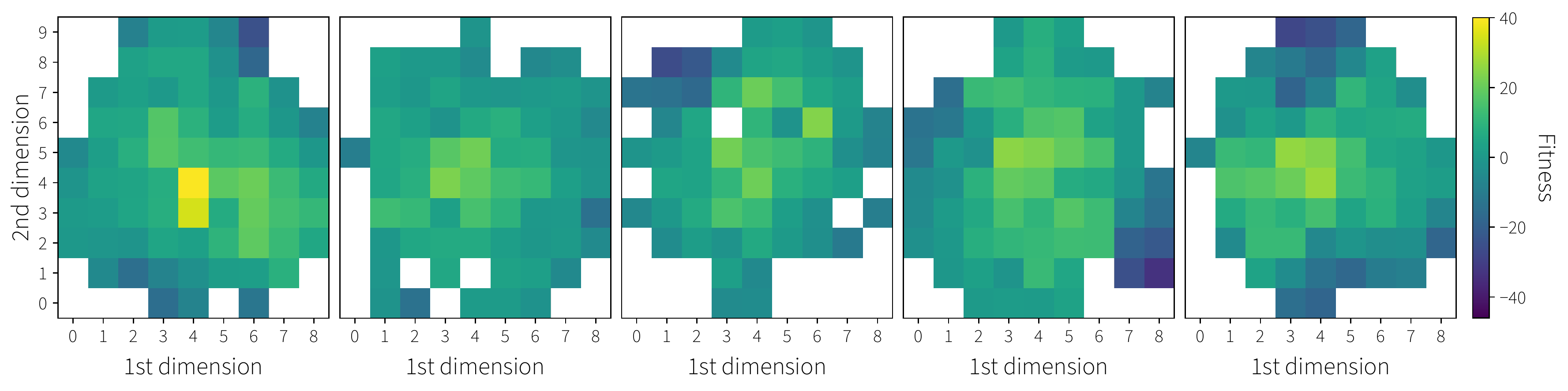}
      \caption{DM-ME, controller archive, goal reaching}
      \label{fig:heatmaps-dm-me-contr-gr}
    \end{subfigure}
    \begin{subfigure}{0.495\textwidth}
      \centering
      \includegraphics[width=\textwidth]{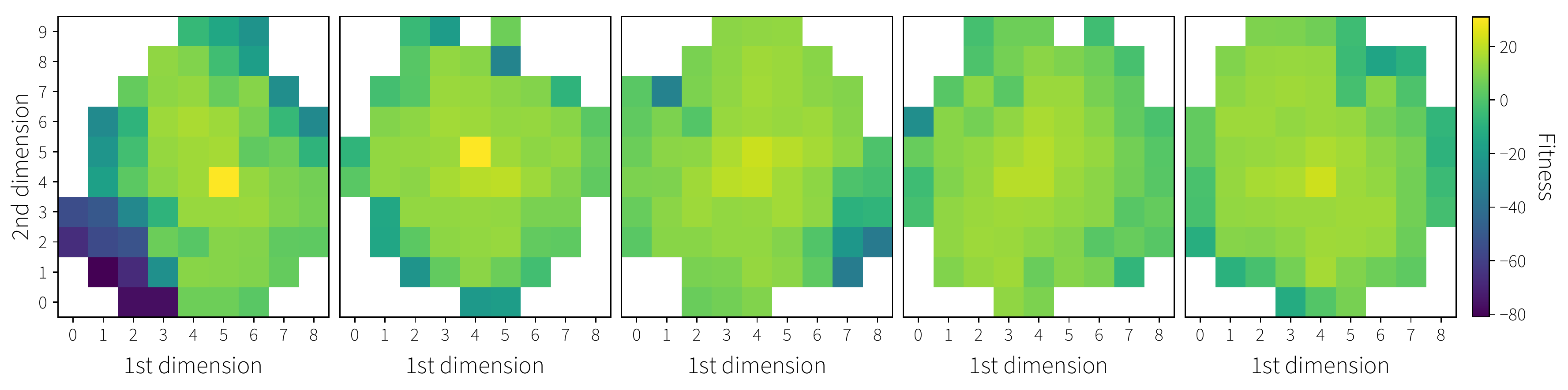}
      \caption{DM-ME, controller archive, squeezing}
      \label{fig:heatmaps-dm-me-contr-sgr}
    \end{subfigure}
    \caption{Archives generated by five runs of each experiment configuration. The first row shows the archives generated by ViE-NEAT, obtained by projecting on the morphology feature space the individuals discovered throughout the evolution (the reconstruction of the controller archives would require modifying the algorithm since ViE-NEAT does not include the PCA). The second and third row show the morphology-related archives generated by MAP-Elites (i.e., the projection from the resulting single archive) and DM-ME, respectively. The fourth and fifth row display the corresponding controller-related archives. In all cases, the fitness values (to be maximized) shown here have been obtained through Eq. \eqref{eq:res-diff-based-fit}.}
    \label{fig:heatmaps}
\end{figure*}

The other evaluation metric we are interested in evaluating here is the \textit{illumination} of the feature space(s). With reference to \cref{fig:heatmaps} (left column, first three rows), which provides a heatmap representation of the archives, or archive projections, generated in five runs of the goal reaching experiments, it can be noted that DM-ME appears to perform better in terms of morphology archives produced (although the difference is not so pronounced), whereas ViE-NEAT shows almost equivalent performance to MAP-Elites. As regards the controller archives, it is worth remembering that each cell corresponds to a different trajectory according to the PCA and, since not all trajectories can lead to good results, it is reasonable to find also bad performing individuals inside them. In addition, the coverage of the controller archives is influenced by the re-discretization of the features space.
Hence, it is reasonable to focus on the morphology archives. The difference in terms of illumination capability between the three algorithms is confirmed by the statistics computed on the QD score (i.e, the sum of the fitness of the individuals in the archive \cite{qd_score}) of the morphology archives. All the ten runs per algorithm have been taken into account for the computation of these statistics, see \cref{tab:qd} (up). Both the superiority of DM-ME on the other two methods and the equivalence between ViE-NEAT and MAP-Elites in terms of QD score are statistically significant, see the results of the Wilcoxon rank-sum test in \cref{tab:qd-ranksum}.

An inverse analysis has been also performed on MAP-Elites and DM-ME. The double archives generated by DM-ME have been projected onto single ones, and then compared with the single archives produced by MAP-Elites. This requires to re-simulate the individuals contained in the morphology archives of DM-ME in order to collect their trajectories and compute the controller-related FDs. 
Specifically, the PCA and the cell boundaries used for the projection (i.e., the final ones) are potentially different from the ones used for the evaluation of the considered individuals during the evolution. Hence, if the reconstructed single archives were projected back onto separate morphology-controller ones, the controller archives would be potentially different from the original ones. The results of the double-to-single projection show that most of the individuals contained in the controller archives generated by DM-ME tend to be concentrated in a small number of cells in the morphological feature space, typically with a medium-high number of modules and very low stiffness. This is reasonable, since those morphological properties allow a greater variety of mid-performing behaviors. Instead, as expected, the single archives generated by MAP-Elites show a higher coverage of the 4-dimensional feature space, especially in the region of medium-low stiffness. The heatmap representation of the single archives can be found in the Supplementary Material.

Concerning the execution time, MAP-Elites ($154.3\pm9.22$ hours) and DM-ME ($145.4\pm10.76$ hours) turn out to be considerably more expensive than ViE-NEAT ($94.3\pm13.15$ hours). It is worth mentioning however that both MAP-Elites and DM-ME present an overhead due to the need for collecting the sensory data (trajectories), which is written by the simulator into temporary files. 

\paragraph{Squeezing}
\cref{fig:fitness-ba-sgr} shows the best fitness trend (mean $\pm$ std. dev. across $10$ runs) for the three algorithms. The black dashed line represents the fitness limit corresponding to a robot entering the aperture in a single simulation. By looking at the plot, it turns out that the three algorithms discover well-performing solutions. In this case, the best individual has been discovered by ViE-NEAT, and again both ViE-NEAT and DM-ME obtain a higher mean best fitness value than MAP-Elites, see the descriptive statistics reported in \cref{tab:sgr} (up). On the other hand, the three algorithms result statistically equivalent, see the results of the Wilcoxon rank-sum test provided in \cref{tab:sgr} (bottom). Nevertheless, by looking at the behavior of the best individuals\footnote{The videos of the best individuals are available at \url{https://tinyurl.com/y3e9neej}.}, it turns out that the best individual evolved by DM-ME is the only one, among the best individuals obtained by all algorithms, that is able to pass through the aperture and reach the target if the simulation is allowed to continue beyond the \SI{40}{s} used in the evolutionary process. This also demonstrates that the task is achievable.

As concerns the illumination of the search space, \cref{fig:heatmaps} (right column) provides a heatmap representation of the archives, or archive projections, generated in five runs of the squeezing experiments. As in the case of goal reaching, we focus the analysis on the morphology archives. In this case, some runs of ViE-NEAT are unable to fill all the cells of the grid. Except for that, the archives generated appear to be very similar, especially the ones produced by MAP-Elites and DM-ME. Nevertheless, by looking at the statistics of the QD score of the archives generated in the ten runs, reported in \cref{tab:qd} (bottom), it turns out that DM-ME outperforms MAP-Elites, which in turn outperforms ViE-NEAT. \cref{tab:qd-ranksum} provides the statistical evidence based on the Wilcoxon rank-sum test.

The heatmap representation of the single archives, the ones generated by MAP-Elites and the ones resulting from the projection of the double archives produced by DM-ME, can be found in the Supplementary Material. The observations made for goal reaching, as concerns this inverse analysis, hold also for squeezing.

As in goal reaching, the execution time of MAP-Elites ($103.3\pm5.87$ hours) and DM-ME ($94.9\pm12.23$ hours) turns out to be higher than that of ViE-NEAT ($70.9\pm6.07$ hours). The lower execution time with respect to goal reaching is mainly due to the number of targets considered for each individual, which in this case is 2 instead of 4 (see \cref{tab:exps-params}). However, a single simulation of the squeezing task tends to require a little more time than one of the goal reaching task, due to the interactions between the robot and the walls.


\section{Conclusions}
\label{sec:conclusion}
In this work we have addressed the joint optimization of morphology and controller of TSMRs. In detail, we have considered three different evolutionary approaches, i.e., MAP-Elites, ViE-NEAT, and DM-ME, with the last two being algorithms proposed here for co-evolving morphologies and controllers. In order to compare the three algorithms, we have conducted an experimental campaign on two robotic tasks: goal reaching and squeezing. As concerns goal reaching, ViE-NEAT outperforms MAP-Elites and results equivalent to DM-ME in terms of best fitness. As regards squeezing, the three algorithms achieve similar results in terms of quality of discovered solutions. Moreover, DM-ME outperforms the other two approaches in terms of illumination\slash exploration of the feature space (measured with QD score) in both tasks. The higher total fitness of the resulting archive implies a higher number of well-performing candidates for the physical realization. Notably, ViE-NEAT achieves very similar performance to MAP-Elites in terms of illumination of the search space in the goal reaching task, although it does not exploit any map. Future work includes testing other configurations of DM-ME and co-evolution, other robotic tasks, and more complex TSMR structures.


\begin{acks}
We thank Daniele Bissoli for implementing the original framework on which this work was built.
\end{acks}

\clearpage

\balance

\bibliographystyle{ACM-Reference-Format}
\bibliography{main_arxiv}


\begin{figure*}[h]
\section*{Supplementary Material}
    \centering
    
    \vspace{10pt} \centerline{\textbf{MAP-Elites}} \vspace{10pt}
    \begin{subfigure}{0.195\textwidth}
      \centering
      \includegraphics[width=\textwidth]{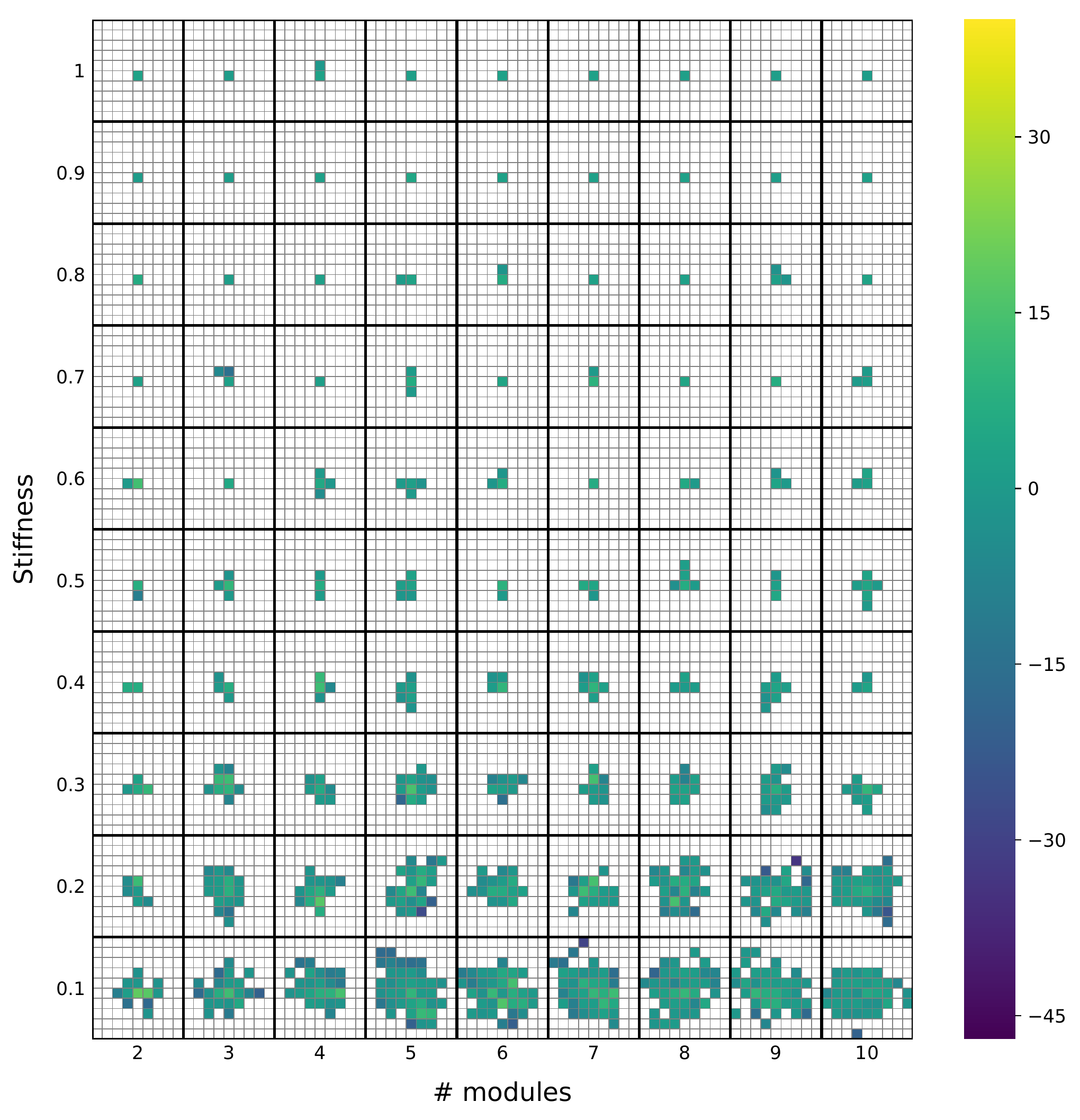}
      \label{fig:s-heat-gr-me-0}
    \end{subfigure} 
    \begin{subfigure}{0.195\textwidth}
      \centering
      \includegraphics[width=\textwidth]{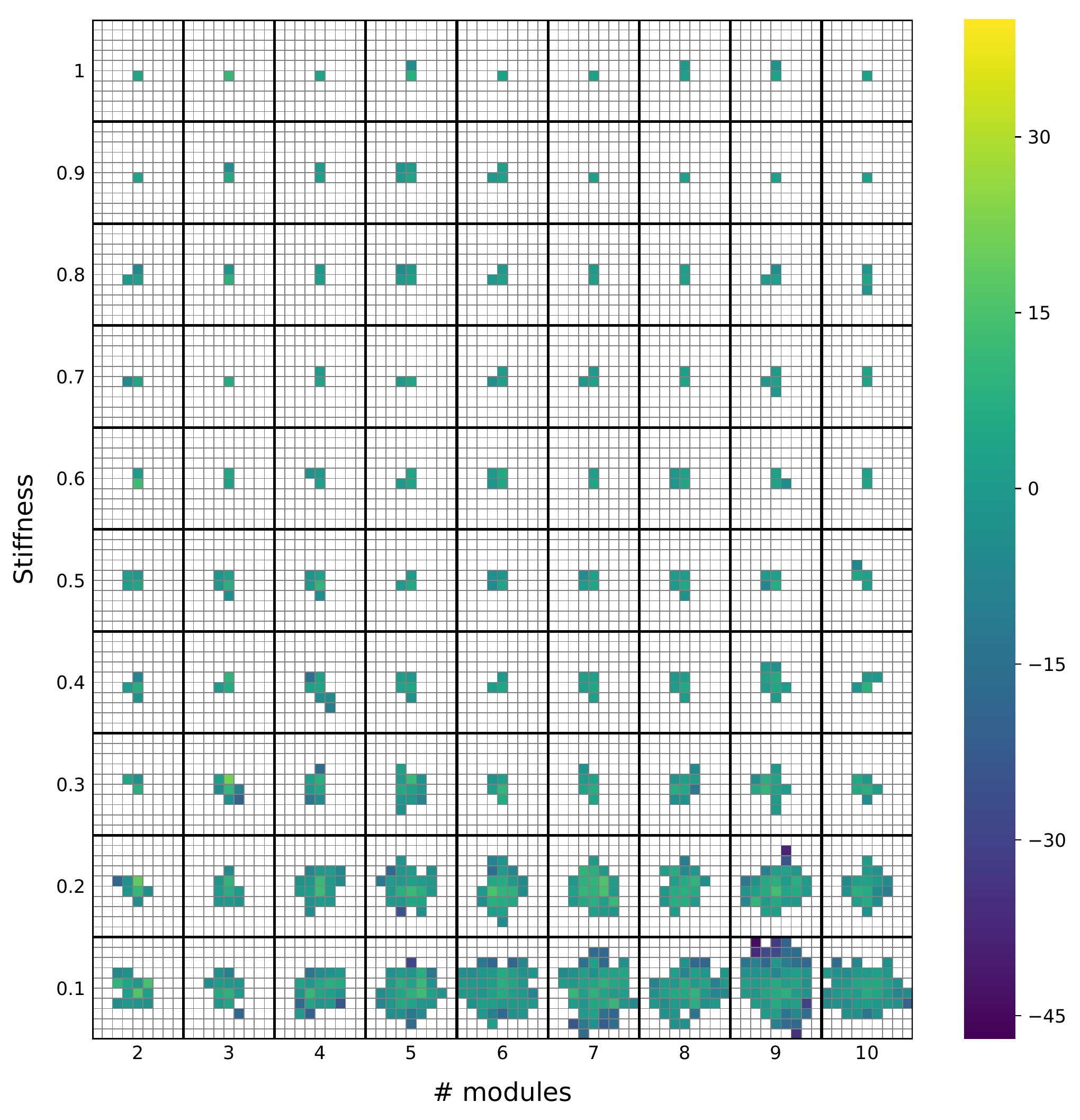}
      \label{fig:s-heat-gr-me-1}
    \end{subfigure} 
    \begin{subfigure}{0.195\textwidth}
      \centering
      \includegraphics[width=\textwidth]{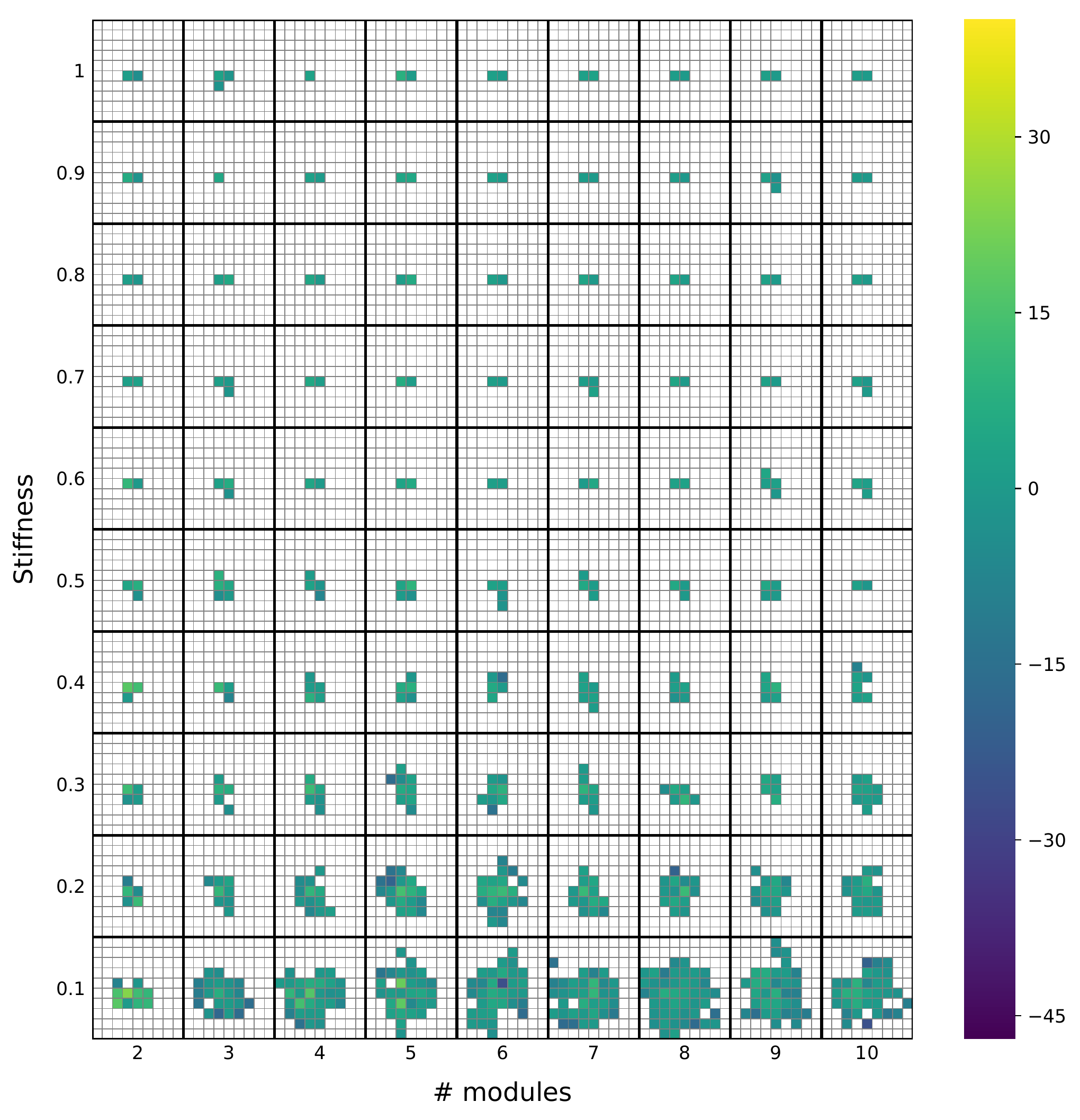}
      \label{fig:s-heat-gr-me-2}
    \end{subfigure} 
    \begin{subfigure}{0.195\textwidth}
      \centering
      \includegraphics[width=\textwidth]{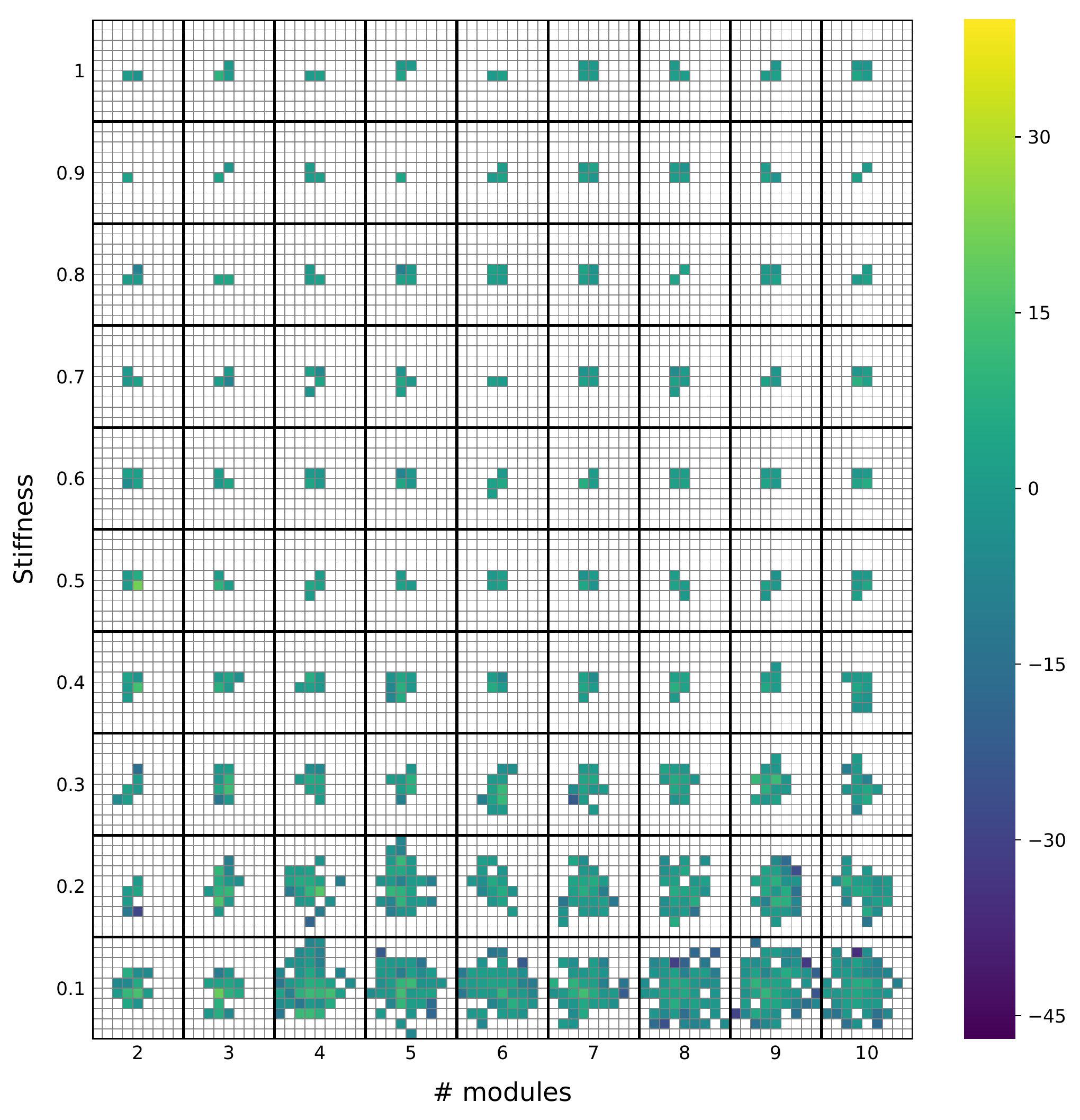}
      \label{fig:s-heat-gr-me-3}
    \end{subfigure} 
    \begin{subfigure}{0.195\textwidth}
      \centering
      \includegraphics[width=\textwidth]{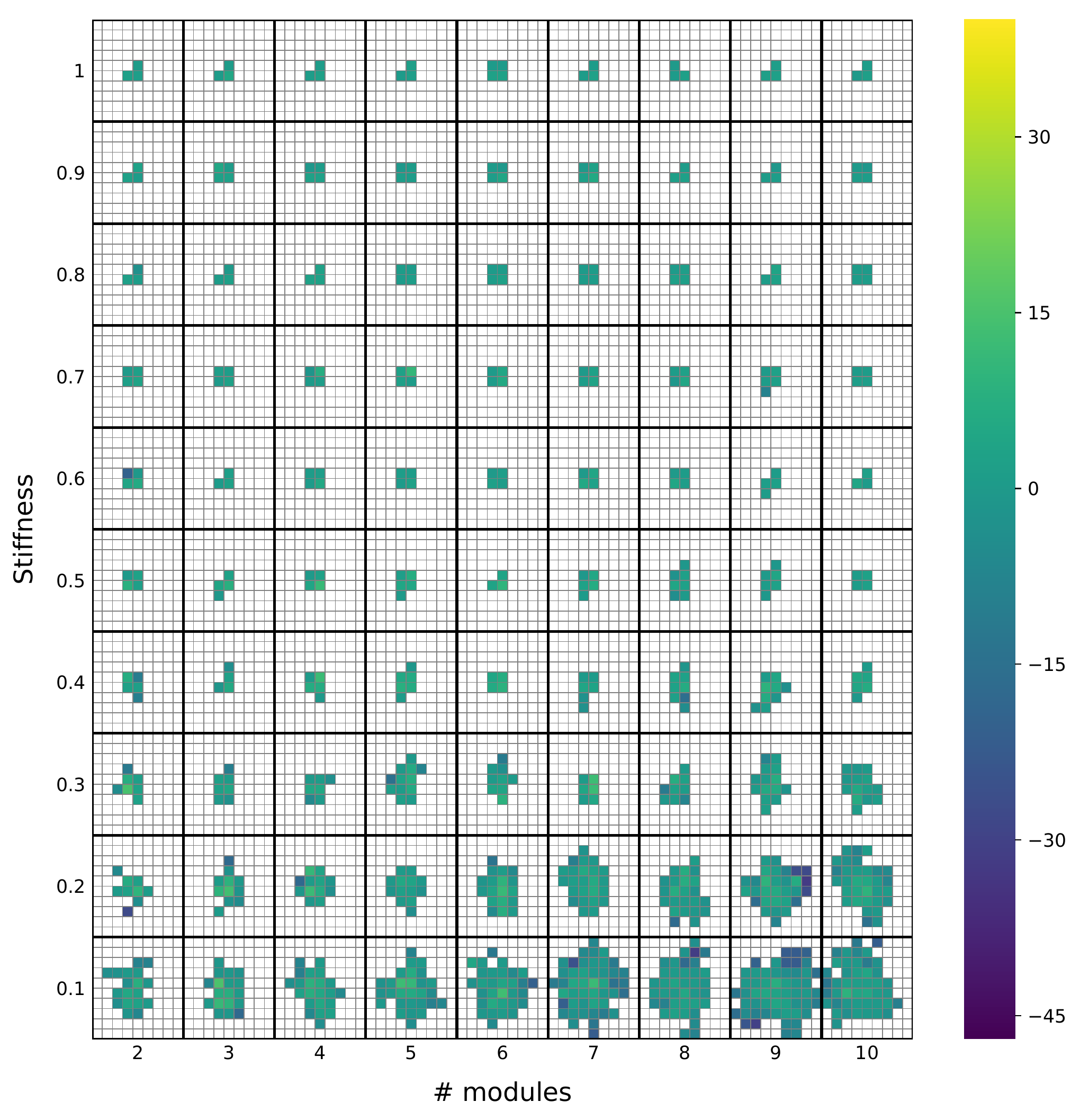}
      \label{fig:s-heat-gr-me-4}
    \end{subfigure} \\ \vspace{-10pt}
    \begin{subfigure}{0.195\textwidth}
      \centering
      \includegraphics[width=\textwidth]{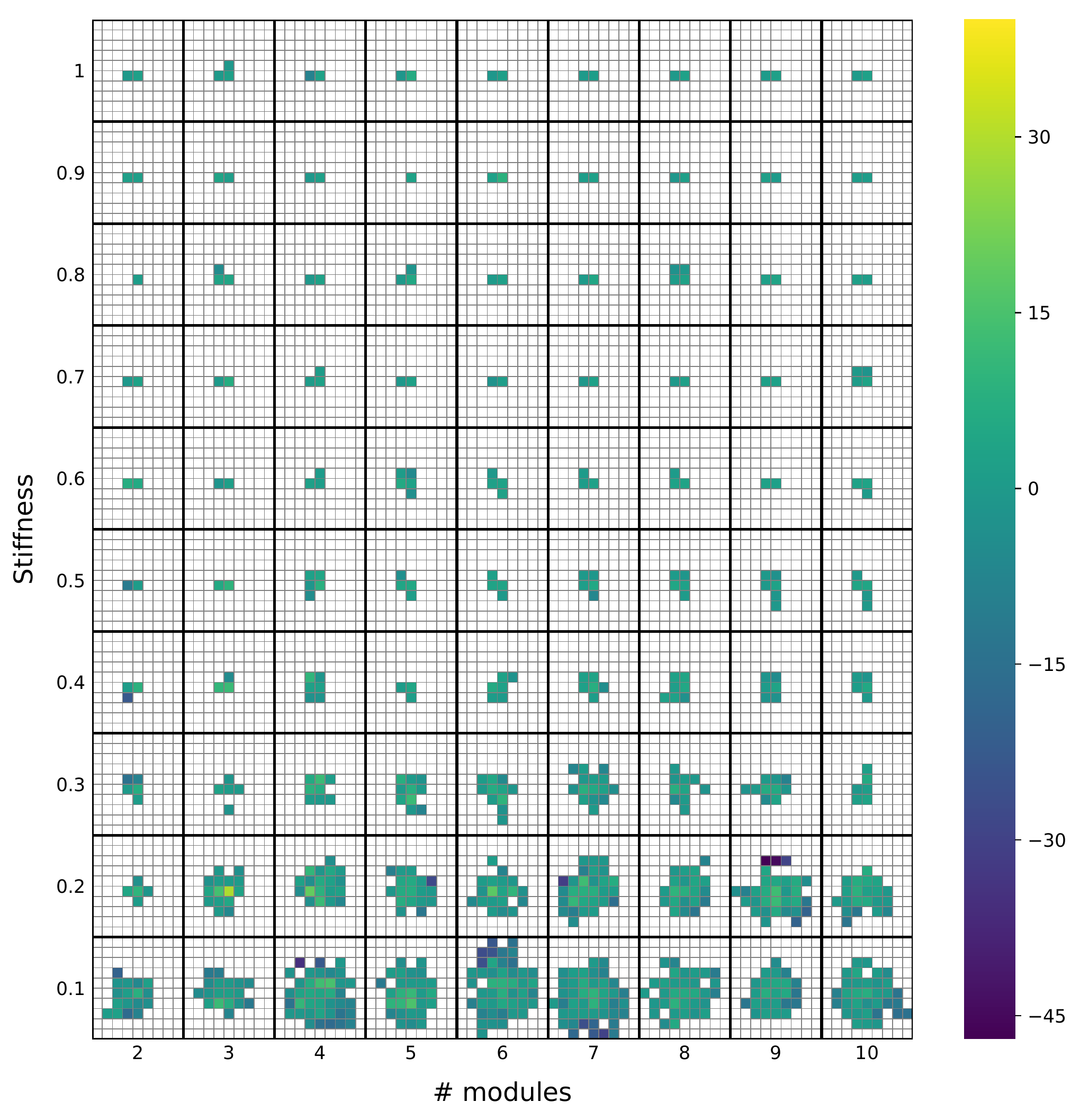}
      \label{fig:s-heat-gr-me-5}
    \end{subfigure} 
    \begin{subfigure}{0.195\textwidth}
      \centering
      \includegraphics[width=\textwidth]{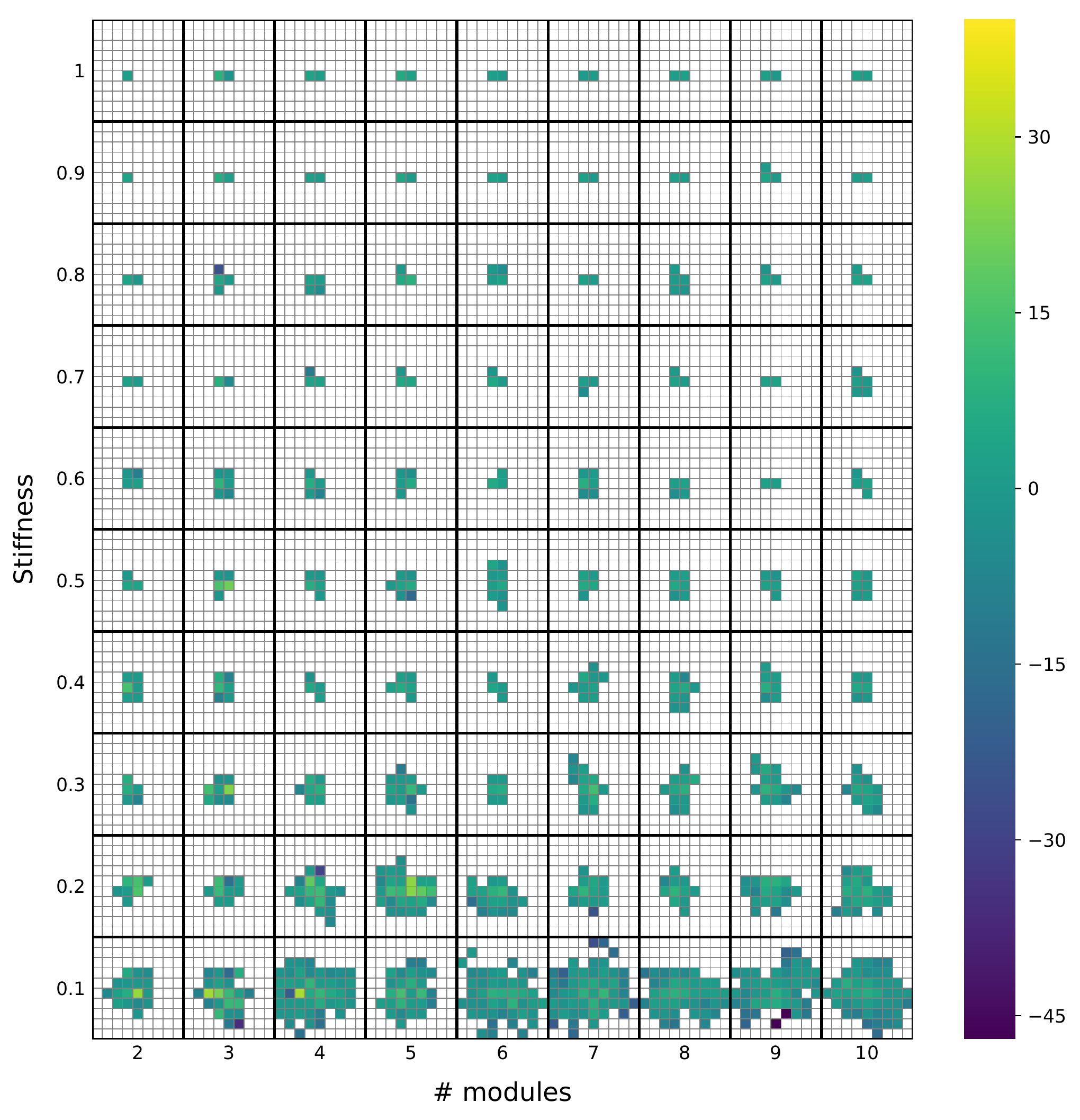}
      \label{fig:s-heat-gr-me-6}
    \end{subfigure} 
    \begin{subfigure}{0.195\textwidth}
      \centering
      \includegraphics[width=\textwidth]{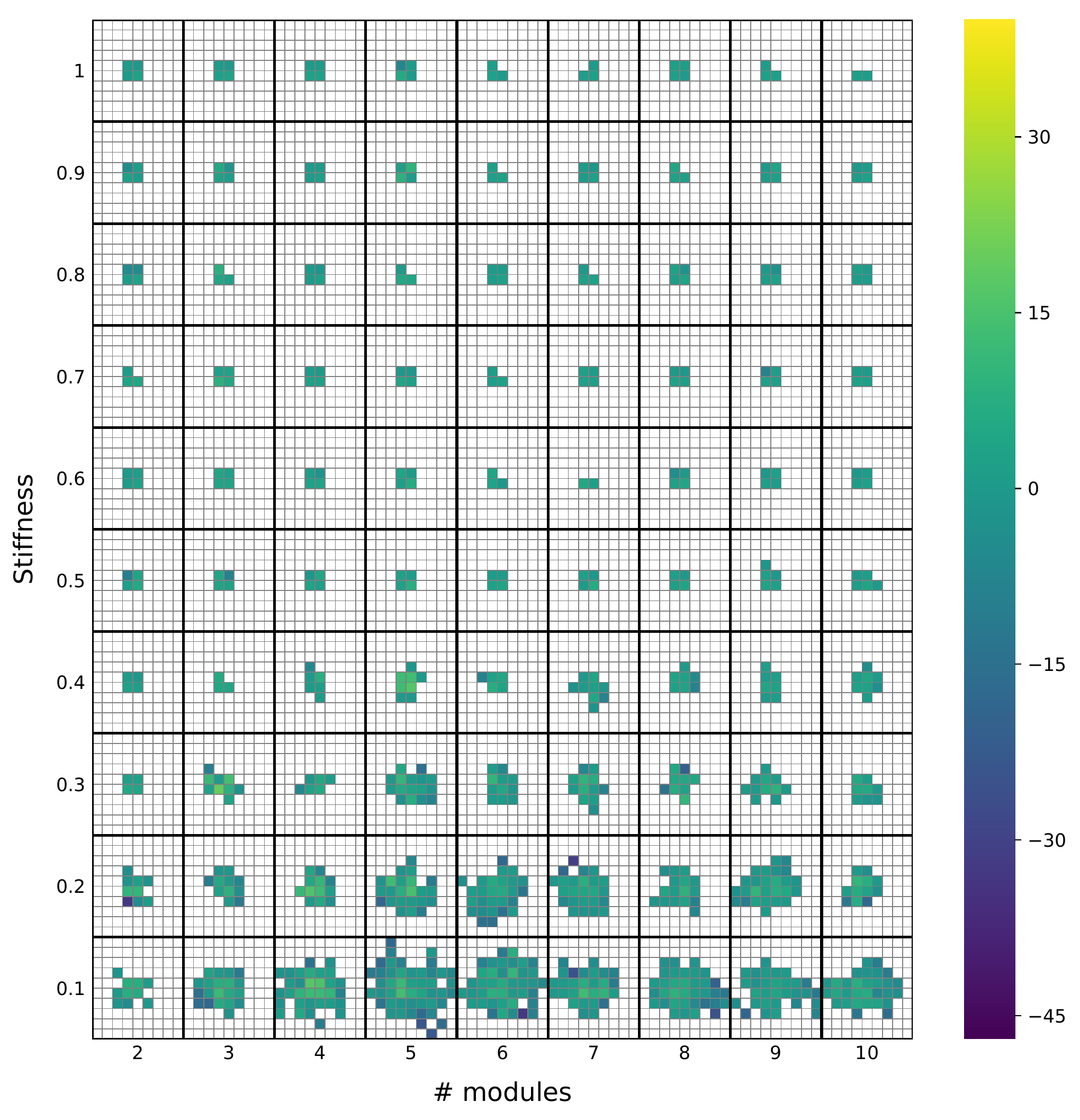}
      \label{fig:s-heat-gr-me-7}
    \end{subfigure} 
    \begin{subfigure}{0.195\textwidth}
      \centering
      \includegraphics[width=\textwidth]{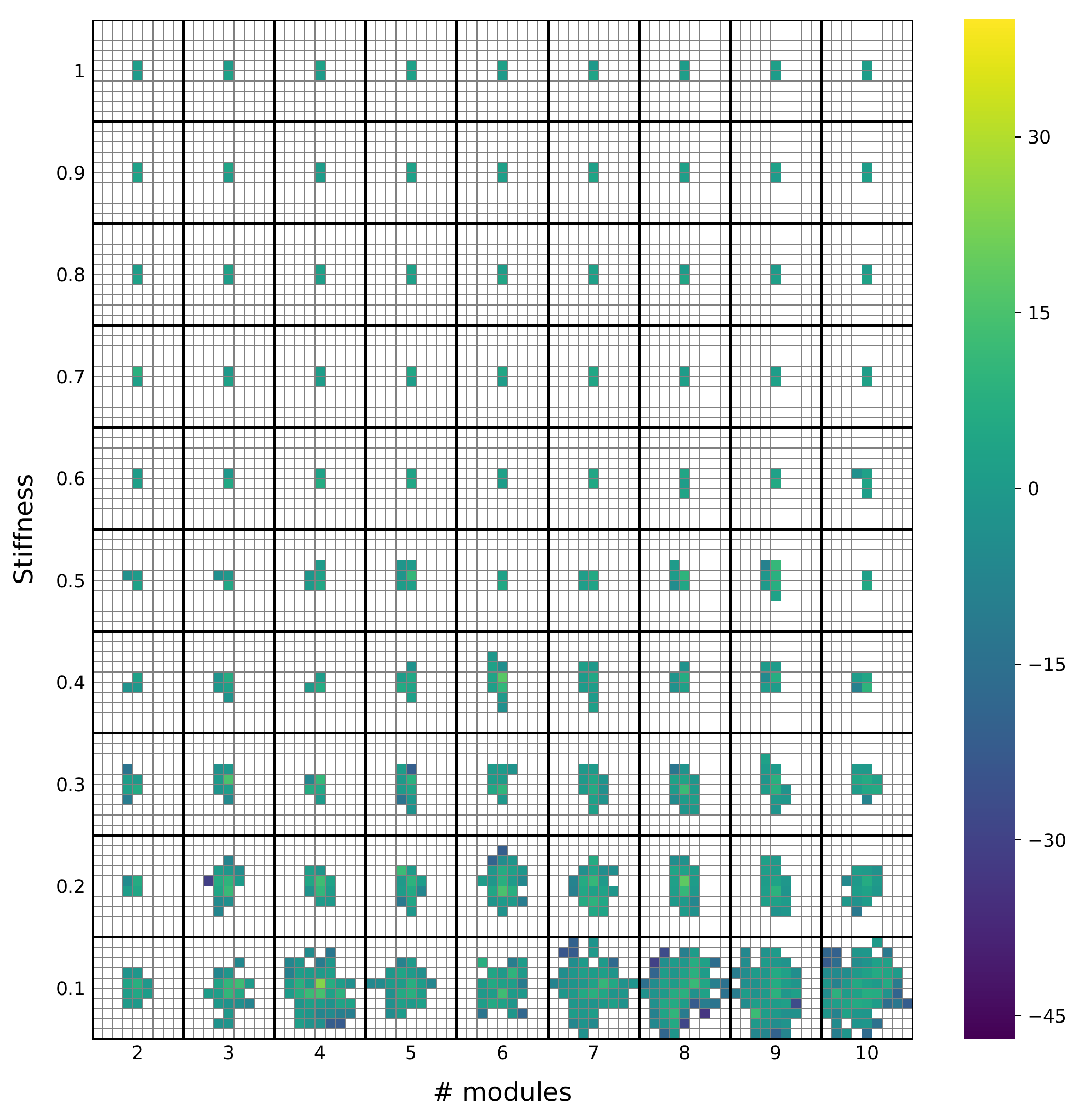}
      \label{fig:s-heat-gr-me-8}
    \end{subfigure} 
    \begin{subfigure}{0.195\textwidth}
      \centering
      \includegraphics[width=\textwidth]{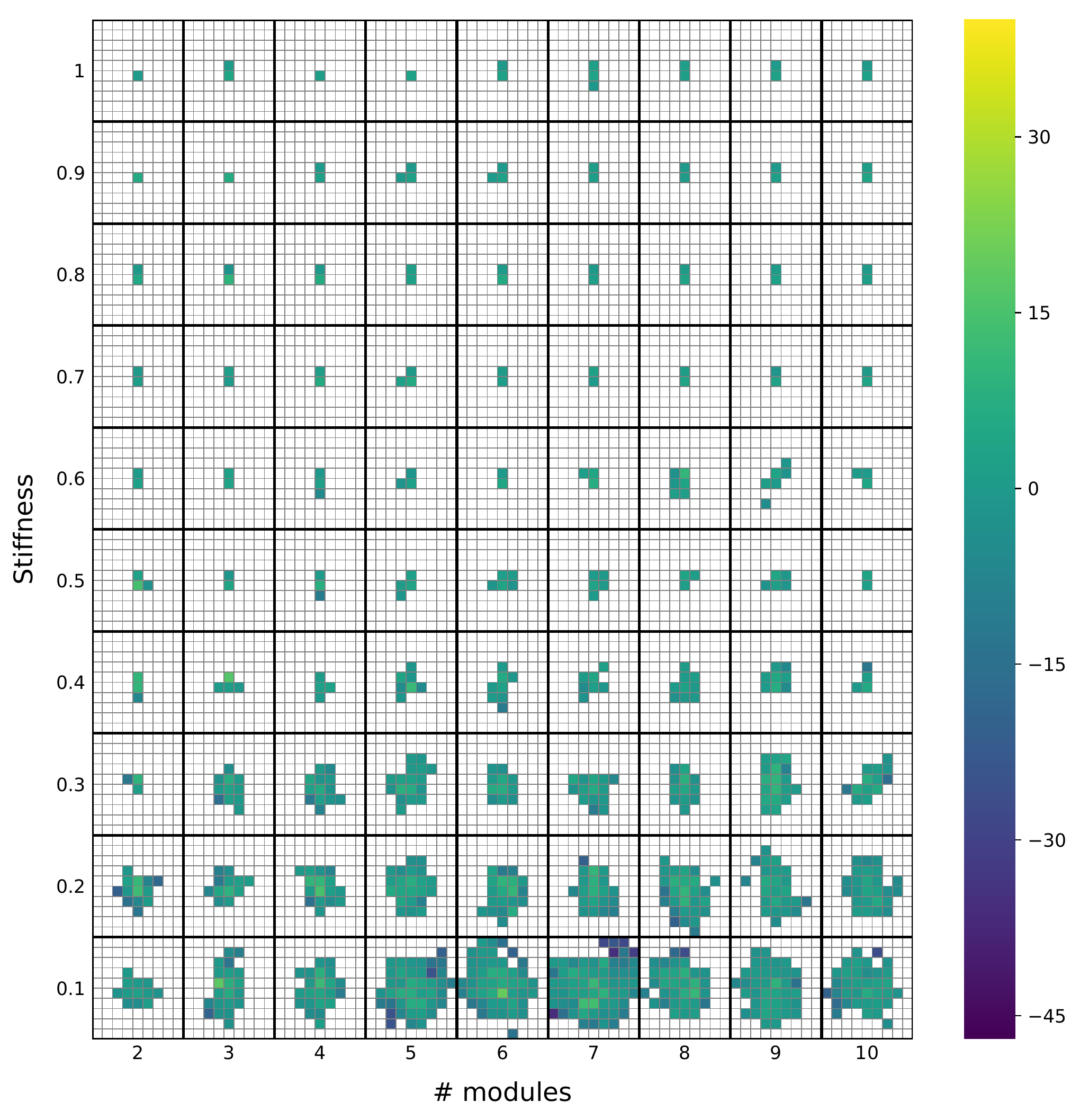}
      \label{fig:s-heat-gr-me-9}
    \end{subfigure} \\ \vspace{10pt}
    
    \vspace{5pt} \centerline{\textbf{Double Map MAP-Elites (DM-ME)}} \vspace{10pt}
    
    \begin{subfigure}{0.195\textwidth}
      \centering
      \includegraphics[width=\textwidth]{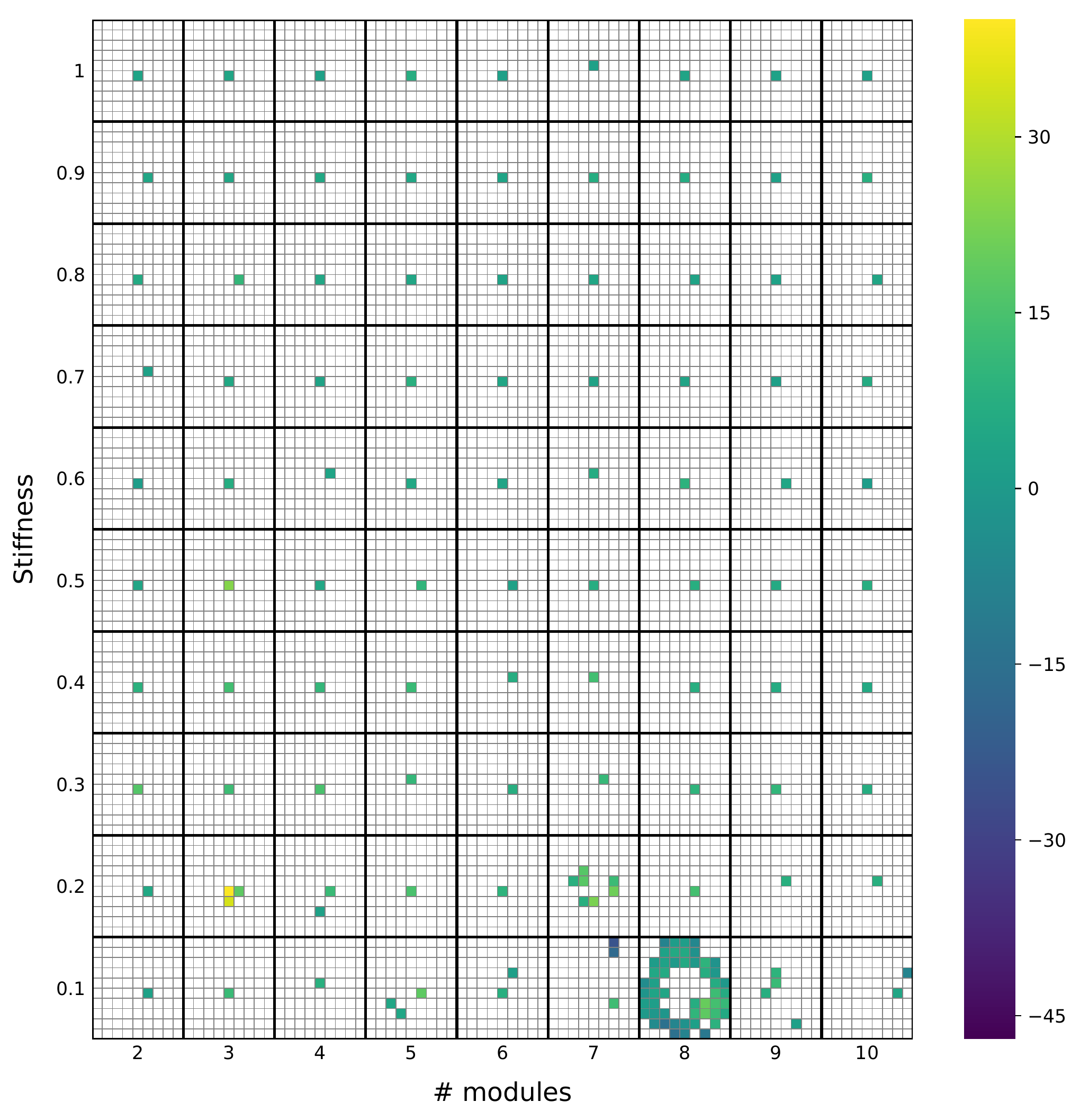}
      \label{fig:s-heat-gr-dm-me-0}
    \end{subfigure} 
    \begin{subfigure}{0.195\textwidth}
      \centering
      \includegraphics[width=\textwidth]{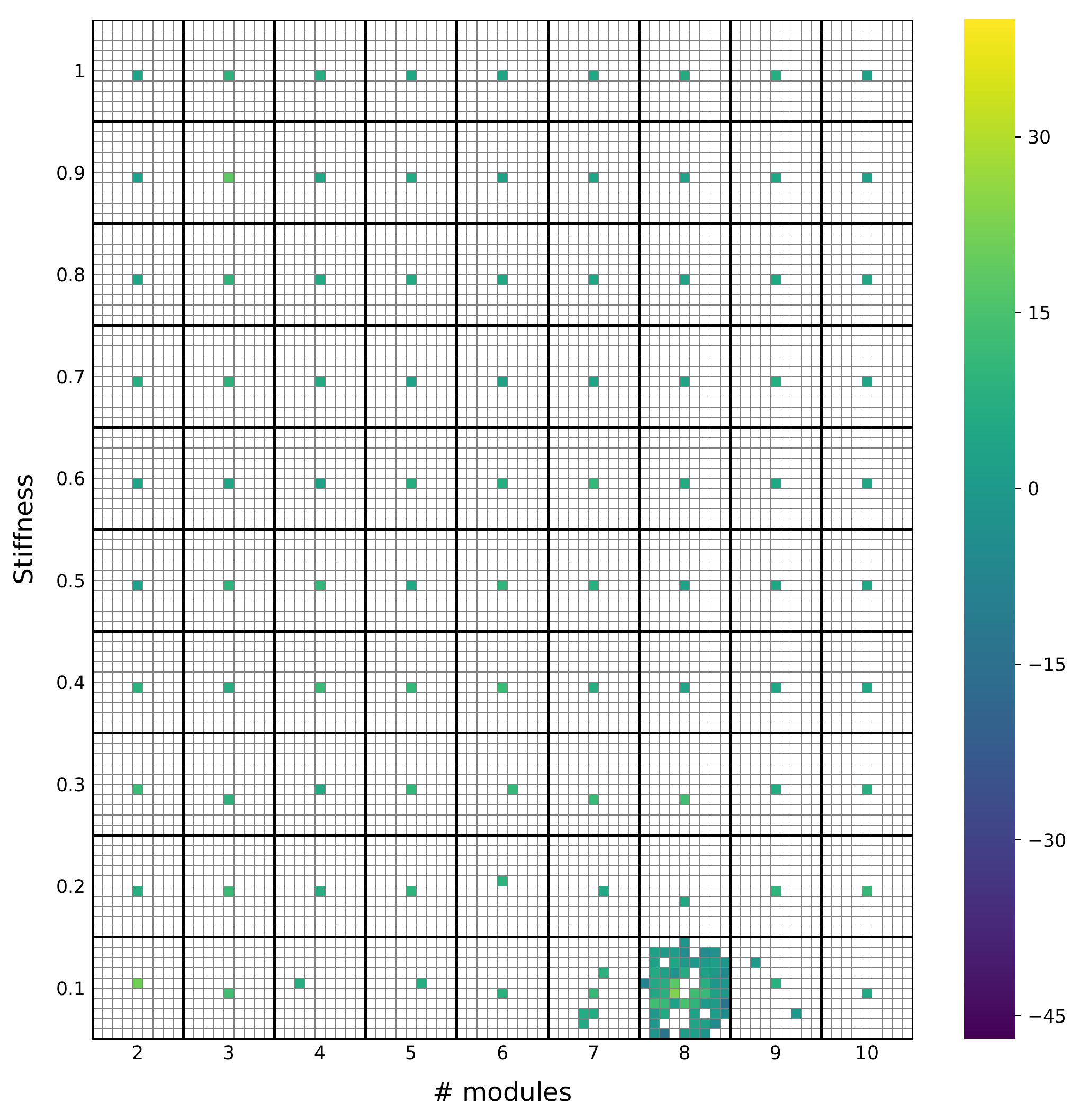}
      \label{fig:s-heat-gr-dm-me-1}
    \end{subfigure} 
    \begin{subfigure}{0.195\textwidth}
      \centering
      \includegraphics[width=\textwidth]{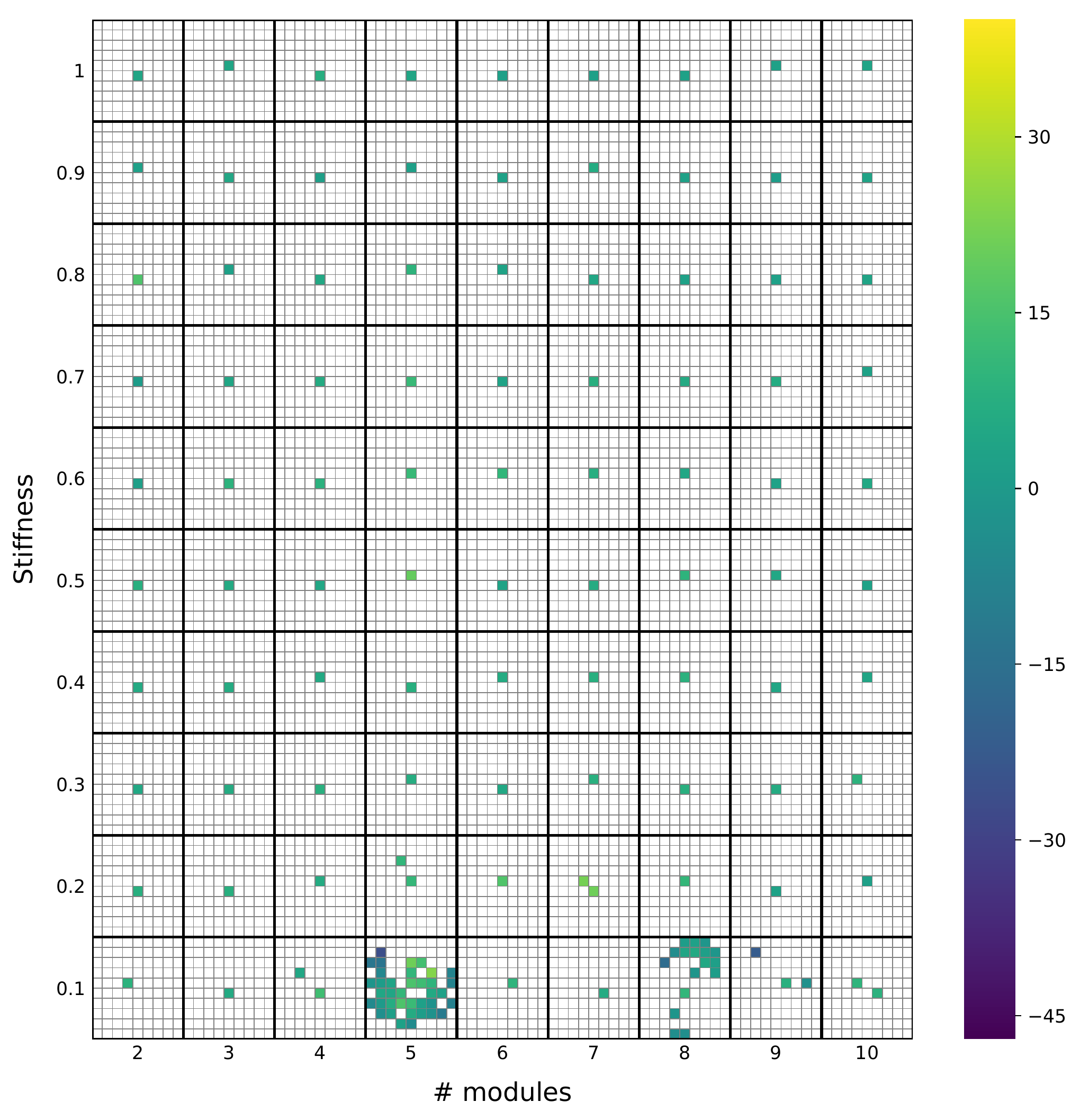}
      \label{fig:s-heat-gr-dm-me-2}
    \end{subfigure} 
    \begin{subfigure}{0.195\textwidth}
      \centering
      \includegraphics[width=\textwidth]{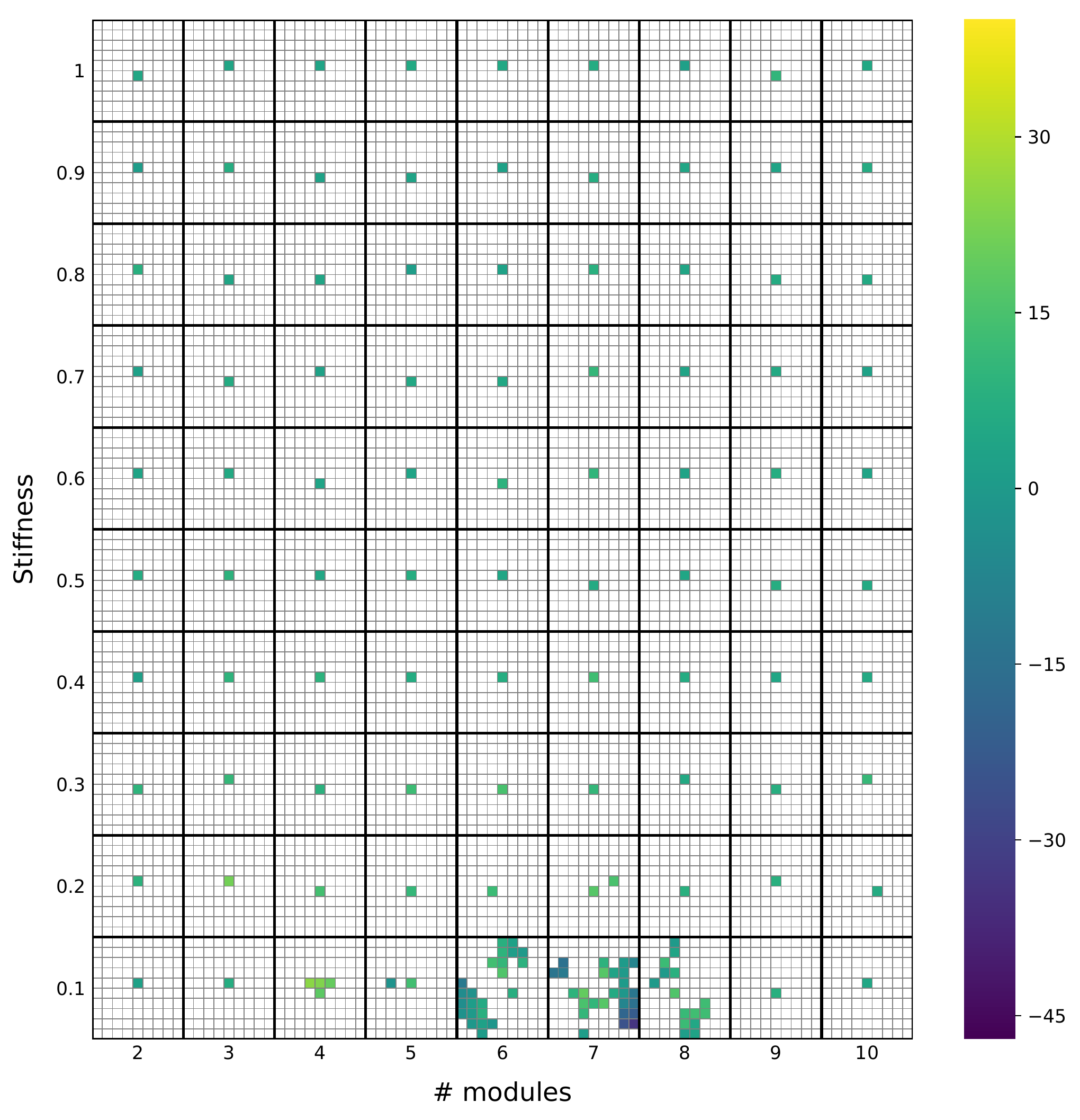}
      \label{fig:s-heat-gr-dm-me-3}
    \end{subfigure} 
    \begin{subfigure}{0.195\textwidth}
      \centering
      \includegraphics[width=\textwidth]{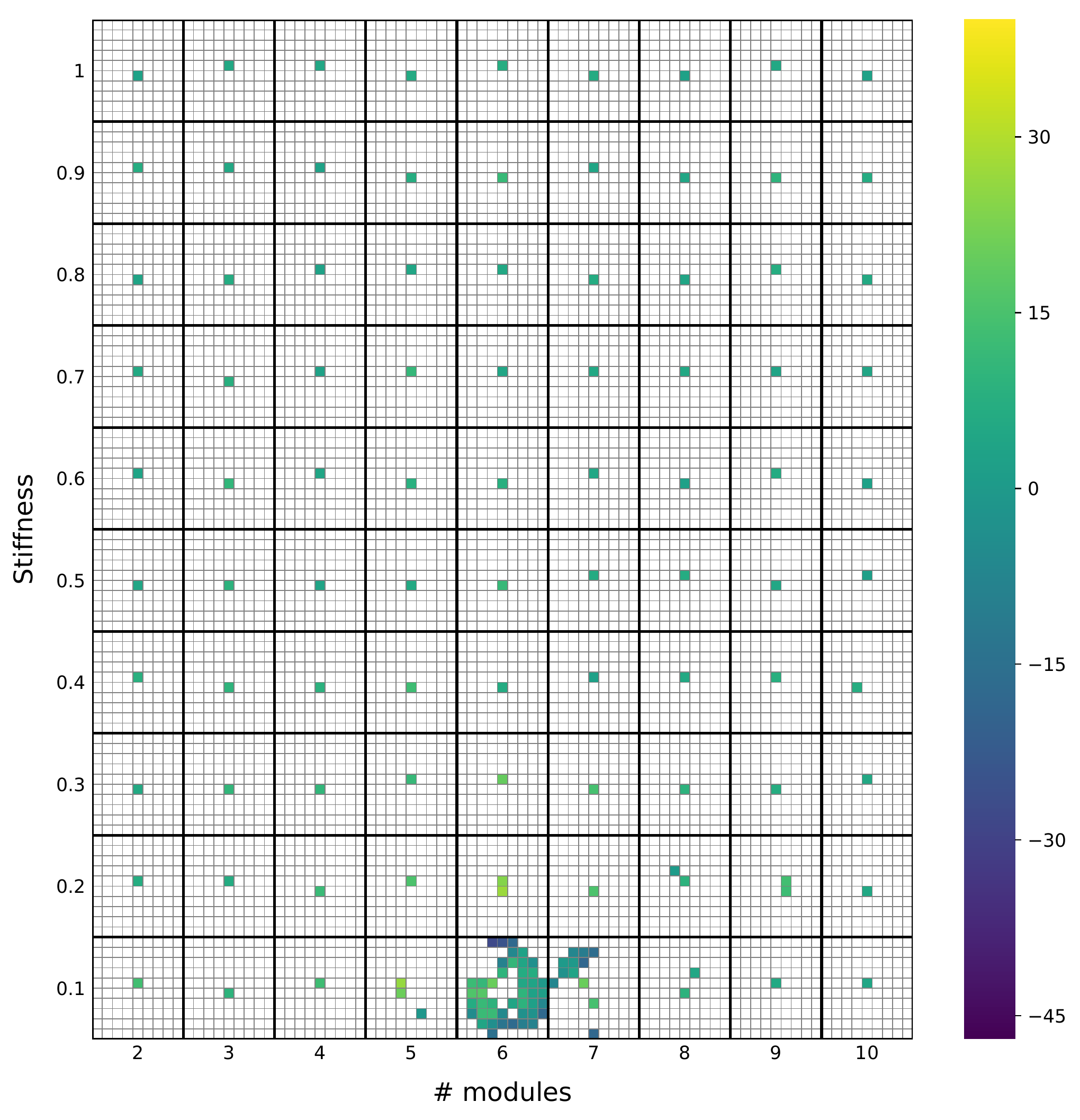}
      \label{fig:s-heat-gr-dm-me-4}
    \end{subfigure} \\ \vspace{-10pt}
    \begin{subfigure}{0.195\textwidth}
      \centering
      \includegraphics[width=\textwidth]{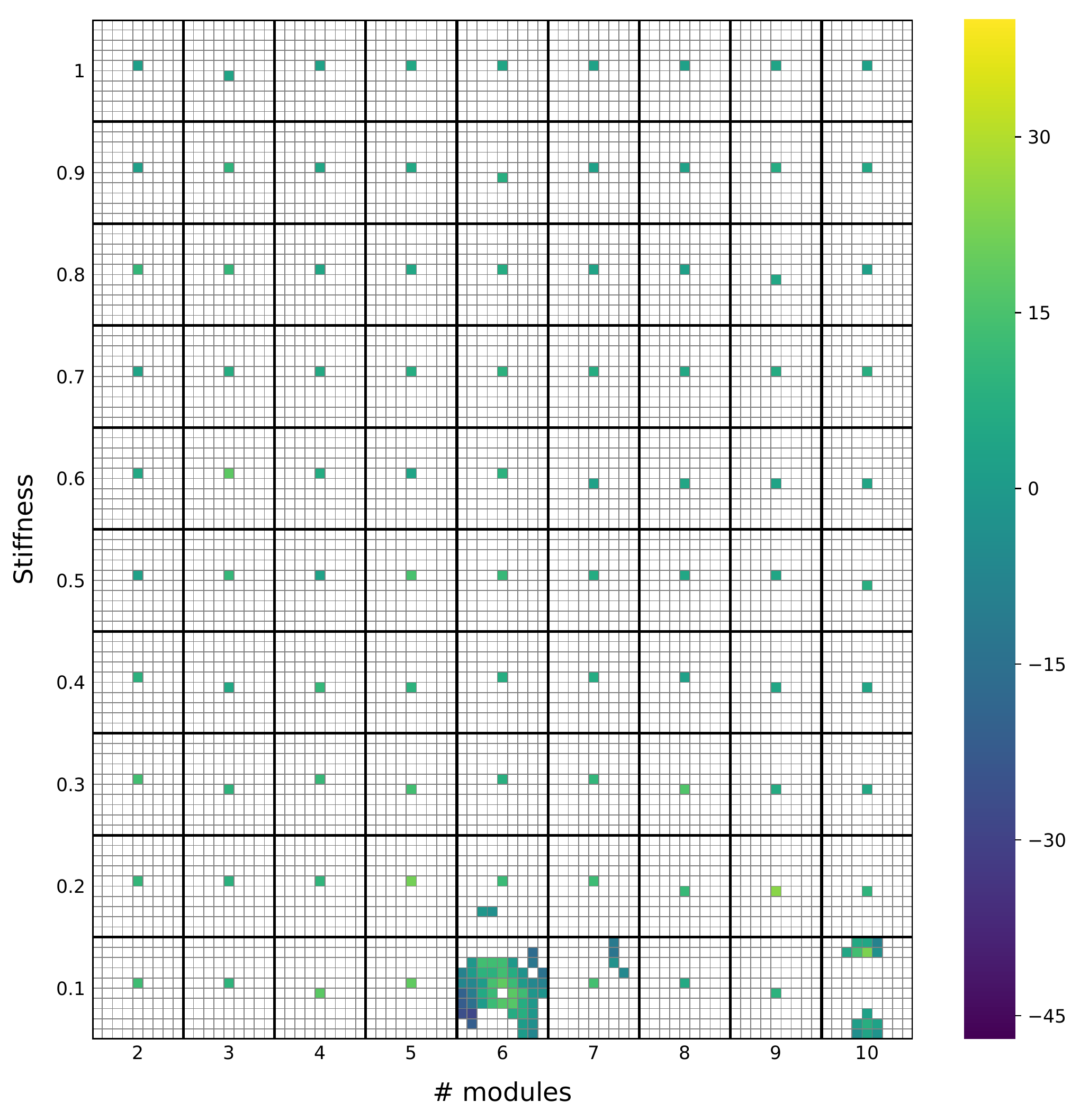}
      \label{fig:s-heat-gr-dm-me-5}
    \end{subfigure} 
    \begin{subfigure}{0.195\textwidth}
      \centering
      \includegraphics[width=\textwidth]{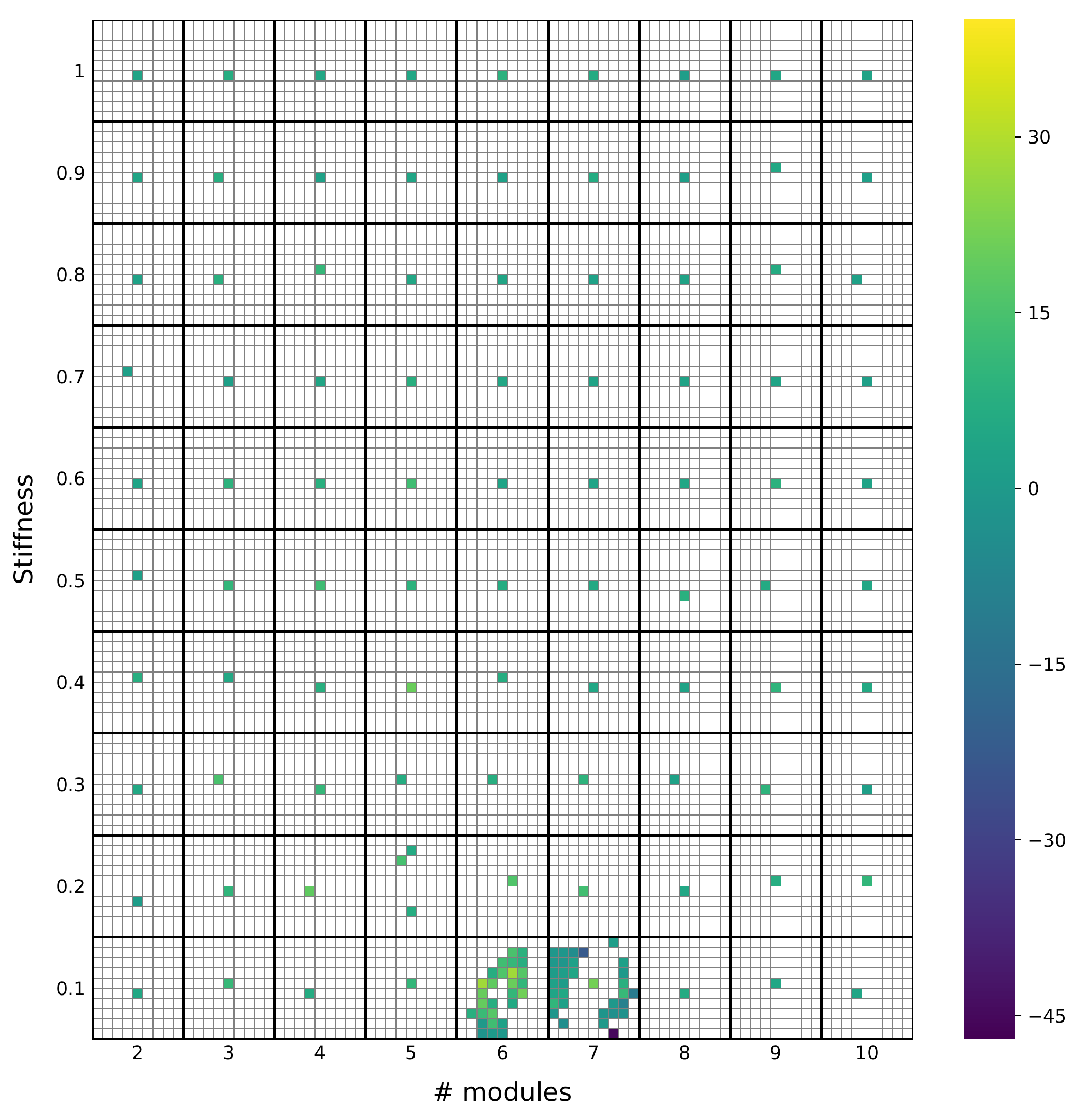}
      \label{fig:s-heat-gr-dm-me-6}
    \end{subfigure} 
    \begin{subfigure}{0.195\textwidth}
      \centering
      \includegraphics[width=\textwidth]{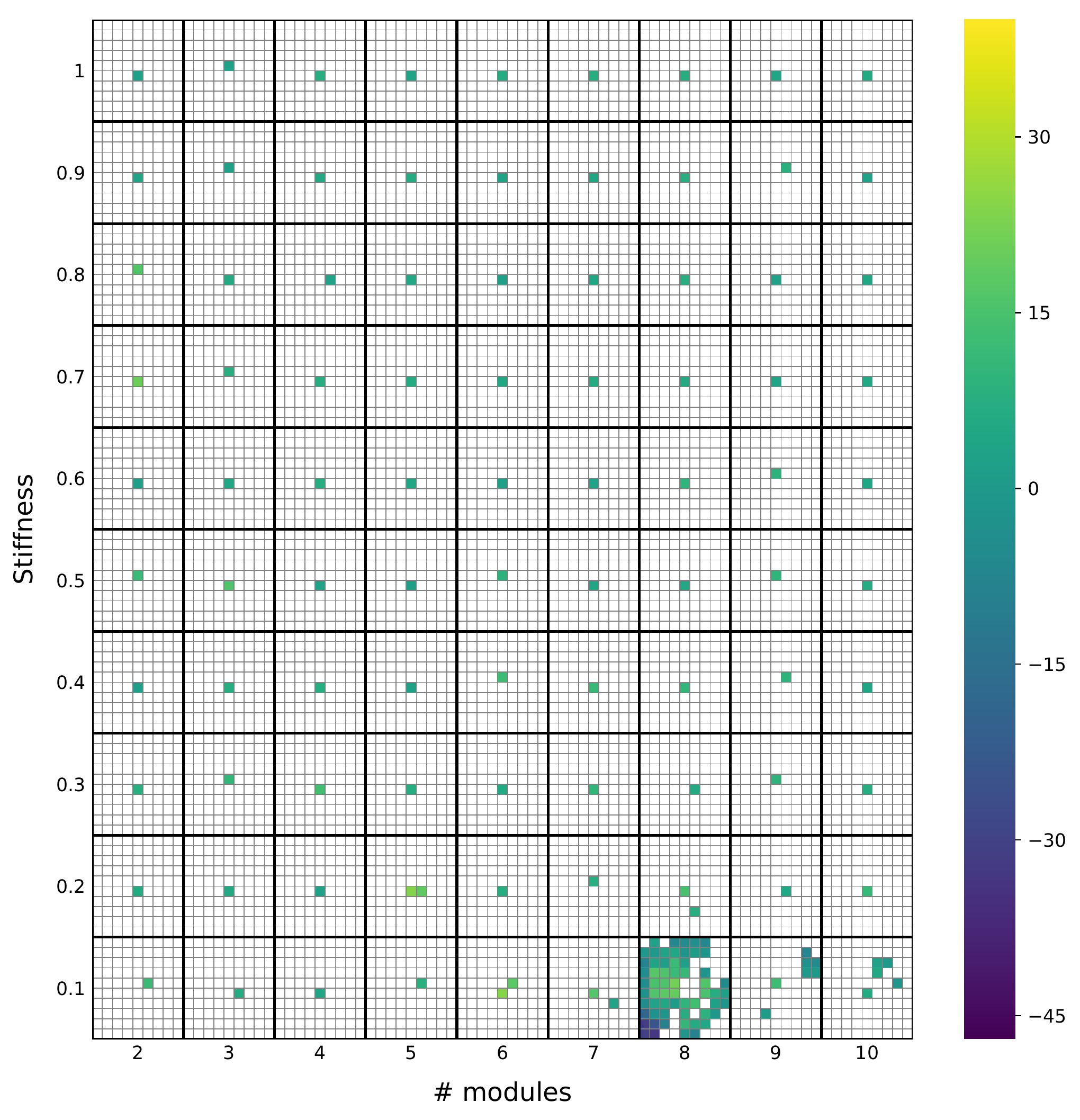}
      \label{fig:s-heat-gr-dm-me-7}
    \end{subfigure} 
    \begin{subfigure}{0.195\textwidth}
      \centering
      \includegraphics[width=\textwidth]{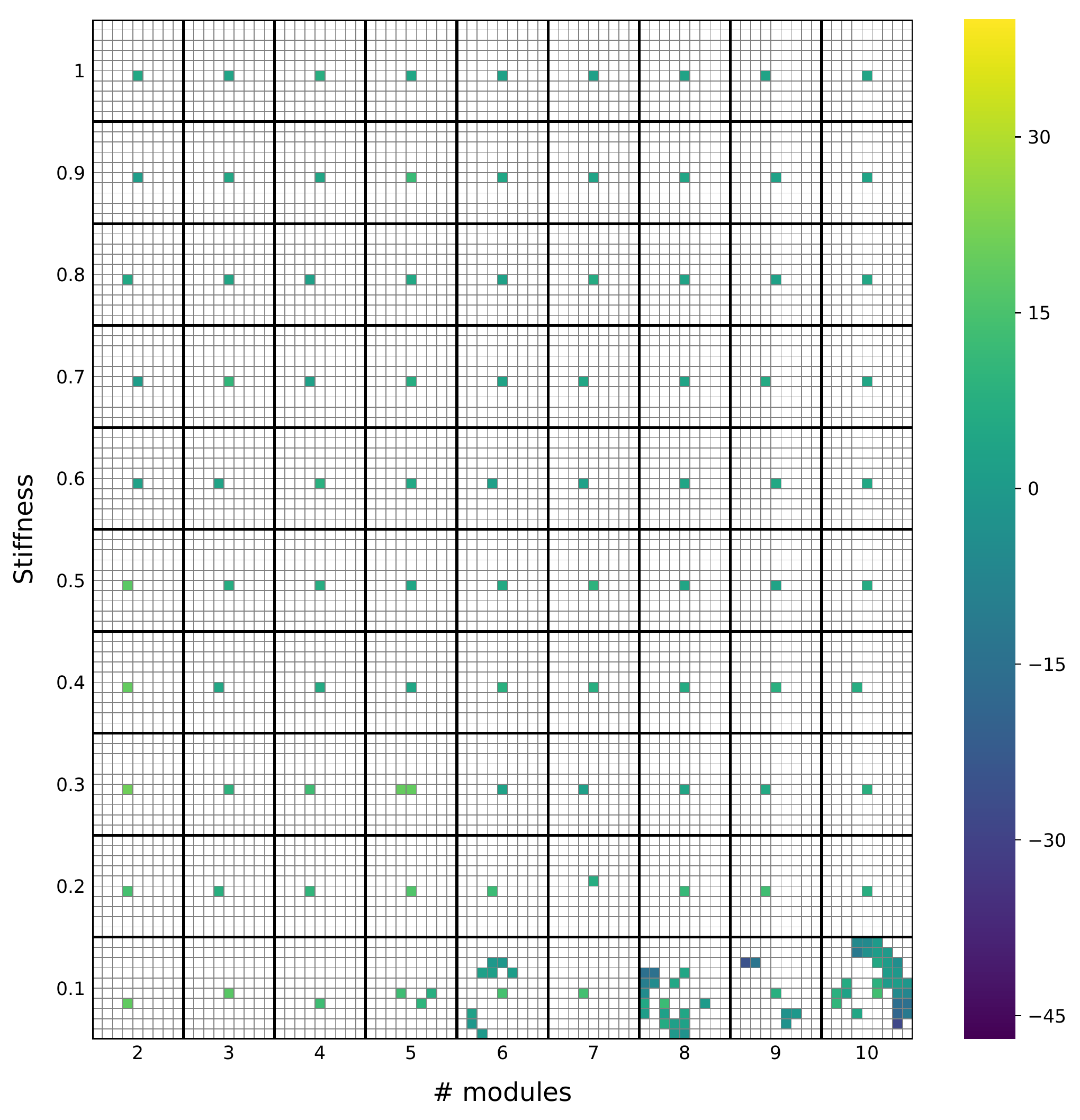}
      \label{fig:s-heat-gr-dm-me-8}
    \end{subfigure} 
    \begin{subfigure}{0.195\textwidth}
      \centering
      \includegraphics[width=\textwidth]{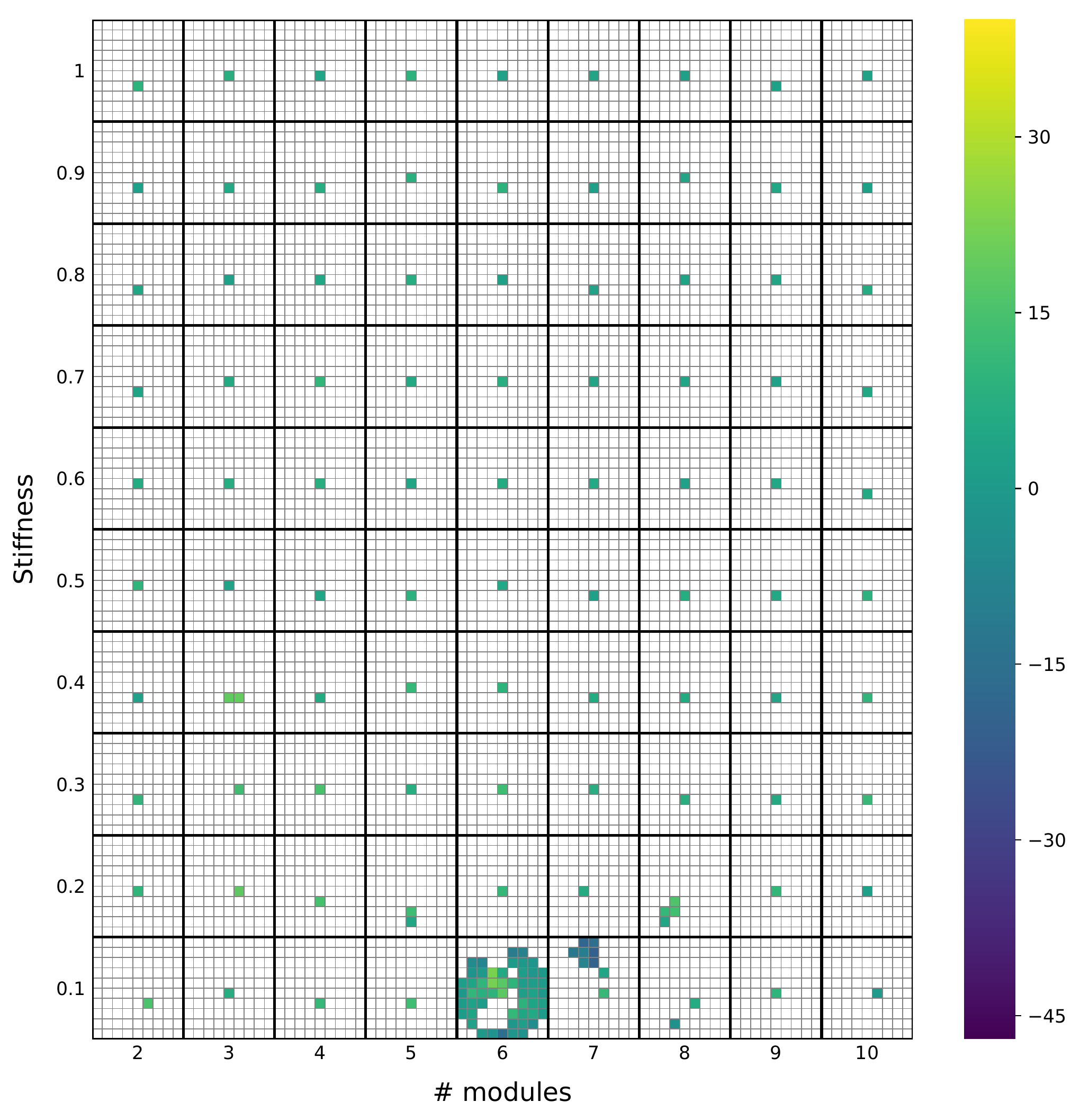}
      \label{fig:s-heat-gr-dm-me-9}
    \end{subfigure}
    \caption{\textbf{Single archives for the goal reaching task. In detail, the first two rows show the single archives generated by the 10 runs of MAP-Elites, whereas the other two show the projection of the double archives produced by Double Map MAP-Elites (DM-ME) onto a single one.}}
    \label{fig:s-heats-gr}
\end{figure*}


\begin{figure*}[h]
    \centering
    
    \vspace{10pt} \centerline{\textbf{MAP-Elites}} \vspace{10pt}
    \begin{subfigure}{0.195\textwidth}
      \centering
      \includegraphics[width=\textwidth]{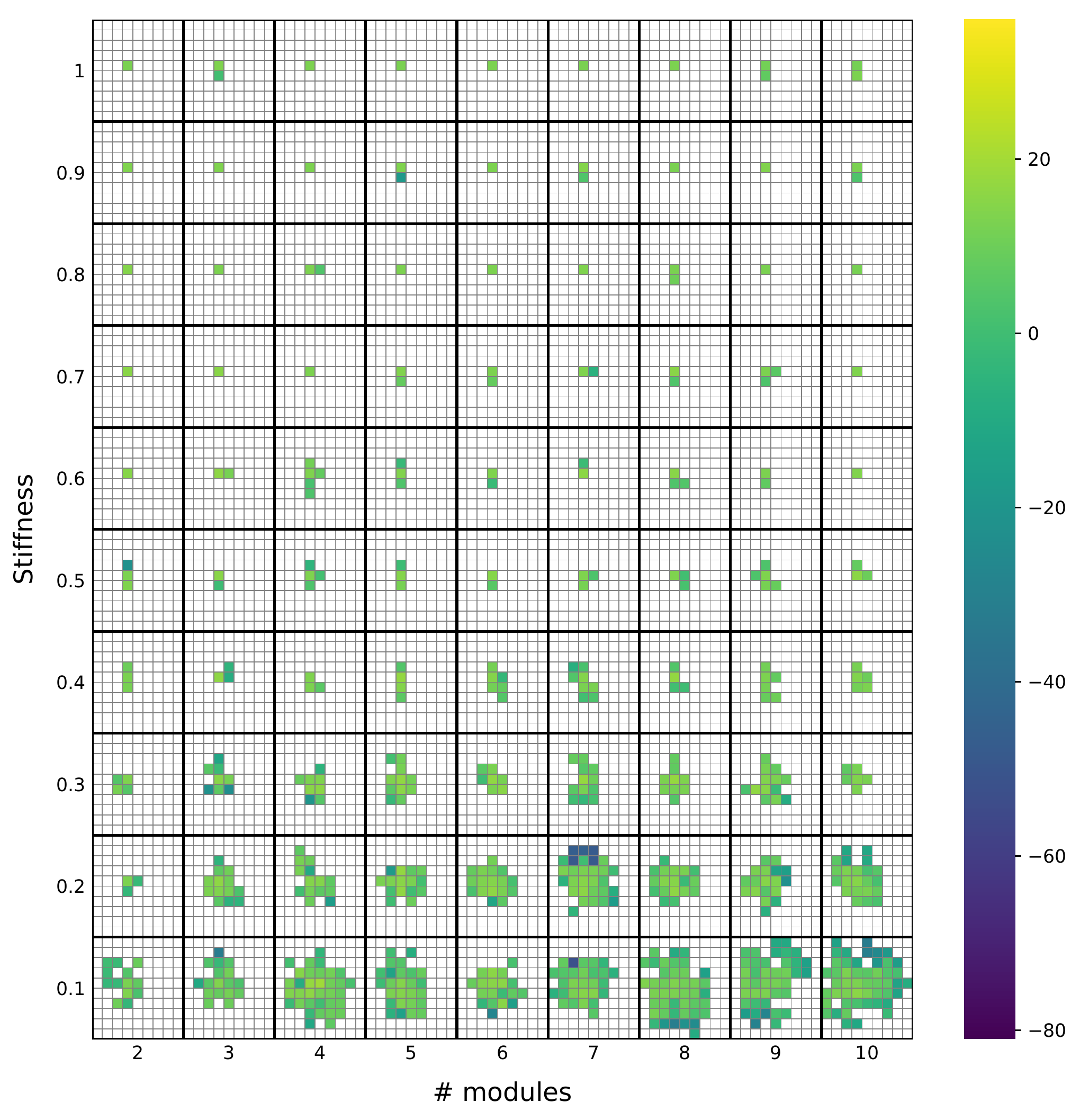}
      \label{fig:s-heat-sgr-me-0}
    \end{subfigure} 
    \begin{subfigure}{0.195\textwidth}
      \centering
      \includegraphics[width=\textwidth]{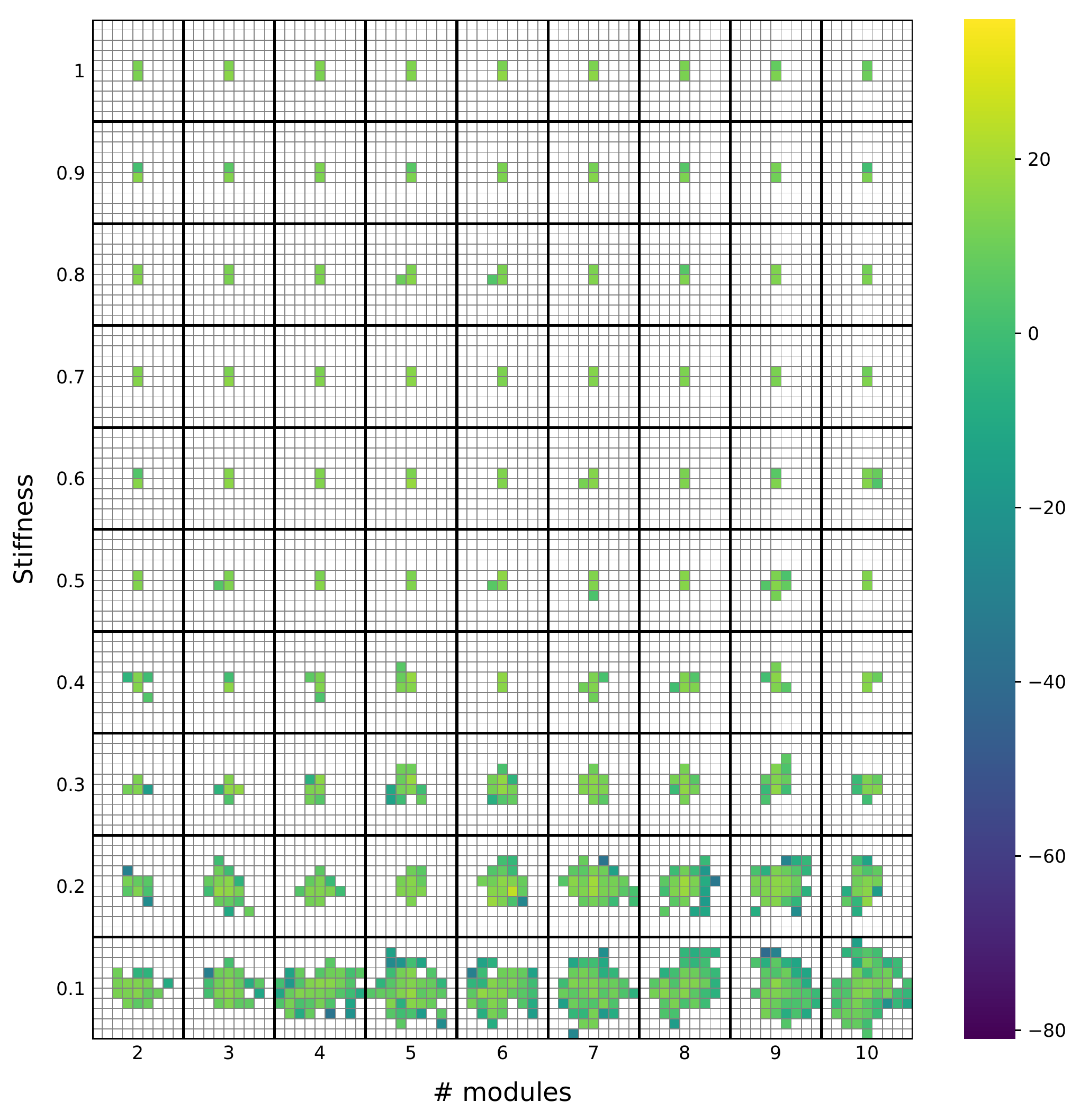}
      \label{fig:s-heat-sgr-me-1}
    \end{subfigure} 
    \begin{subfigure}{0.195\textwidth}
      \centering
      \includegraphics[width=\textwidth]{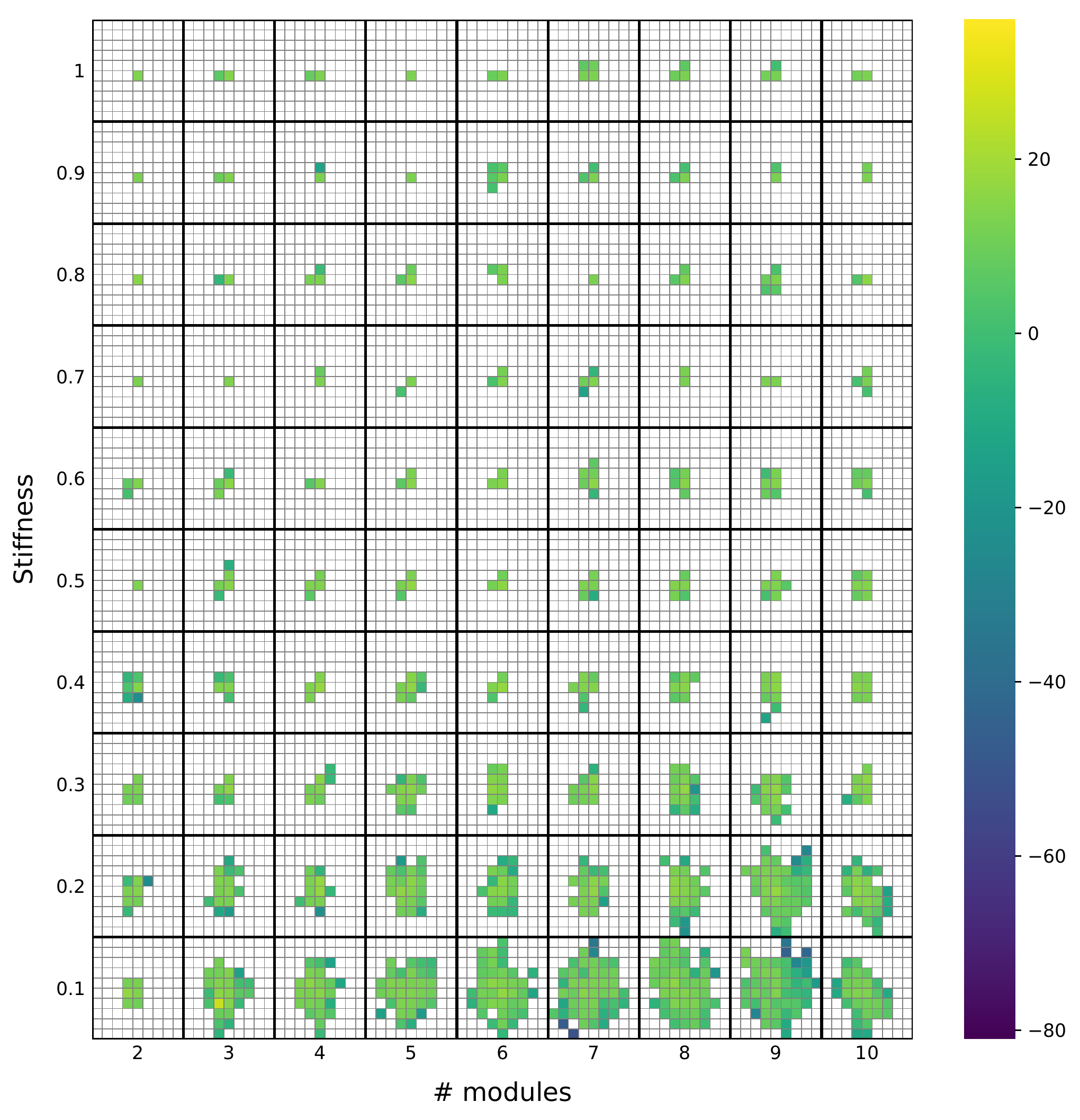}
      \label{fig:s-heat-sgr-me-2}
    \end{subfigure} 
    \begin{subfigure}{0.195\textwidth}
      \centering
      \includegraphics[width=\textwidth]{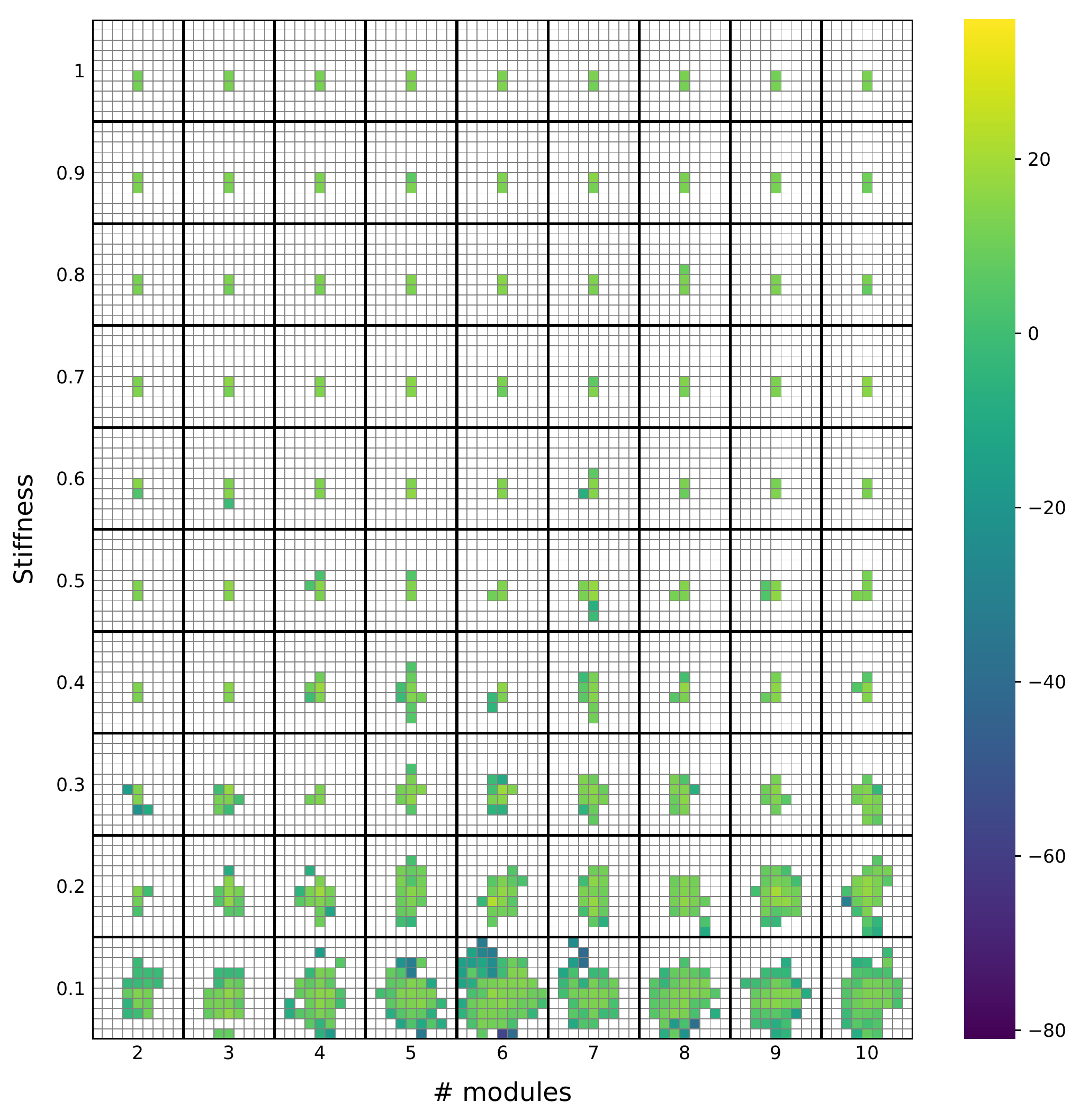}
      \label{fig:s-heat-sgr-me-3}
    \end{subfigure} 
    \begin{subfigure}{0.195\textwidth}
      \centering
      \includegraphics[width=\textwidth]{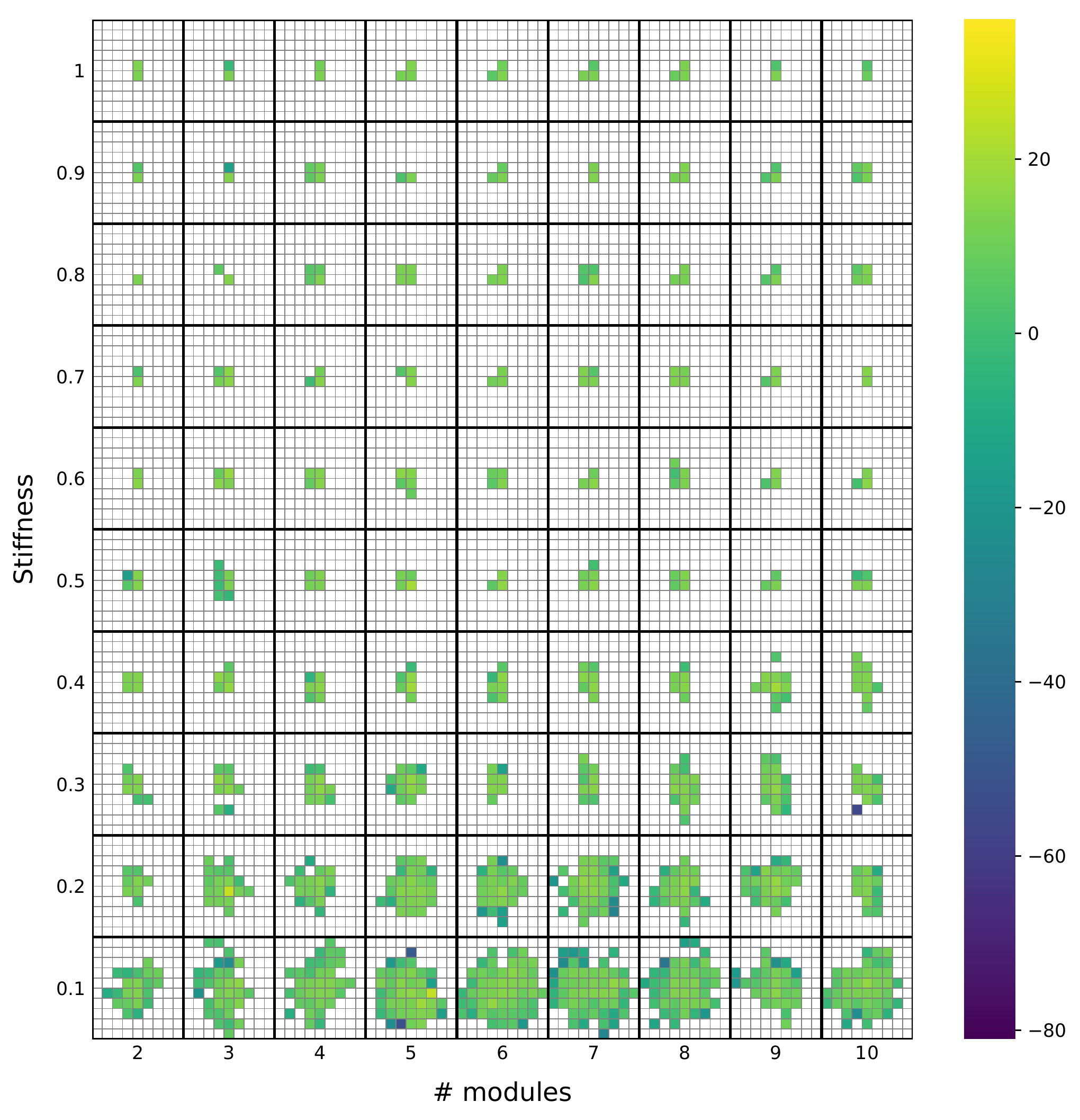}
      \label{fig:s-heat-sgr-me-4}
    \end{subfigure} \\ \vspace{-10pt}
    \begin{subfigure}{0.195\textwidth}
      \centering
      \includegraphics[width=\textwidth]{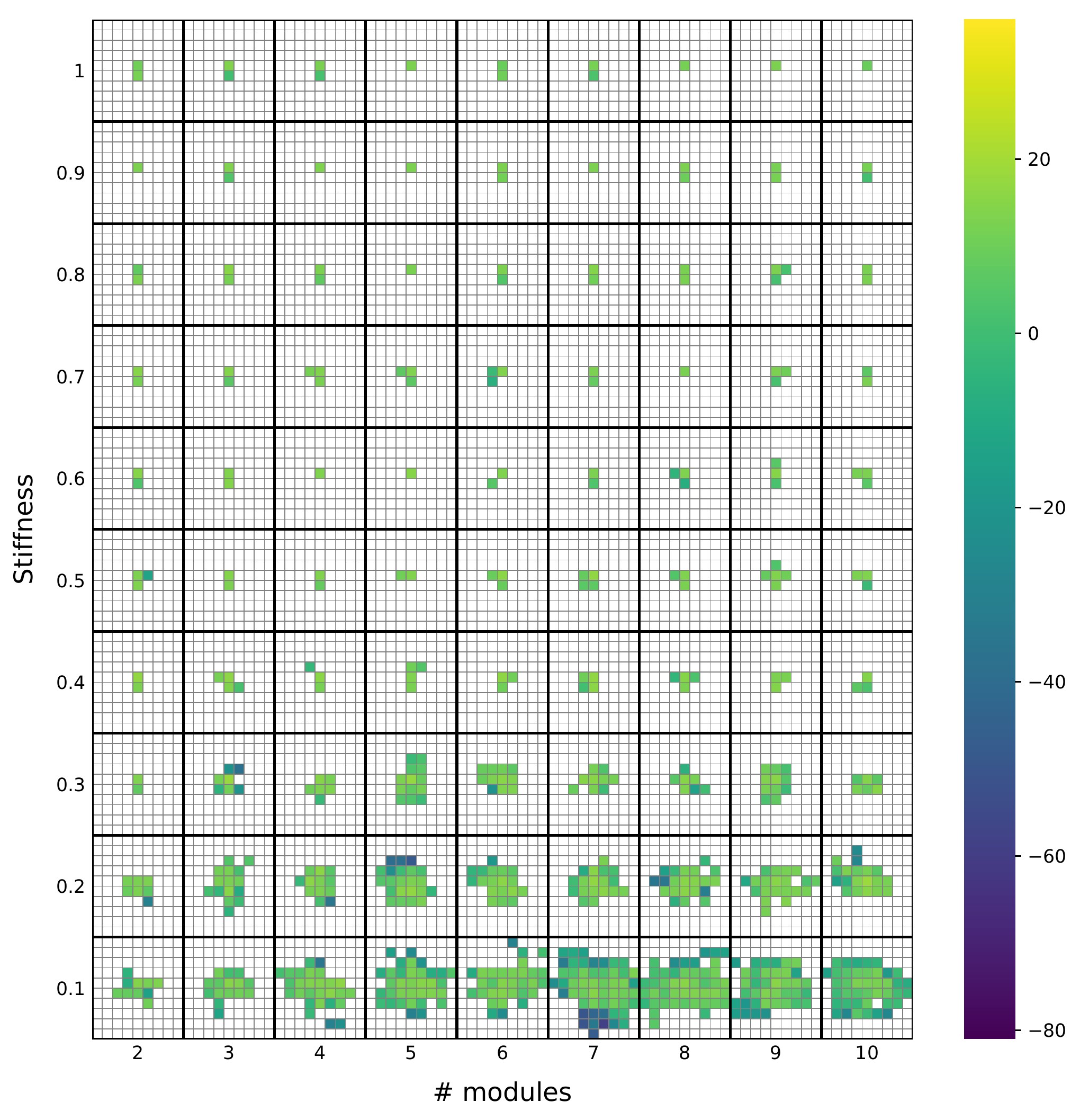}
      \label{fig:s-heat-sgr-me-5}
    \end{subfigure} 
    \begin{subfigure}{0.195\textwidth}
      \centering
      \includegraphics[width=\textwidth]{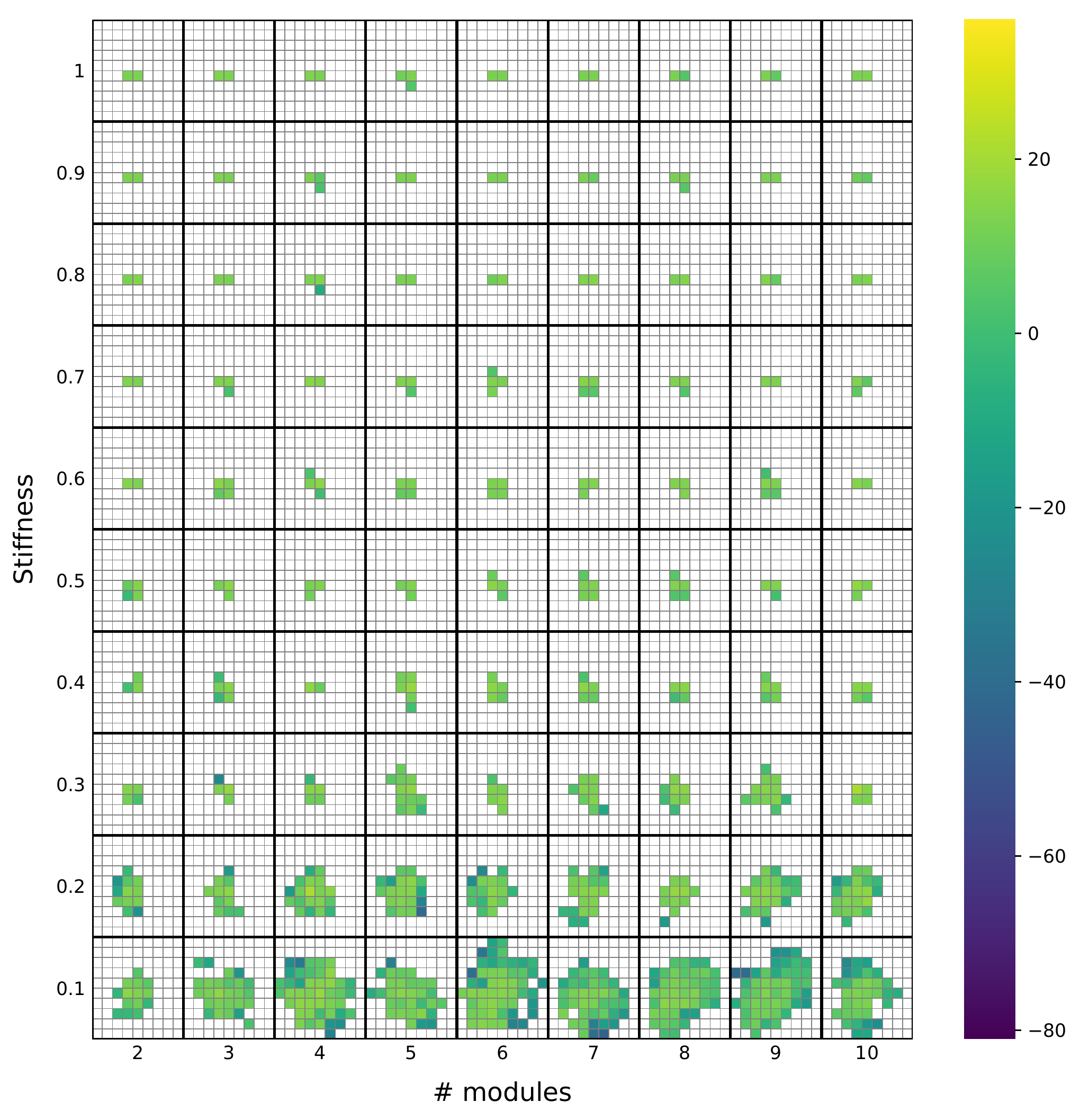}
      \label{fig:s-heat-sgr-me-6}
    \end{subfigure} 
    \begin{subfigure}{0.195\textwidth}
      \centering
      \includegraphics[width=\textwidth]{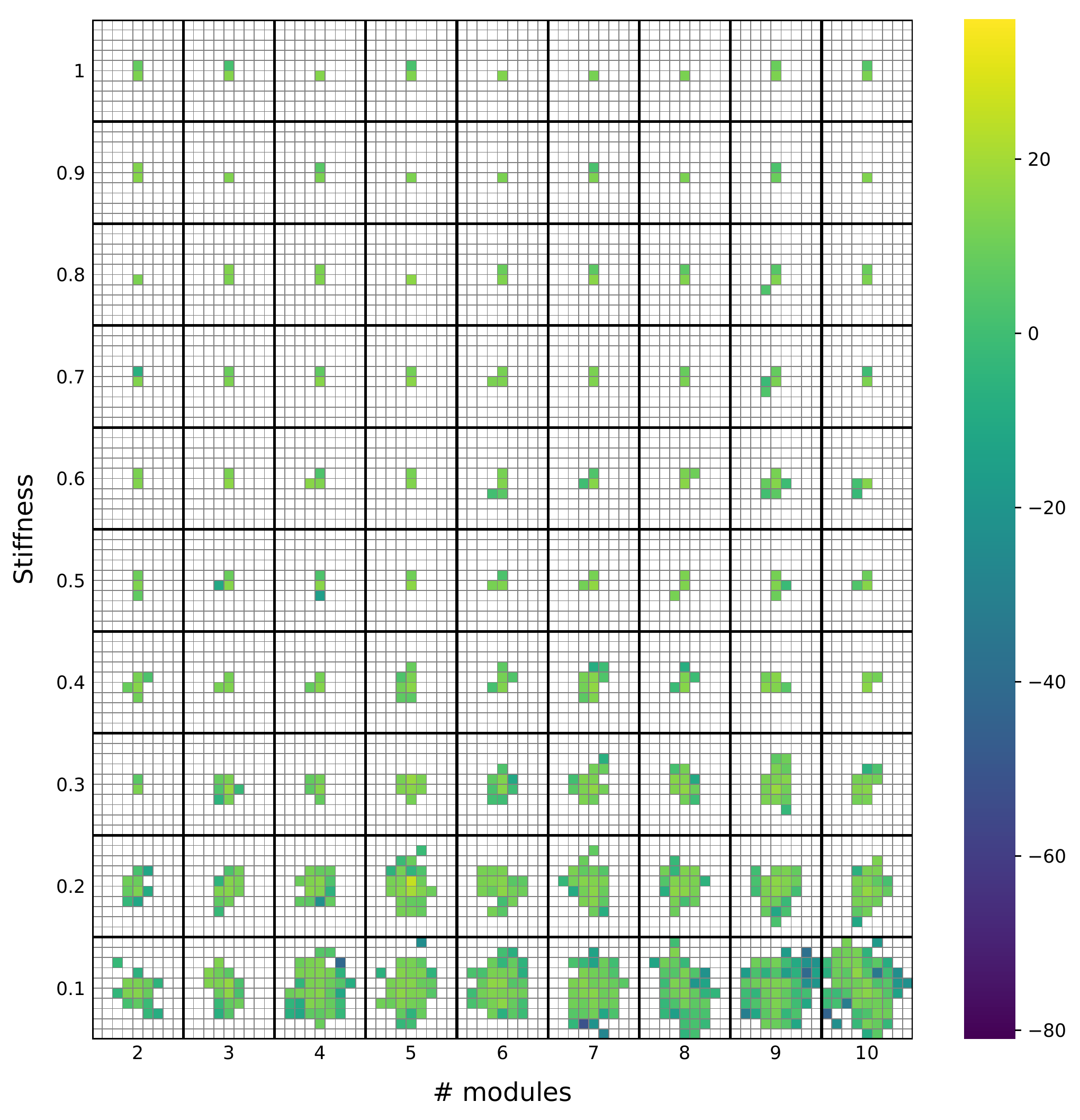}
      \label{fig:s-heat-sgr-me-7}
    \end{subfigure} 
    \begin{subfigure}{0.195\textwidth}
      \centering
      \includegraphics[width=\textwidth]{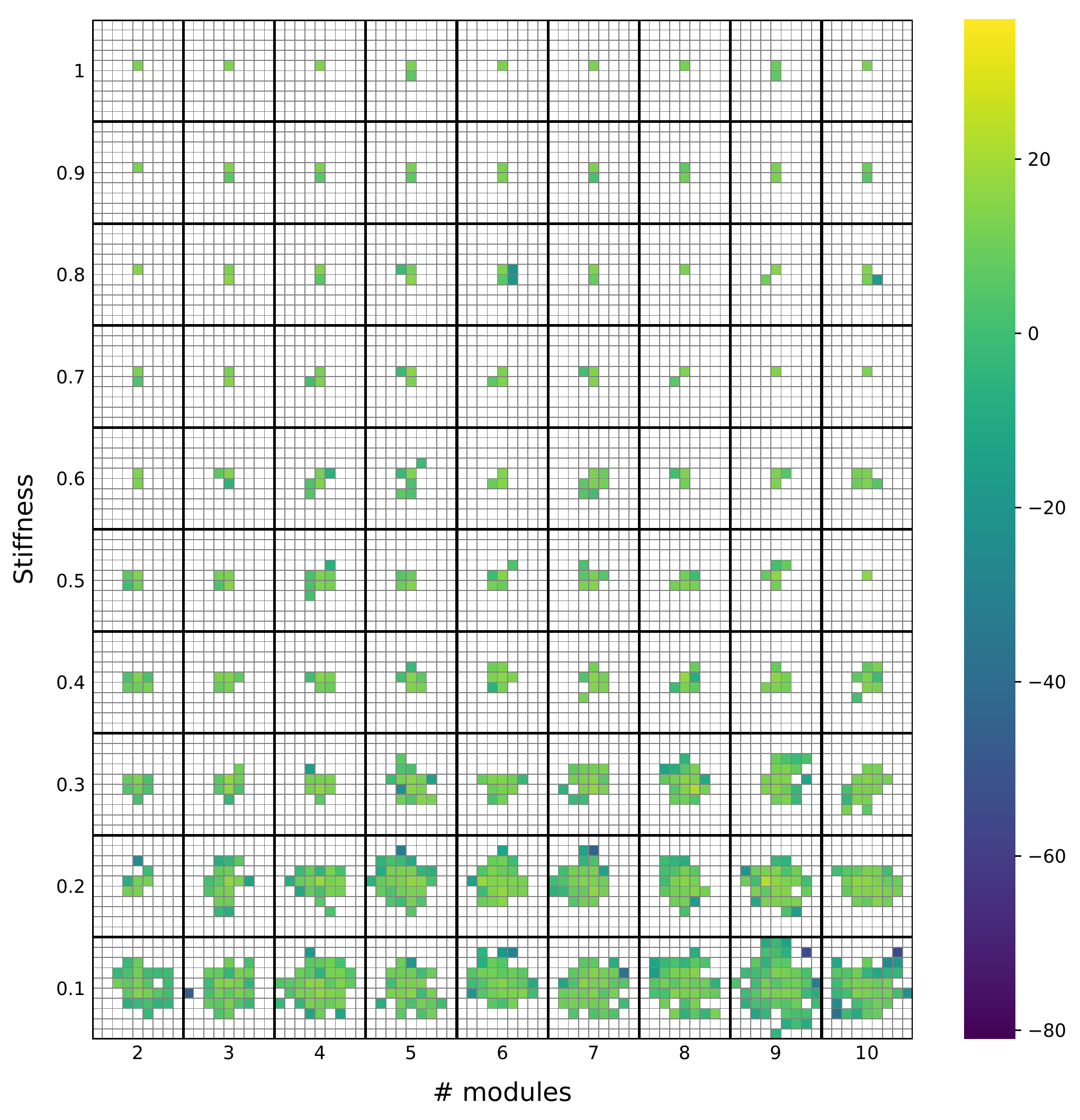}
      \label{fig:s-heat-sgr-me-8}
    \end{subfigure} 
    \begin{subfigure}{0.195\textwidth}
      \centering
      \includegraphics[width=\textwidth]{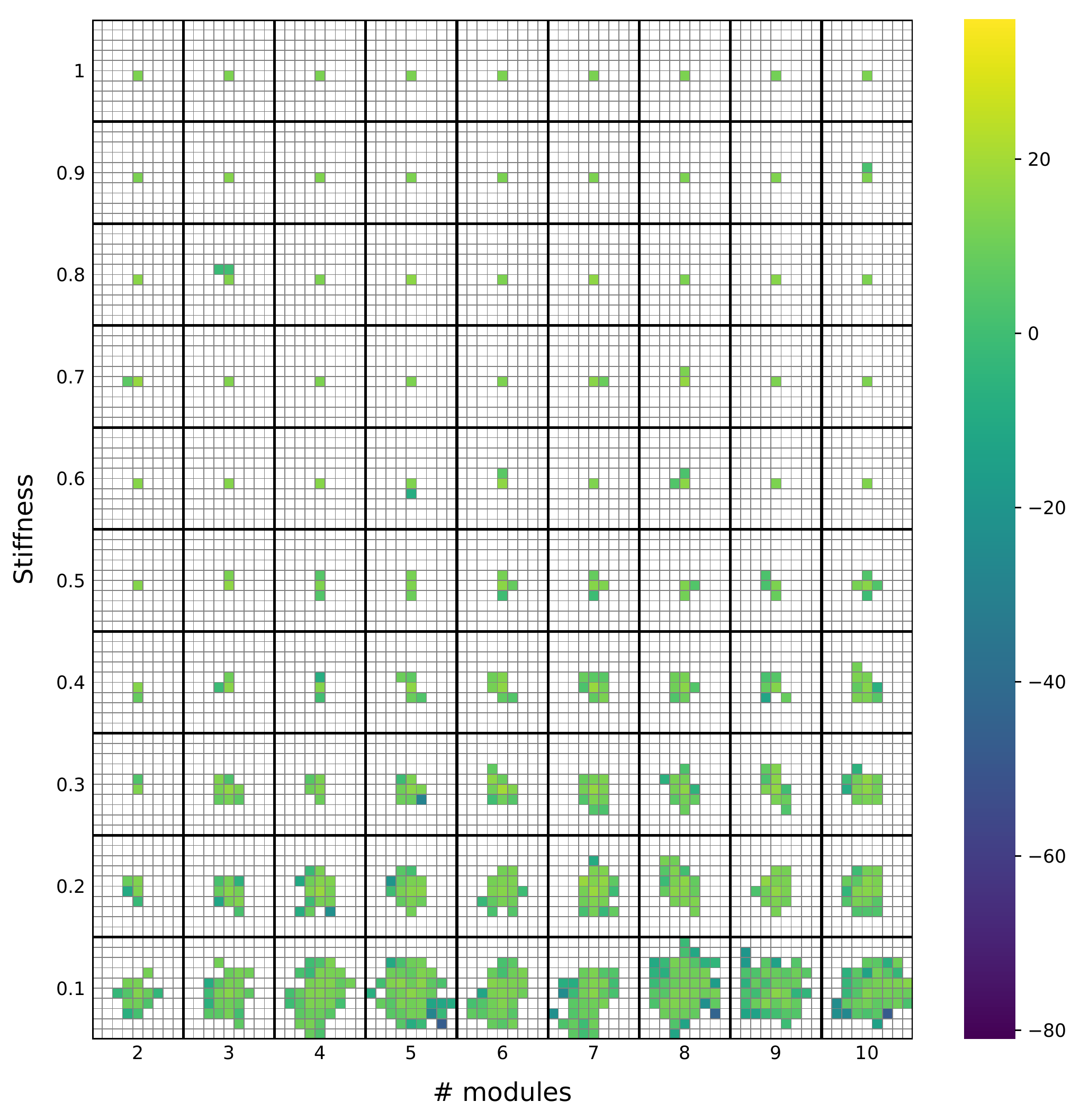}
      \label{fig:s-heat-sgr-me-9}
    \end{subfigure} \\
    
    \vspace{5pt} \centerline{\textbf{Double Map MAP-Elites (DM-ME)}} \vspace{10pt}
    
    \begin{subfigure}{0.195\textwidth}
      \centering
      \includegraphics[width=\textwidth]{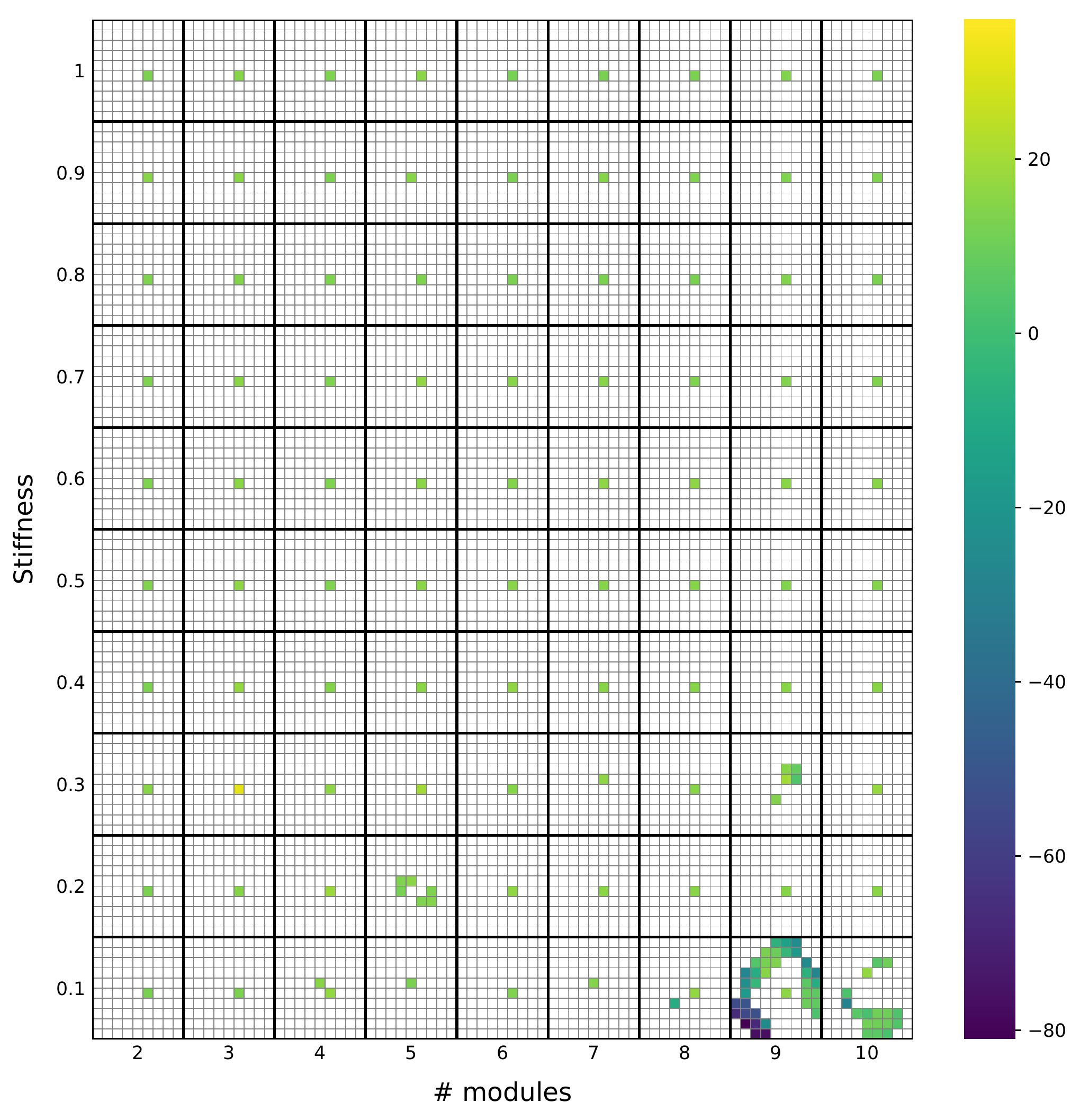}
      \label{fig:s-heat-sgr-dm-me-0}
    \end{subfigure} 
    \begin{subfigure}{0.195\textwidth}
      \centering
      \includegraphics[width=\textwidth]{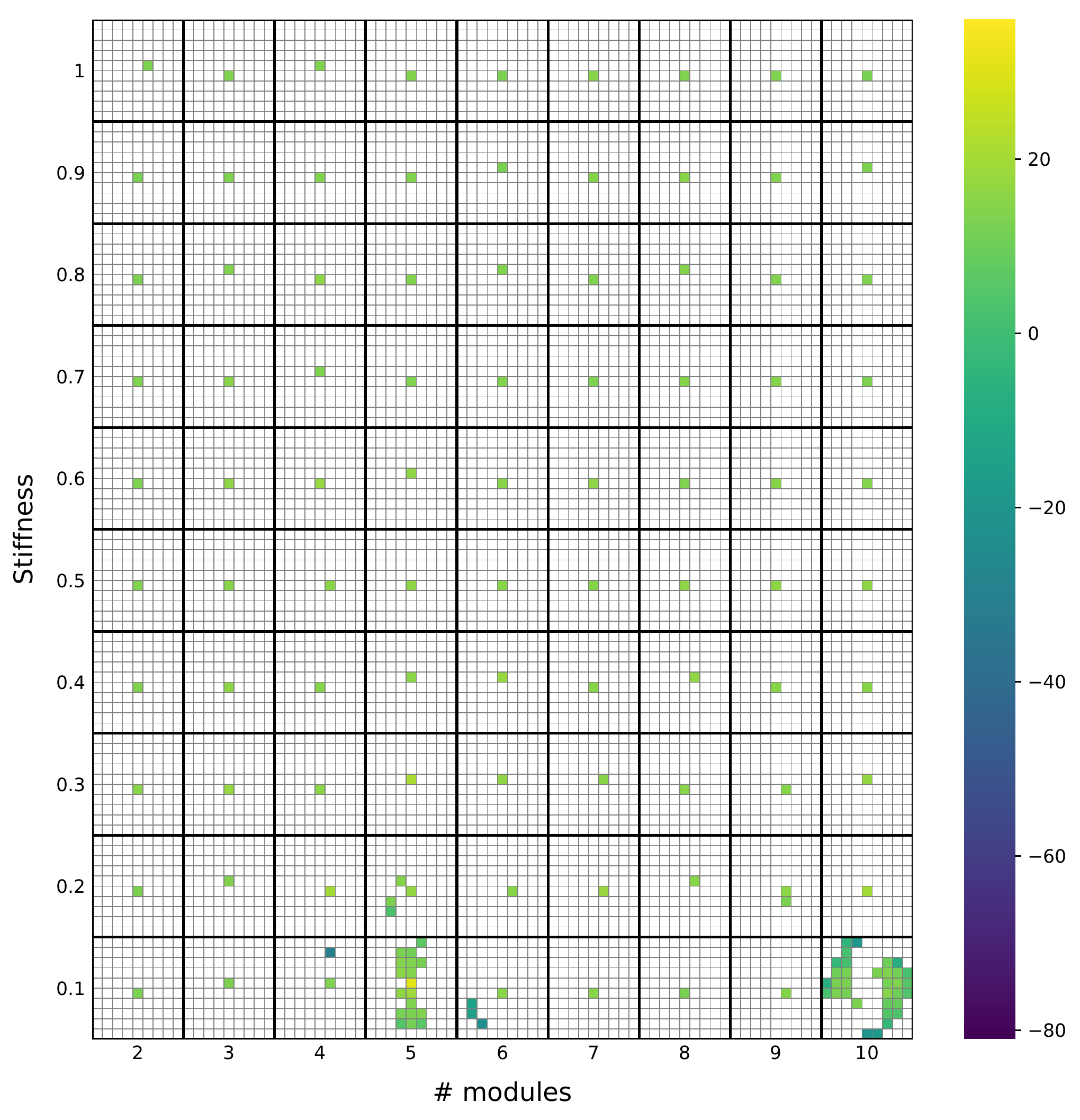}
      \label{fig:s-heat-sgr-dm-me-1}
    \end{subfigure} 
    \begin{subfigure}{0.195\textwidth}
      \centering
      \includegraphics[width=\textwidth]{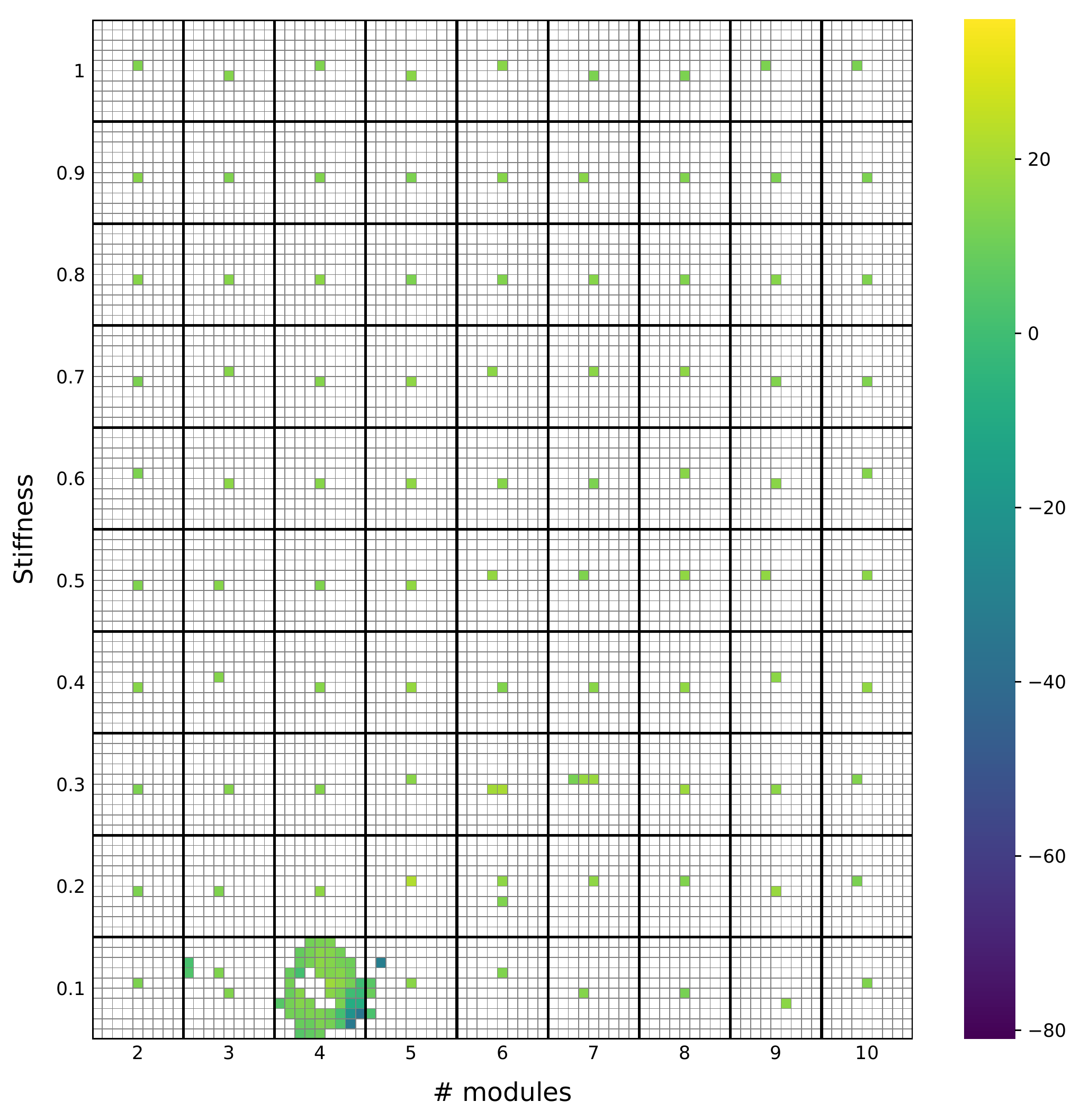}
      \label{fig:s-heat-sgr-dm-me-2}
    \end{subfigure} 
    \begin{subfigure}{0.195\textwidth}
      \centering
      \includegraphics[width=\textwidth]{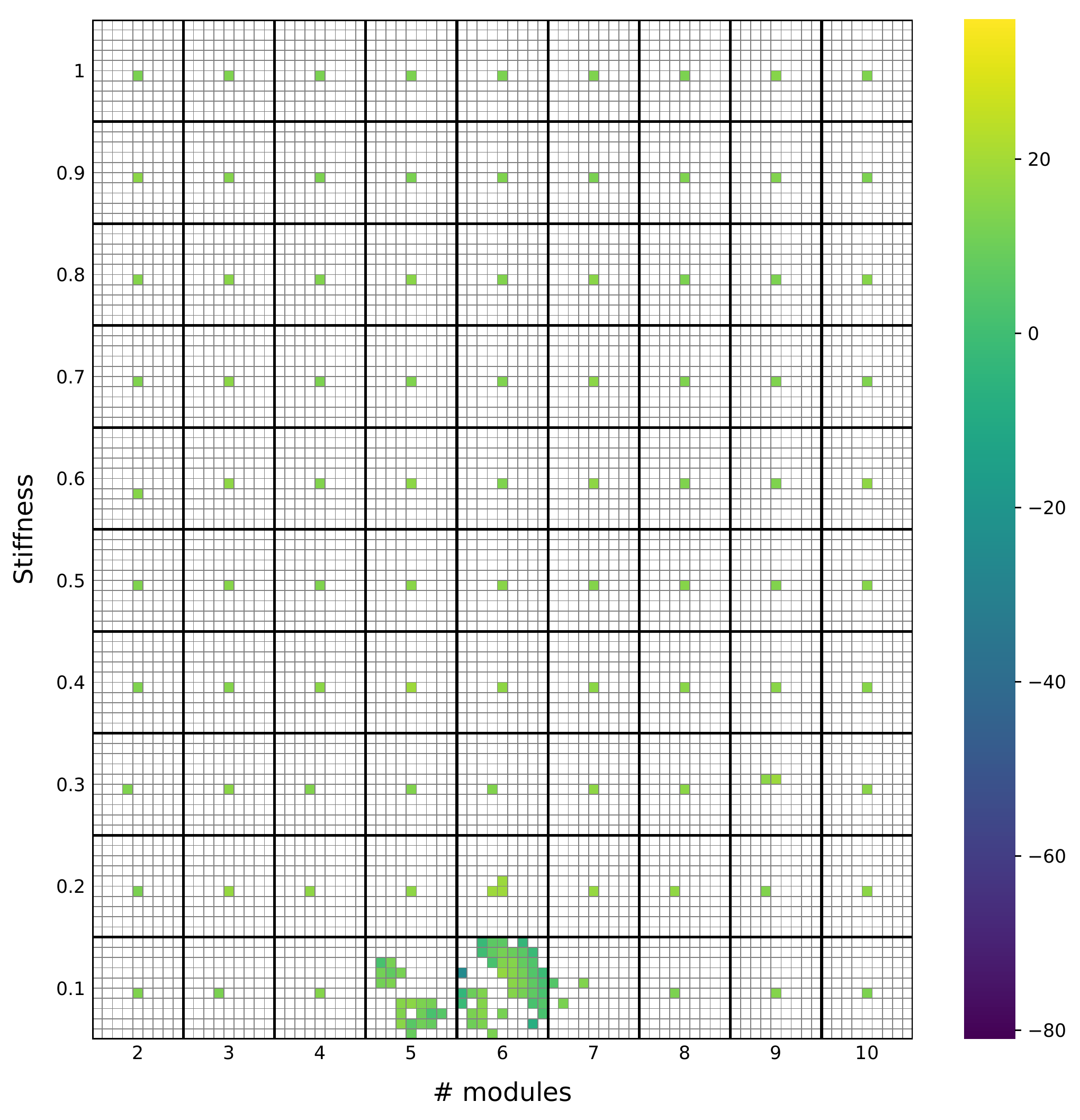}
      \label{fig:s-heat-sgr-dm-me-3}
    \end{subfigure} 
    \begin{subfigure}{0.195\textwidth}
      \centering
      \includegraphics[width=\textwidth]{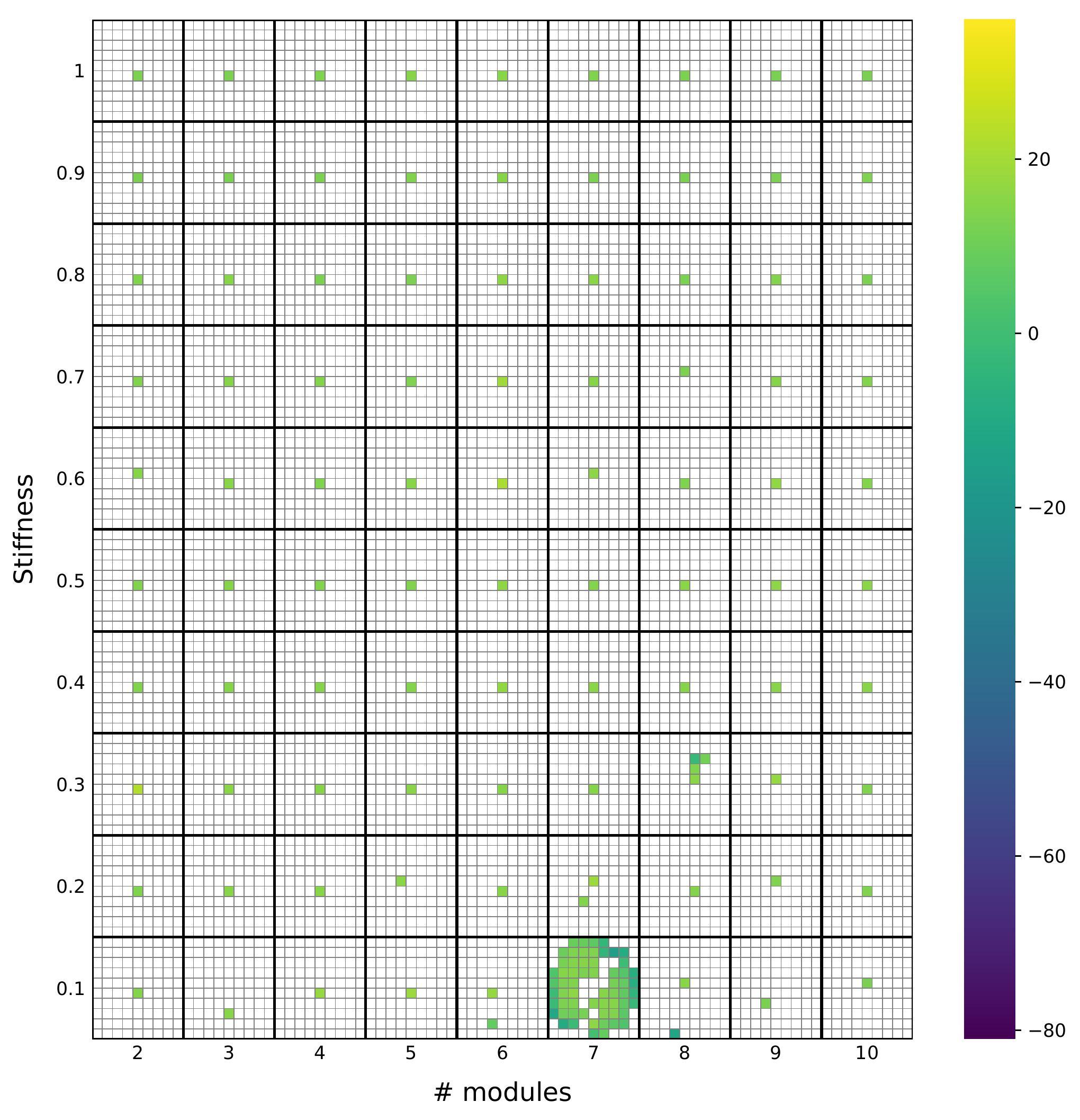}
      \label{fig:s-heat-sgr-dm-me-4}
    \end{subfigure} \\ \vspace{-10pt}
    \begin{subfigure}{0.195\textwidth}
      \centering
      \includegraphics[width=\textwidth]{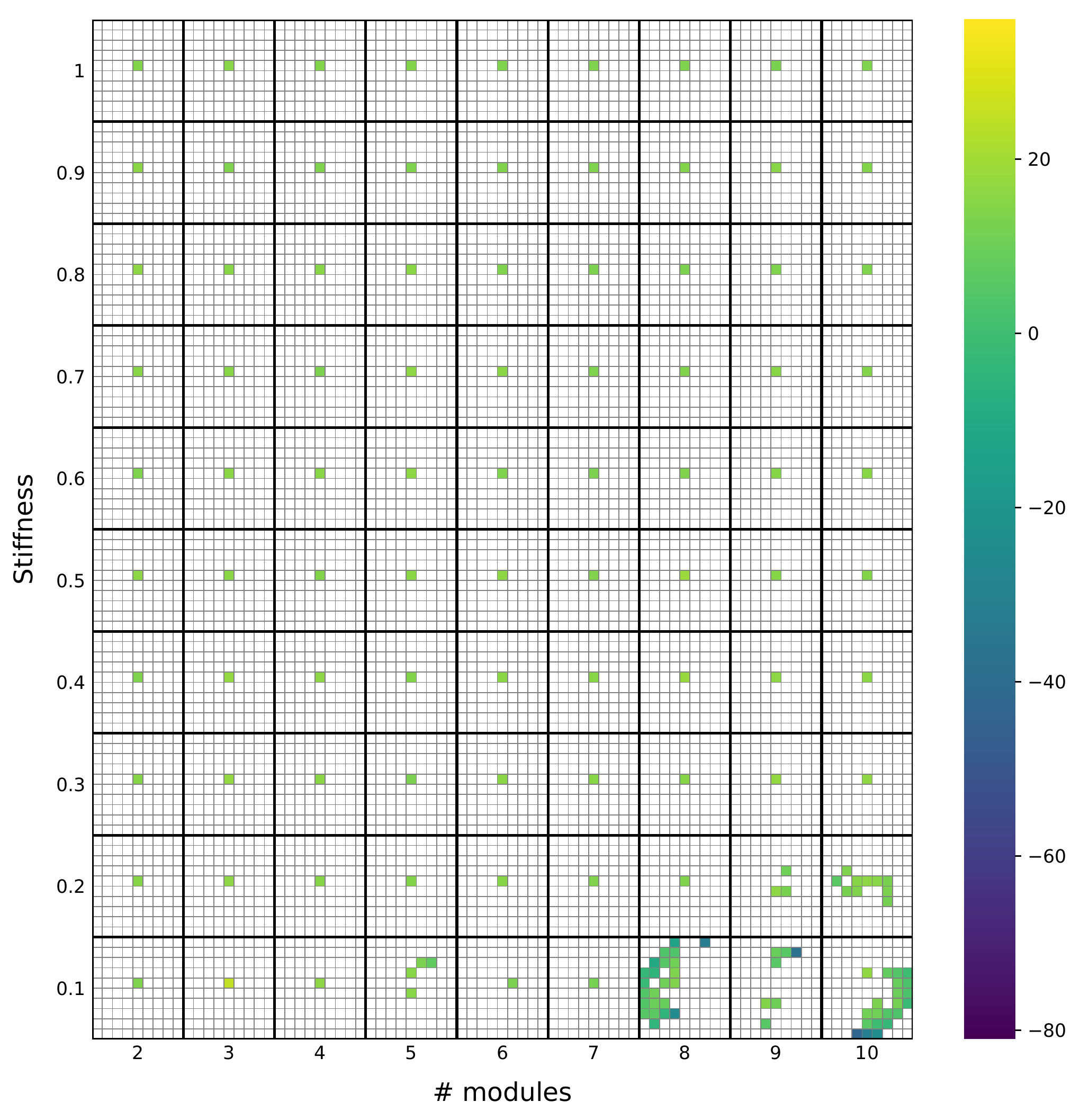}
      \label{fig:s-heat-sgr-dm-me-5}
    \end{subfigure} 
    \begin{subfigure}{0.195\textwidth}
      \centering
      \includegraphics[width=\textwidth]{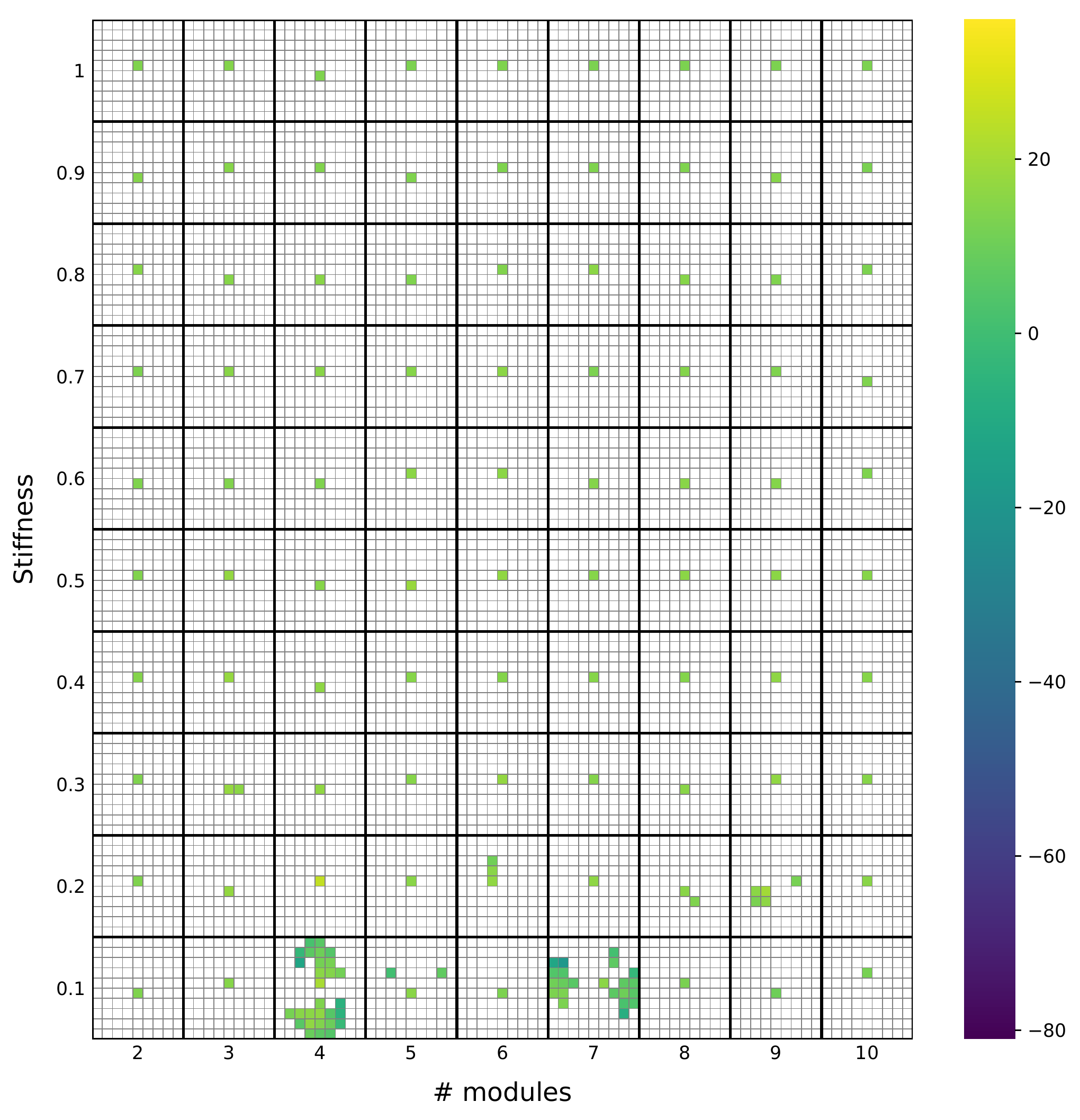}
      \label{fig:s-heat-sgr-dm-me-6}
    \end{subfigure} 
    \begin{subfigure}{0.195\textwidth}
      \centering
      \includegraphics[width=\textwidth]{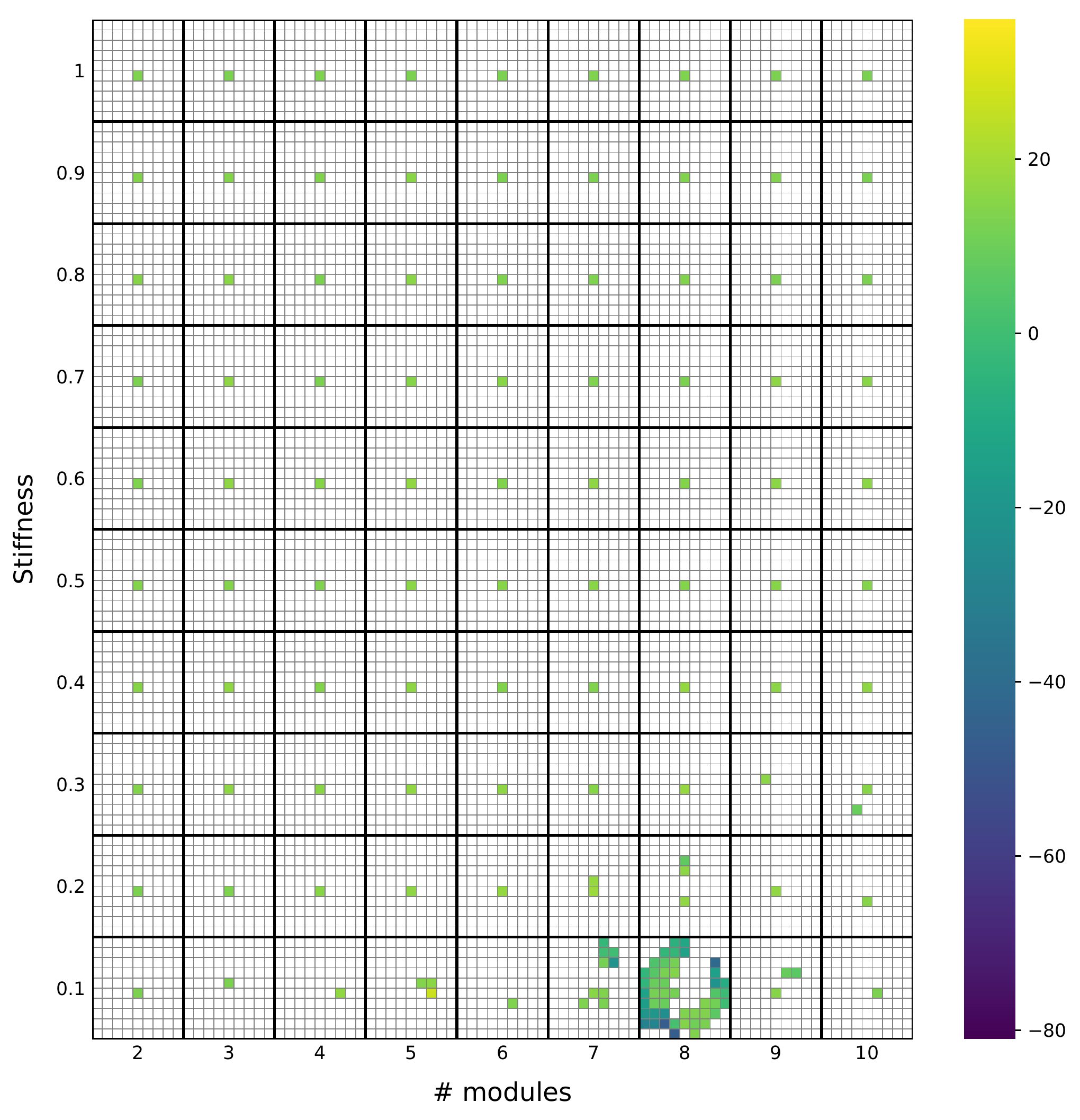}
      \label{fig:s-heat-sgr-dm-me-7}
    \end{subfigure} 
    \begin{subfigure}{0.195\textwidth}
      \centering
      \includegraphics[width=\textwidth]{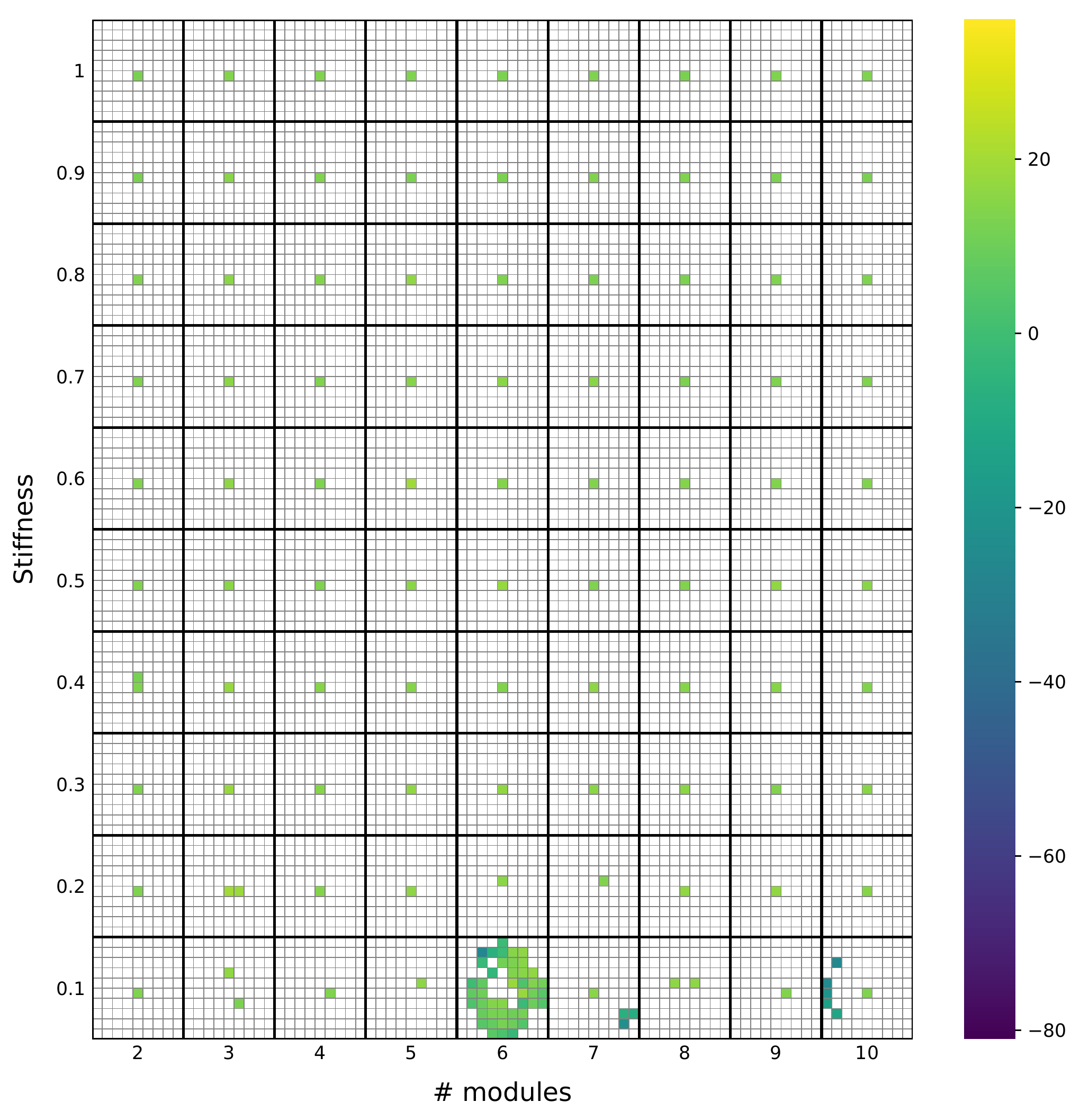}
      \label{fig:s-heat-sgr-dm-me-8}
    \end{subfigure} 
    \begin{subfigure}{0.195\textwidth}
      \centering
      \includegraphics[width=\textwidth]{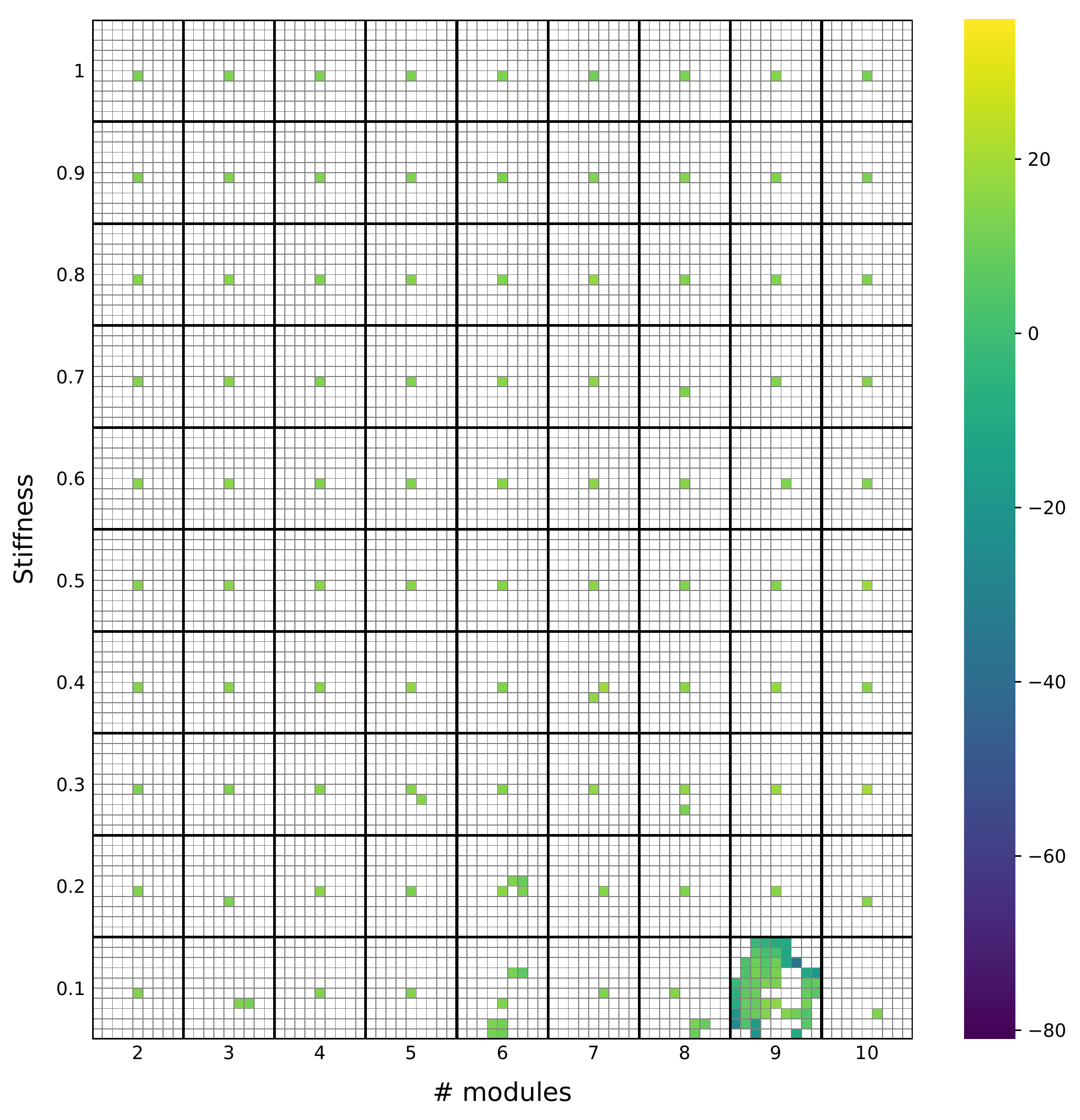}
      \label{fig:s-heat-sgr-dm-me-9}
    \end{subfigure}
    \caption{\textbf{Single archives for the squeezing task. In detail, the first two rows show the single archives generated by the 10 runs of MAP-Elites, whereas the other two show the projection of the double archives produced by Double Map MAP-Elites (DM-ME) onto a single one.}}
    \label{fig:s-heats-sgr}
\end{figure*}

\end{document}